\documentclass[pdflatex,sn-mathphys-num]{sn-jnl}% Math and Physical Sciences Numbered Reference Style
\usepackage{graphicx}%
\usepackage{multirow}%
\usepackage{makecell}
\usepackage{amsmath,amssymb,amsfonts}%
\usepackage{amsthm}%
\usepackage{mathrsfs}%
\usepackage[title]{appendix}%
\usepackage{xcolor}%
\usepackage{textcomp}%
\usepackage{manyfoot}%
\usepackage{booktabs}%
\usepackage{algorithm}%
\usepackage{algorithmic}%
\usepackage{listings}%
\usepackage{subcaption}
\usepackage{geometry}
\usepackage[export]{adjustbox}%%%%
\usepackage{bm}
\usepackage{xspace}
\usepackage{pifont}
\usepackage{bbding}
\usepackage{lineno}

\newcommand{\model}{TNFL\xspace}

\theoremstyle{thmstyleone}%
\theoremstyle{thmstyletwo}%

\theoremstyle{thmstylethree}%

\begin{document}

\title[A Trust-Network-Based Federated Learning Framework for Multi-Center Aging Clock Prediction]{A Trust-Network-Based Federated Learning Framework for Multi-Center Aging Clock Prediction}

%%=============================================================%%
%% GivenName	-> \fnm{Joergen W.}
%% Particle	-> \spfx{van der} -> surname prefix
%% FamilyName	-> \sur{Ploeg}
%% Suffix	-> \sfx{IV}
%% \author*[1,2]{\fnm{Joergen W.} \spfx{van der} \sur{Ploeg} 
%%  \sfx{IV}}\email{iauthor@gmail.com}
%%=============================================================%%

\author[1]{\fnm{Chunxu Zhang}}\email{chunxu.zhang@polyu.edu.hk}

\author[1]{\fnm{Bo Li}}\email{cnroselearn@gmail.com}

\author[2]{\fnm{Wenliang Wang}}\email{barry@quantumlife.tech}

\author[1,4]{\fnm{Yang Liu}}\email{yang-veronica.liu@polyu.edu.hk}

\author[1,4]{\fnm{Di Jiang}}\email{di-prof.jiang@polyu.edu.hk}

\author[2]{\fnm{Yuan Huang}}\email{christine@quantumlife.tech}

\author[5]{\fnm{Yo-ichi Nabeshima}}\email{nabeshima.yoichi.7n@kyoto-u.ac.jp}

\author[5]{\fnm{Akinori Yamamura}}\email{yamamura.akinori.4y@kyoto-u.ac.jp}

\author[6]{\fnm{Bo Yang}}\email{ybo@jlu.edu.cn}

\author*[1,3]{\fnm{Qiang Yang}}\email{profqiang.yang@polyu.edu.hk}

\affil*[1]{\orgdiv{PolyU Academy for Artificial Intelligence}, \orgname{The Hong Kong Polytechnic University}, \orgaddress{\city{Hung Hom}, \state{Hong Kong}, \country{China}}}

\affil[2]{\orgdiv{Quantum Life}, \orgaddress{\city{Hong Kong}, \country{China}}}

\affil[3]{\orgdiv{Department of Data Science and Artificial Intelligence}, \orgname{The Hong Kong Polytechnic University}, \orgaddress{\city{Hung Hom}, \state{Hong Kong}, \country{China}}}

\affil[4]{\orgdiv{Department of Computing}, \orgname{The Hong Kong Polytechnic University}, \orgaddress{\city{Hung Hom}, \state{Hong Kong}, \country{China}}}

\affil[5]{\orgdiv{Department of Aging Science and Medicine}, \orgname{Graduate School of Medicine, Kyoto University}, \orgaddress{\city{Kyoto}, \country{Japan}}}

\affil[6]{\orgdiv{College of Computer Science and Technology}, \orgname{Jilin University}, \orgaddress{\city{Changchun}, \state{Jilin}, \country{China}}}

%%==================================%%
%% Sample for unstructured abstract %%
%%==================================%%

\abstract{
Aging clocks quantify biological aging and provide a means to characterize individual health status. What kinds of protein interactions are important for determining a high-quality aging clock? Are these interactions zeroth-order or higher-order? Addressing these questions requires learning from large and diverse molecular datasets distributed across multiple medical centers. However, privacy and security constraints often prevent individual-level data from being centrally shared, resulting in isolated data silos. Federated learning offers a natural way to enable collaborative learning without centralizing raw data, but its application to this setting faces four key challenges: First, limited local sample sizes constrain reliable predictive modeling and make local generative modeling particularly difficult at data-limited centers. Second, inter-center collaboration is often governed by sparse and directional trust relations rather than a globally trusted coordinator. Third, aging clocks are primarily formulated as discriminative predictors and should retain age prediction while supporting interpretation of the learned age-related patterns. Finally, heterogeneous cross-center data can drive model drift and weaken previously acquired knowledge during continued collaborative learning. To address these challenges, we propose \model, a trust-network-based federated learning framework for multi-center modeling, which we used to further answer the above biological questions.

Specifically, \model organizes federated learning as progressive model propagation over directed pairwise trust relations, where a trust relation indicates whether a center can trust another center with information privacy and security. We build a model progressively along a path of trust relations without centralized aggregation. \model combines an age-aware mixture-of-experts model with generative replay, using generated pseudo-samples to preserve previously learned information and reduce forgetting and model drift. Experiments across multiple molecular datasets show that \model enables effective aging-clock prediction with limited local data. \model provides interpretable age-dependent prediction patterns and maintains stable performance across interaction orders without systematic forgetting. 

To answer the above biological questions concerning protein relevance to aging and the order of protein interactions, we analyze the model-identified pairwise protein interactions using \model and then investigate how these relationships organize at higher orders through functional and network analyses. The identified interactions repeatedly form coordinated higher-order subnetworks spanning multiple aging-related biological systems, with several proteins repeatedly appearing across different subnetworks. These findings indicate that the model captures molecular relationships that extend beyond isolated pairwise associations and exhibit coherent higher-order biological organization associated with aging.

}

\keywords{Biological Aging Clock, Trust Network, Sequential Federated Learning}

%%\pacs[JEL Classification]{D8, H51}

%%\pacs[MSC Classification]{35A01, 65L10, 65L12, 65L20, 65L70}
% \linenumbers

\maketitle
Aging clocks estimate biological age from molecular measurements and provide quantitative indicators of aging-related biological changes~\cite{jylhava2017biological,argentieri2024proteomic}. These models capture inter-individual variation in aging and are associated with functional decline, age-related diseases, and mortality risk~\cite{fransquet2019epigenetic,warner2024systematic}. High-quality aging clocks can help identify which protein interactions are important for age prediction and characterize both individual protein effects and higher-order interactions. Investigating these questions increasingly relies on large and diverse datasets that capture variation across populations~\cite{rutledge2022measuring}. However, such datasets are often distributed across multiple medical centers~\cite{min2024critical} and cannot be centrally shared~\cite{bonomi2020privacy,wan2022sociotechnical}. This setting therefore calls for a federated learning framework that can learn jointly from data distributed across different centers.
% aging clock重要，高质量的aging clocl可以帮助我们回答这两个问题，但是我们现在面临分布式场景，而且有四个challenges。

% two linked contributions. bio和算法结构和文字的平衡都要对应，逻辑保持和摘要一致。

% Aging clocks estimate biological age from molecular measurements and provide quantitative indicators of aging-related biological changes~\cite{jylhava2017biological,argentieri2024proteomic}. These models capture inter-individual variation in aging and are associated with functional decline, age-related diseases, and mortality risk~\cite{argentieri2024proteomic,fransquet2019epigenetic,warner2024systematic}. Reliable aging-clock development increasingly depends on large and diverse datasets that capture variation across populations~\cite{rutledge2022measuring}. Yet such datasets are often distributed across multiple centers~\cite{min2024critical} and cannot be centrally shared~\cite{bonomi2020privacy,wan2022sociotechnical}, making reliable aging-clock development inherently a multi-center collaborative learning problem.

However, applying federated learning on multi-center aging-clock modeling faces four key challenges. \textbf{First, limited local data pose a fundamental constraint on both predictive and generative modeling.} Small centers may lack sufficient samples to train reliable aging clocks independently, while the same data scarcity also makes local generative modeling difficult. \textbf{Second, cross-center collaboration is restricted by partial and asymmetric trust.} In practice, centers may permit model communication only with selected partners, and these relations can be directional rather than globally shared among participating centers. \textbf{Third, aging-clock modeling calls for discriminative and interpretable prediction.} Aging clocks should maintain reliable predictive performance while also enabling interpretation of the age-related patterns learned by the model. \textbf{Finally, cross-center heterogeneity introduces model drift and knowledge forgetting.} Continued adaptation to different local data distributions may shift the model toward recently encountered patterns and weaken information acquired from earlier centers. Together, these challenges characterize the capabilities required for a federated learning framework to support reliable multi-center aging-clock modeling.

% Beyond age prediction, aging clocks also provide a basis for examining molecular changes associated with biological aging~\cite{argentieri2024proteomic}. Aging itself reflects coordinated alterations across multiple cellular and physiological processes rather than isolated molecular changes~\cite{lopez2023hallmarks}. Recent molecular aging studies further suggest that age-related signals can exhibit coherent organization across biological systems and organs~\cite{wang2025organ}. This motivates an important biological question in the multi-center setting: \textbf{whether models learned from multiple small and heterogeneous populations can capture coordinated biological patterns associated with aging.}

% 放在discussion里面，再找计算机里面做aging clock，没有考虑生物学分析。这里面先给出结论，大概是算法层面和生物层面都没有做好，然后指引具体内容见后面章节
% dicussion先把实验结论给个摘要，然后就说别人没做过，最后说limitation and future，一共三个部分。
Standard federated learning relies on a central server trusted by all participating centers for model coordination and aggregation, an assumption that does not hold under sparse and directional inter-center trust. Beyond this conventional setting, trust-related federated~\cite{zhang2020enabling,chen2023trustnetfl,he2019central}, decentralized federated learning~\cite{warnat2021swarm,sun2022decentralized}, and sequential federated learning~\cite{chang2018distributed,kampfederated,wang2024one} have explored alternative collaboration mechanisms. These approaches each address part of the challenges arising in multi-center aging-clock modeling, but do not provide a single framework that addresses them together. In parallel, machine-learning and deep-learning studies on aging clocks have increasingly moved beyond age prediction toward biological interpretation~\cite{galkin2021deepmage,de2022pan}, including the identification of age-relevant molecular features~\cite{vijayakumar2022pan} and their associated biological pathways~\cite{martinez2023ncae,prosz2024biologically,lin2026deepstrataage}. However, how model-identified molecular interactions organize across different orders, particularly into higher-order relationships, remains less explored. Taken together, existing work does not yet provide a unified approach that simultaneously supports effective multi-center collaboration and examines which protein interactions are relevant to aging-clock prediction and how they organize across different orders. We therefore propose \model to address both aspects within a unified multi-center framework.

% Standard federated learning provides the canonical paradigm for collaborative model training across distributed centers, typically relying on a central server to coordinate local updates and aggregate them into a shared global model~\cite{mcmahan2017communication}. This centralized organization assumes that participating centers can rely on a common coordinating party for model aggregation. Beyond this conventional setting, several lines of federated learning have explored alternative collaboration structures. Trust-related FL incorporates trust into collaborative learning to characterize participant reliability or regulate communication among selected parties~\cite{zhang2020enabling,chen2023trustnetfl,he2019central}, with OPS~\cite{he2019central} further supporting collaboration over a directed trust graph. Decentralized FL removes the fixed central aggregator and coordinates learning through model exchange, aggregation, or synchronization among connected participants~\cite{warnat2021swarm,sun2022decentralized}. Sequential FL instead transfers models across centers so that information can accumulate through successive local updates~\cite{chang2018distributed,kampfederated,wang2024one}; CWT~\cite{chang2018distributed} and FedELMY~\cite{wang2024one} employ cumulative cross-center training, while FedDC~\cite{kampfederated} transfers multiple models with periodic aggregation. FedELMY further constrains local models relative to the preceding model to reduce drift and preserve previously acquired information. 

\begin{figure}[!t]
  \centering          
  \includegraphics[width=1.\textwidth]{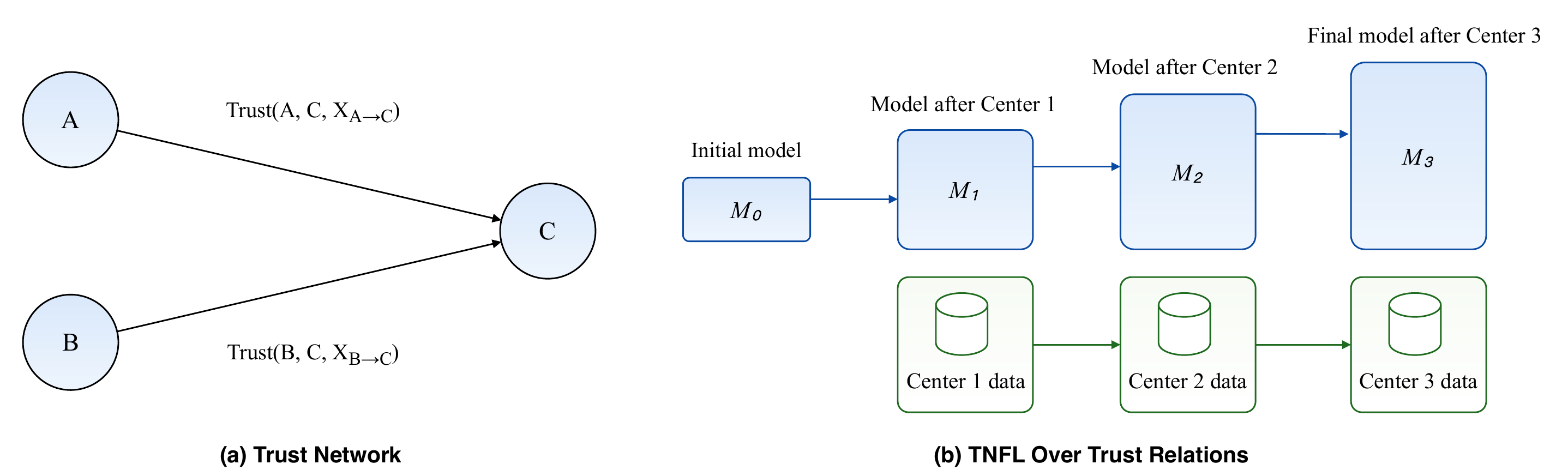} 
  \vspace{1pt}
  \caption{Trust-constrained model propagation in \model. (a) Directed pairwise trust relations define which model transmissions are permitted between centers, forming the trust network over which collaboration can proceed. (b) \model realizes progressive cross-center learning along an admissible trust sequence. The aging-clock model is updated at each visited center, with its model capability progressively enhanced as information from more centers is incorporated.
  % (b) \model realizes progressive cross-center learning along an admissible trust sequence, where a single aging-clock model evolves cumulatively and is updated using the local data of each visited center.
  }
  \label{fig:motivation} 
\end{figure}

% As summarized in Table~\ref{tab:related_work_comparison}, the three alternative federated learning paradigms introduced above address different aspects of multi-center collaboration, but each covers only part of the four computational challenges identified above. To address these challenges, we propose \model, a trust-network-based federated learning framework for multi-center modeling, and further use the learned model to address the biological questions posed above.

Specifically, \model organizes federated learning as progressive model propagation over directed pairwise trust relations without relying on a globally trusted aggregation server. As illustrated in Fig.~\ref{fig:motivation}(a), directed pairwise trust relations define which model transmissions are permitted between centers and collectively form the trust network. Each edge governs an authorized model transmission between two centers, and \model constructs valid trust sequences over this network. Fig.~\ref{fig:motivation}(b) illustrates how the evolving model is progressively updated along an admissible trust sequence, with each visited center continuing from the model state accumulated through preceding updates. This allows centers with limited local data to benefit from information incorporated earlier in the sequence. For aging-clock modeling, \model employs an age-aware mixture-of-experts predictor. Its routing behavior supports interpretation of age-dependent prediction patterns. \model further incorporates generative replay to preserve information across heterogeneous center updates and reduce model drift and knowledge forgetting. Experiments across multiple molecular datasets show that \model remains effective when local data are limited. \model provides interpretable age-dependent prediction patterns and also maintains stable performance across different interaction orders without systematic forgetting.

To further address the biological questions concerning which protein relationships are relevant to aging-clock prediction and whether these relationships extend to higher-order interactions, we first characterize the biological coherence of the protein interactions identified by \model. Functional analyses show that these interactions are associated with multiple aging-related biological processes, while several pairs are also supported by known functional associations. We then examine whether these pairwise relationships further organize at higher orders through complementary analyses based on age prediction and expert-routing outputs across different population settings. Across these analyses, the identified protein groups repeatedly form coordinated higher-order subnetworks spanning multiple aging-related biological systems, with several proteins repeatedly appearing across different subnetworks. Together, these results show that the protein relationships identified by the model are not limited to isolated pairwise associations but exhibit coherent higher-order organization associated with aging. Importantly, these patterns emerge without incorporating prior protein-interaction or pathway knowledge during model training, further supporting their biological coherence.

\section{Results}
In Section 1.1, we use \model to investigate the biological questions of which proteins are relevant to aging-clock prediction and how their effects and interactions are organized across different orders. In Section 1.2, we evaluate \model computationally against the key challenges of multi-center aging-clock modeling, focusing on predictive performance and robustness.

\begin{figure}[!t]
    \centering
    % 全局间距优化：减小列间距，保证布局紧凑
    \setlength{\tabcolsep}{0.5em}
    \renewcommand{\arraystretch}{1.0}

    % ========== 第一排：左侧文字块(0.48) + 右侧子图a(0.48) ==========
    \begin{subfigure}[t]{0.5\textwidth}
        \centering
        \includegraphics[width=\linewidth]{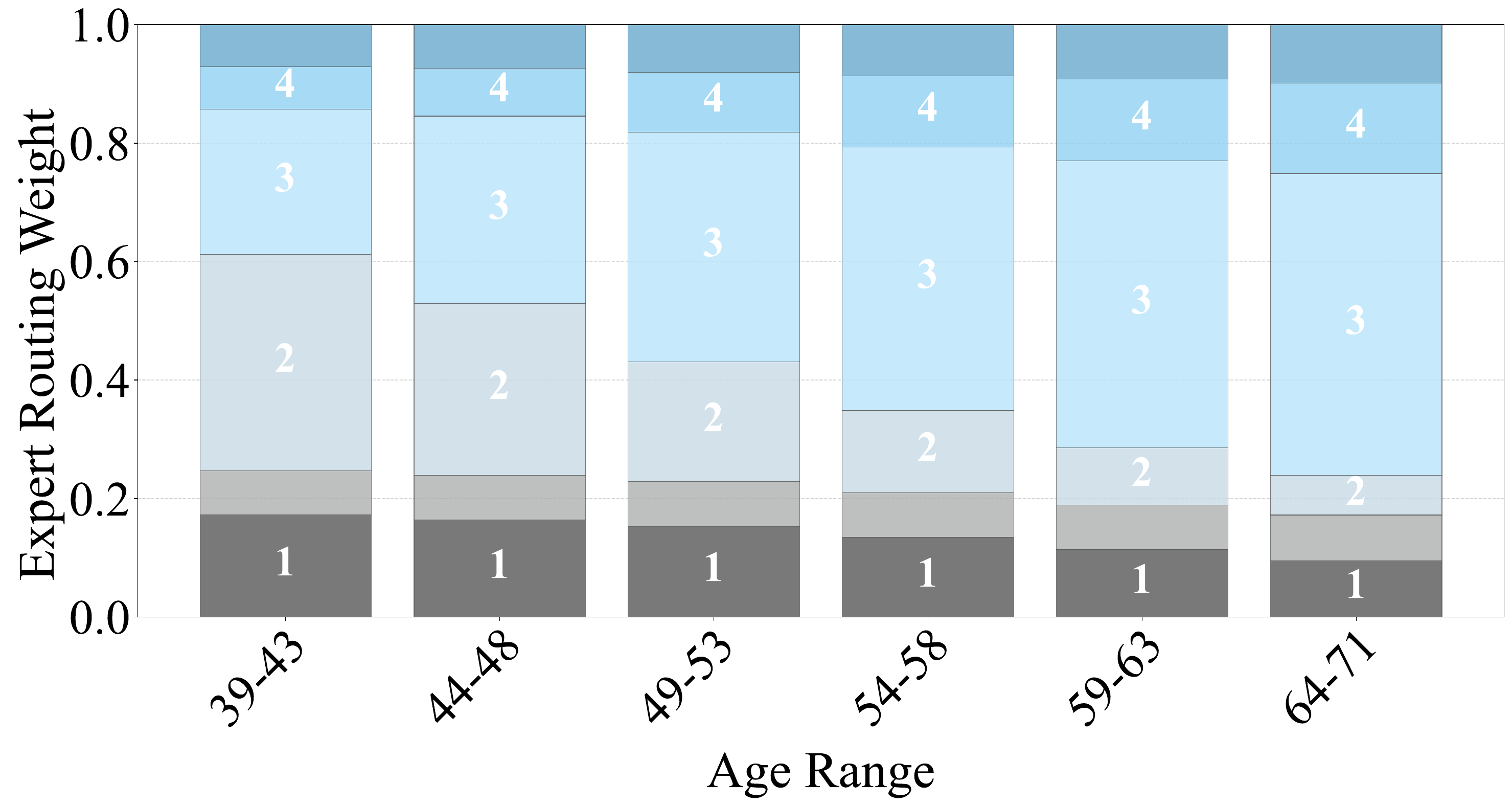}
        \subcaption{UKB-Center}
    \end{subfigure}

    \par
    \vspace{4pt}
    \noindent\makebox[\linewidth]{\dotfill}
    \vspace{4pt}

    % ========== 第二排：子图==========
    \begin{subfigure}[t]{\textwidth}
        \centering
        \includegraphics[width=0.9\linewidth]{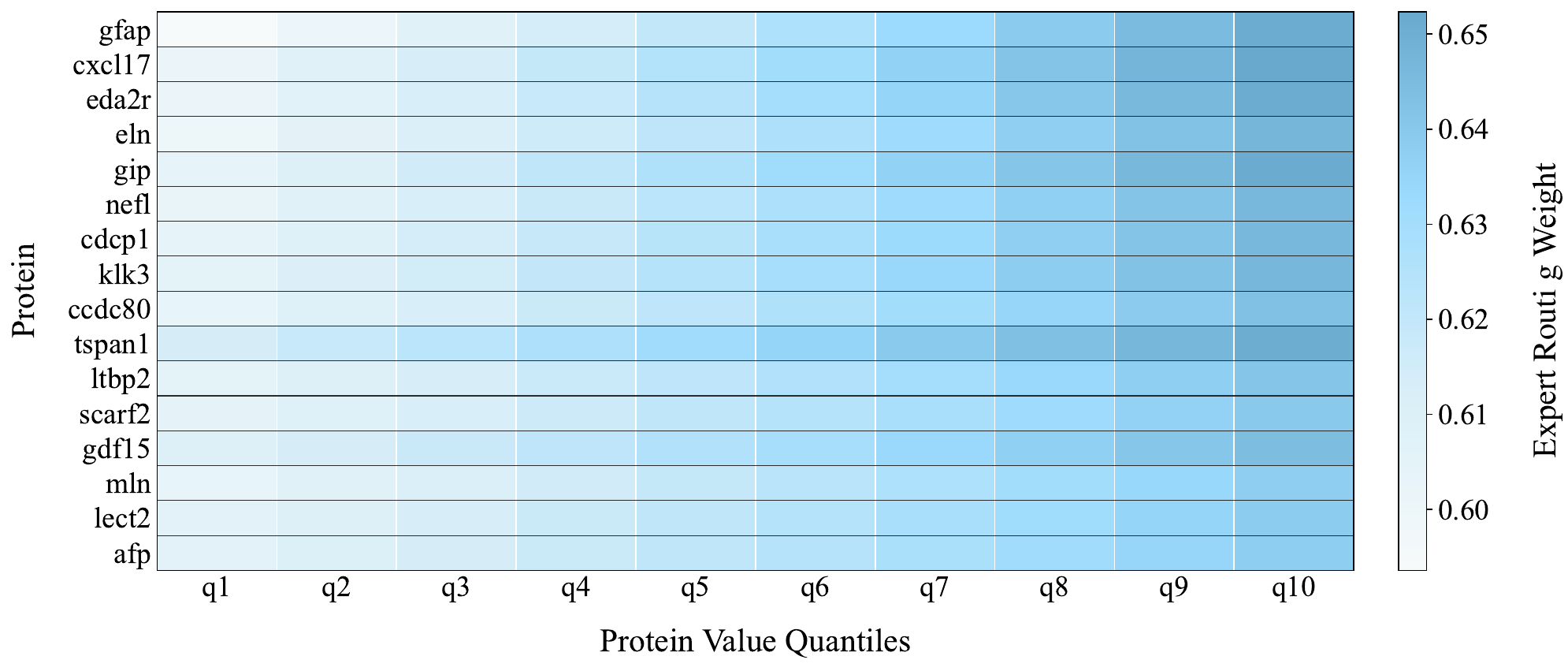}
        \subcaption{Top-ranked proteins with monotonically increasing expert routing weights}
        % \caption{Subfig b}
        % \label{subfig:b}
    \end{subfigure}

    \par
    \vspace{4pt}
    \noindent\makebox[\linewidth]{\dotfill}
    \vspace{4pt}

    % ========== (c) Functional enrichment ==========
    \begin{subfigure}[t]{\textwidth}
        \centering
        \includegraphics[width=\linewidth]{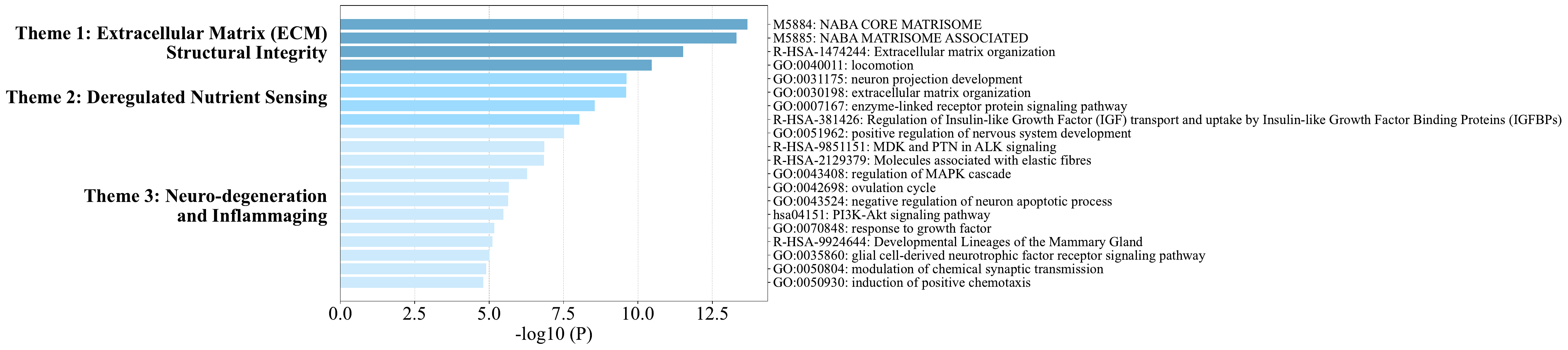}
        \subcaption{Functional and pathway enrichment of top-100 proteins}
    \end{subfigure}

    \caption{Visualization of expert routing patterns across samples (a), top-ranked proteins showing a positive relationship with Expert 3 routing weight, where higher protein values correspond to higher weight (b), and functional and pathway enrichment of the top-100 proteins ranked by their influence on expert routing (c).}
    \label{fig:analysis}
\end{figure}
\subsection{Biological Findings from \model}
Biological analyses are conducted using the main \model implementation on the UKB-Center proteomic dataset. We first use the age-dependent routing behavior of AgeMoE to identify proteins associated with aging-related prediction patterns. We then examine these molecular effects across increasing interaction orders, from individual proteins to pairwise interactions and higher-order protein organizations, to characterize both their biological relevance and their broader coordination in aging.

\subsubsection{Individual Protein Relevance to Aging-Clock Prediction} 
We first examine the age-dependent routing behavior of the AgeMoE predictor used in the main \model implementation and use it to identify individual proteins associated with aging-related prediction patterns. In AgeMoE, each expert is a learnable prediction sub-module, and a routing network assigns sample-specific weights to combine the outputs of different experts. We therefore analyze these routing weights to examine how the model organizes its predictions across age groups. Samples in the global test set are grouped into age intervals, and the average routing weight assigned to each expert is computed within each group. 

As shown in Fig.~\ref{fig:analysis}(a), experts display distinct age-dependent routing patterns: some maintain relatively stable contributions across ages, whereas others receive systematically higher or lower weights in older or younger groups. For example, in the UKB-Center dataset, the routing weights of Experts 1 and 2 decrease with age, whereas those of Experts 3 and 4 increase. These results indicate that AgeMoE adjusts the relative contributions of its prediction sub-modules according to age, providing an interpretable view of how its predictive behavior changes across age groups. 

To further connect these age-dependent routing patterns to molecular features, we analyze how individual protein values influence the routing weight assigned to Expert 3 in the UKB-Center proteomic dataset. Expert 3 is selected because its routing weight increases with sample age (Fig.~\ref{fig:analysis}(a)), making it informative for examining molecular features associated with this age-dependent model behavior. To assess protein-specific effects, each protein is systematically varied across ten quantile levels while keeping all other features fixed, and the resulting changes in the routing weight of Expert 3 are recorded. We quantify each protein's influence as the signed change in Expert 3 routing weight induced by increasing that protein from its lowest to highest quantile level while keeping all other features fixed, and rank proteins according to this score. 

As shown in Fig.~\ref{fig:analysis}(b), we highlight a subset of the top 20 proteins with positive effects on Expert 3 routing, for which higher protein values lead to higher routing weights. Because Expert 3 is increasingly utilized at older ages, these proteins are associated with a model-routing pattern that becomes more prominent with age. Several highlighted proteins, including GDF15, GFAP, and NEFL, have previously been reported to increase with age~\cite{tanaka2018plasma,pereira2021plasma,lewczuk2018plasma}, supporting the consistency of the learned routing patterns with known age-associated molecular changes. 

We next examine the broader biological relevance of these model-identified proteins. Using the UKB-Center proteomic dataset, we selected the top 100 proteins ranked by their influence on expert routing for downstream biological characterization. Functional enrichment analysis using Metascape~\cite{zhou2019metascape} revealed 20 significantly enriched ontology clusters, which can be summarized into three major biological themes: Extracellular Matrix (ECM) Structural Integrity, Deregulated Nutrient Sensing, and Neuro-degeneration and Inflammaging (Fig.~\ref{fig:analysis}(c)). 

We further assessed biological relevance by comparing these proteins against curated aging-related databases. Several proteins overlap with GenAge~\cite{tacutu2012human} and the SASP Atlas~\cite{basisty2020proteomic}, including SOD2, ELN, EFEMP1, COL6A3, GDF15, and PTX3. In addition, we evaluated protein localization and tissue origin using secretome annotations and data from the Human Protein Atlas~\cite{uhlen2015tissue}. The results indicate that most proteins are secreted or ECM-associated, with additional expression in brain and other major tissues such as vascular, liver, lung, and skin. Single-cell annotations further suggest expression across neural, stromal, and immune cell types. Collectively, these analyses show that the model-identified proteins are functionally enriched in aging-related processes and span multiple tissues and cell types relevant to systemic aging. 

\subsubsection{Pairwise Protein Interactions}
We next extend the analysis from individual protein effects to pairwise protein interactions identified by the model. The top-10 model-identified protein pairs were further examined using STRING functional association analysis~\cite{szklarczyk2021string}. Three pairs, (GIP, CGA), (INSL3, CGA), and (GIP, INSL3), showed known STRING-supported associations related to endocrine and metabolic signaling. The remaining pairs showed no direct STRING links but exhibited coherent functional convergence. GFAP-related pairs were associated with neuroinflammatory and neuroendocrine processes~\cite{yang2015glial}, while (GFAP, NPPB) reflected neural and cardiac stress signaling~\cite{daniels2007natriuretic}. Other pairs, including (EDA2R, CGA) and CCDC80-related pairs, involved inflammatory~\cite{barbera2025increased}, endocrine, and extracellular matrix remodeling functions. These results indicate that the identified pairwise interactions capture biologically coherent relationships that are not limited to direct interactions already represented in curated databases, motivating further analysis of whether these pairwise relationships organize into higher-order protein subnetworks.

\subsubsection{Higher-Order Organization of Protein Interactions} 
We therefore investigated whether model-identified pairwise interactions repeatedly organize into higher-order protein subnetworks. Higher-order groups were examined through STRING analysis from complementary analytical perspectives, including protein coordination associated with final biological age predictions and with latent AgeMoE routing behavior, as well as analyses over the full population and age-restricted young subsets. These complementary settings allow us to examine whether coordinated protein patterns recur across different views of the learned aging signals rather than being specific to a single analysis setting. 

\begin{figure}[!t]
    \centering
    % 全局间距优化：减小列间距，保证布局紧凑
    \setlength{\tabcolsep}{0.5em}
    \renewcommand{\arraystretch}{1.0}

    % ========== 第一排：子图==========
    \begin{subfigure}[t]{0.48\textwidth}
        \centering
        \includegraphics[width=\linewidth]{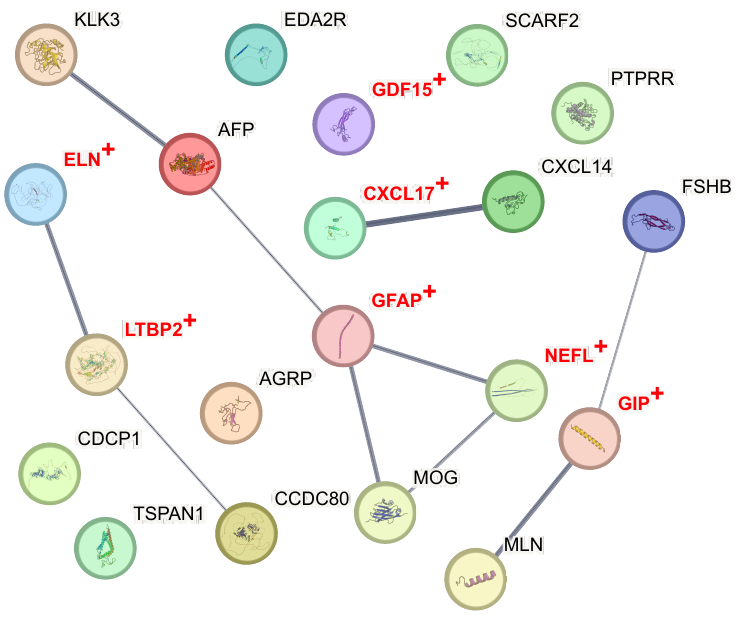}
        \subcaption{Protein coordination under full-population expert routing ($P = 5.47 \times 10^{-6}$)}
        % \label{subfig:b}
    \end{subfigure}
    \hfill
    \begin{subfigure}[t]{0.48\textwidth}
        \centering
        \includegraphics[width=\linewidth]{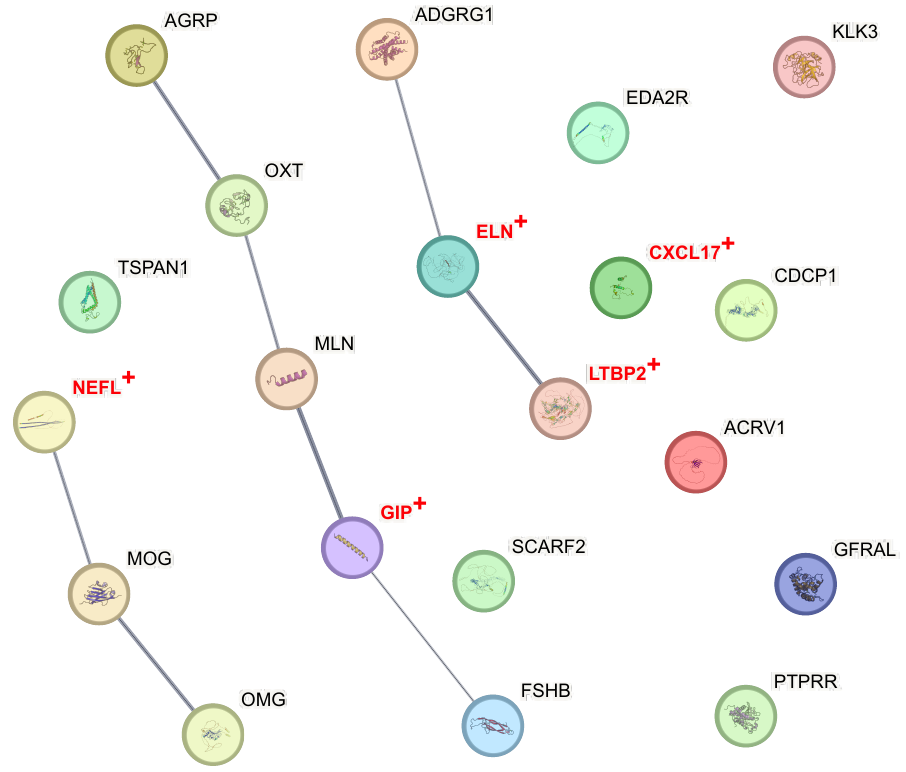}
        \subcaption{Protein coordination under full-population age prediction ($P = 6.83 \times 10^{-6}$)}
        % \label{subfig:c}
    \end{subfigure}

    \par
    % \vspace{1pt}
    \noindent\makebox[\linewidth]{\dotfill}
    % \vspace{1pt}

    % ========== 第三排：左侧文字块(0.48) + 右侧子图d(0.48) ==========
    \begin{subfigure}[t]{0.48\textwidth}
        \centering
        \includegraphics[width=\linewidth]{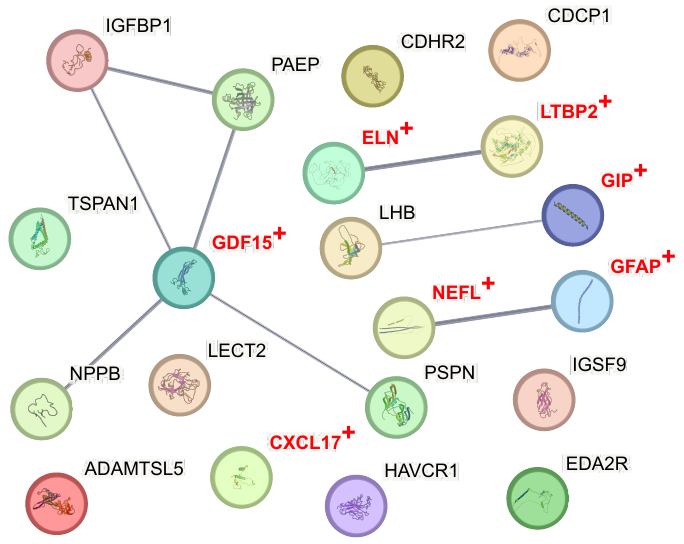}
        \subcaption{Protein coordination under young-subset expert routing ($P = 5.85 \times 10^{-6}$)}
        % \label{subfig:e}
    \end{subfigure}
    \hfill
    \begin{subfigure}[t]{0.48\textwidth}
        \centering
        \includegraphics[width=\linewidth]{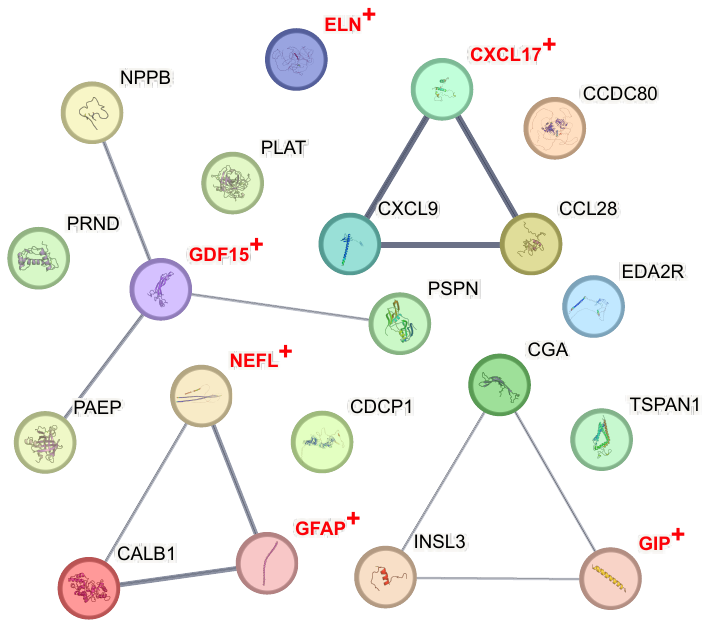}
        \subcaption{Protein coordination under young-subset age prediction ($P = 2.17 \times 10^{-7}$)}
        % \label{subfig:f}
    \end{subfigure}
    \caption{Biological interpretation of STRING-derived synergistic protein subnetworks identified under complementary analytical perspectives (a)-(d) on the UKB-Center dataset. In subnetworks (a)-(d), nodes represent proteins and edges denote STRING functional associations, with thicker edges indicating stronger association confidence. Recurrent proteins appearing across multiple subnetworks are highlighted in red and marked with $^+$. The reported $P$ values correspond to STRING functional association enrichment, assessing whether the observed connectivity within a protein group exceeds that expected for random protein sets with comparable background connectivity.}
    \label{fig:bio_analysis}
\end{figure}

Across these complementary perspectives, the identified subnetworks repeatedly converged on coherent aging-related biological organizations. One recurrent organization centered on coordinated neurovascular-endocrine modules involving neural injury markers (GFAP, NEFL), extracellular matrix remodeling proteins (ELN, LTBP2), and endocrine-metabolic regulators (GIP, FSHB) (Fig.~\ref{fig:bio_analysis}(a))~\cite{cai2017dysfunction}. A second organization further incorporated hypothalamic and neuroendocrine regulators, including AGRP and OXT, linking hormonal coordination with metabolic and neural processes (Fig.~\ref{fig:bio_analysis}(b))~\cite{yoo2021neuroendocrine}. Another recurrent module emphasized systemic stress and cardiometabolic signaling through recurrent inclusion of GDF15 and NPPB (Fig.~\ref{fig:bio_analysis}(c))~\cite{wang2021gdf15,zois2014natriuretic}. The most integrated organization ultimately converged on broader multi-system coordination simultaneously involving inflammatory chemokines, neural injury markers, endocrine regulators, stress-response proteins, and vascular remodeling factors (Fig.~\ref{fig:bio_analysis}(d))~\cite{li2023inflammation,chung2019redefining}. 

Notably, several proteins, including GFAP, NEFL, GDF15, GIP, ELN, LTBP2, and CXCL17, appeared repeatedly across these subnetworks, suggesting shared hub-like signals that connect multiple aging-related biological processes. Together, the recurrent appearance of these proteins and subnetworks indicates higher-order coordination across neural, vascular, endocrine, metabolic, inflammatory, and stress-related systems. This organization is in line with the hallmarks-of-aging framework~\cite{lopez2023hallmarks}, which links molecular and cellular hallmarks to broader integrative dysfunctions during aging. 

Taken together, these analyses show that the model-identified protein interactions are not limited to isolated pairwise associations, but repeatedly organize into higher-order subnetworks spanning multiple aging-related biological systems. Importantly, these patterns emerge without incorporating prior protein-interaction or pathway knowledge during model training. These findings demonstrate that aging-related molecular effects captured by \model extend from individual proteins and pairwise interactions to coherent higher-order organization. Additional details of the biological analyses are provided in Appendix~\ref{app:geo_bio_analysis}.

\begin{figure}[htbp]
    \centering
    \captionsetup[subfigure]{skip=1pt}
    % 全局间距优化：减小列间距，保证布局紧凑
    \setlength{\tabcolsep}{0.5em}
    \renewcommand{\arraystretch}{1.0}

    % ========== 第一排：左侧文字块(0.48) + 右侧子图a(0.48) ==========
    \begin{subfigure}[t]{0.48\textwidth}  % [t] 顶端对齐
        \centering
        \includegraphics[width=\linewidth, height=0.52\linewidth]{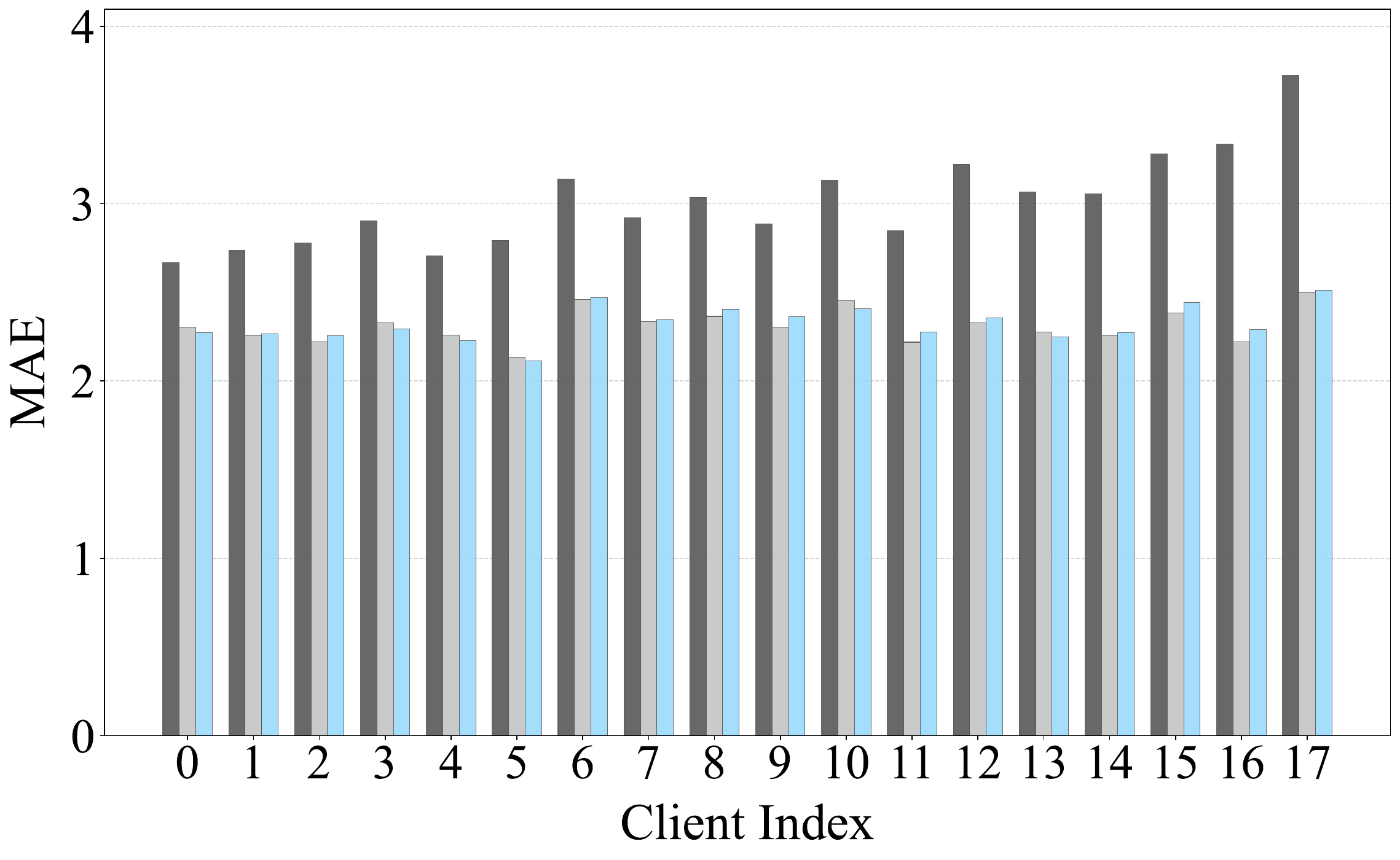}
        \subcaption{UKB-Center}
    \end{subfigure}
    \hfill  % 自动填充列间空白
    \begin{subfigure}[t]{0.48\textwidth}  % [t] 顶端对齐
        \centering
        \includegraphics[width=\linewidth, height=0.52\linewidth]{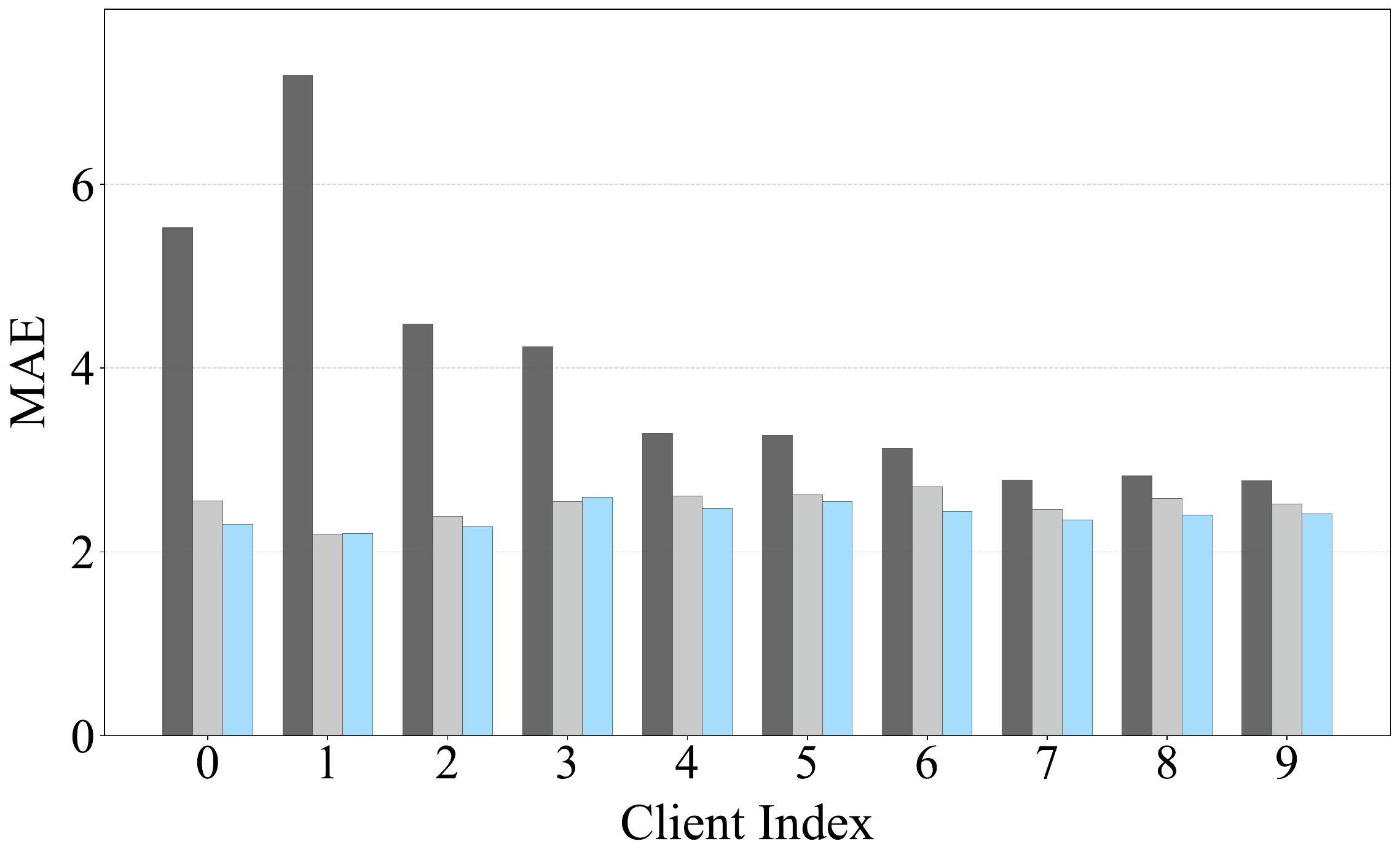}
        % 可选：添加子图标题（不需要则注释）
        \subcaption{UKB-Imbalance}
        % \label{subfig:2a}
    \end{subfigure}

    % \vspace{0.2em}  % 行间距，可按需调整

    % ========== 第二排：子图b(0.48) + 子图c(0.48) ==========
    \begin{subfigure}[t]{0.48\textwidth}
        \centering
        \includegraphics[width=\linewidth, height=0.52\linewidth, valign=t]{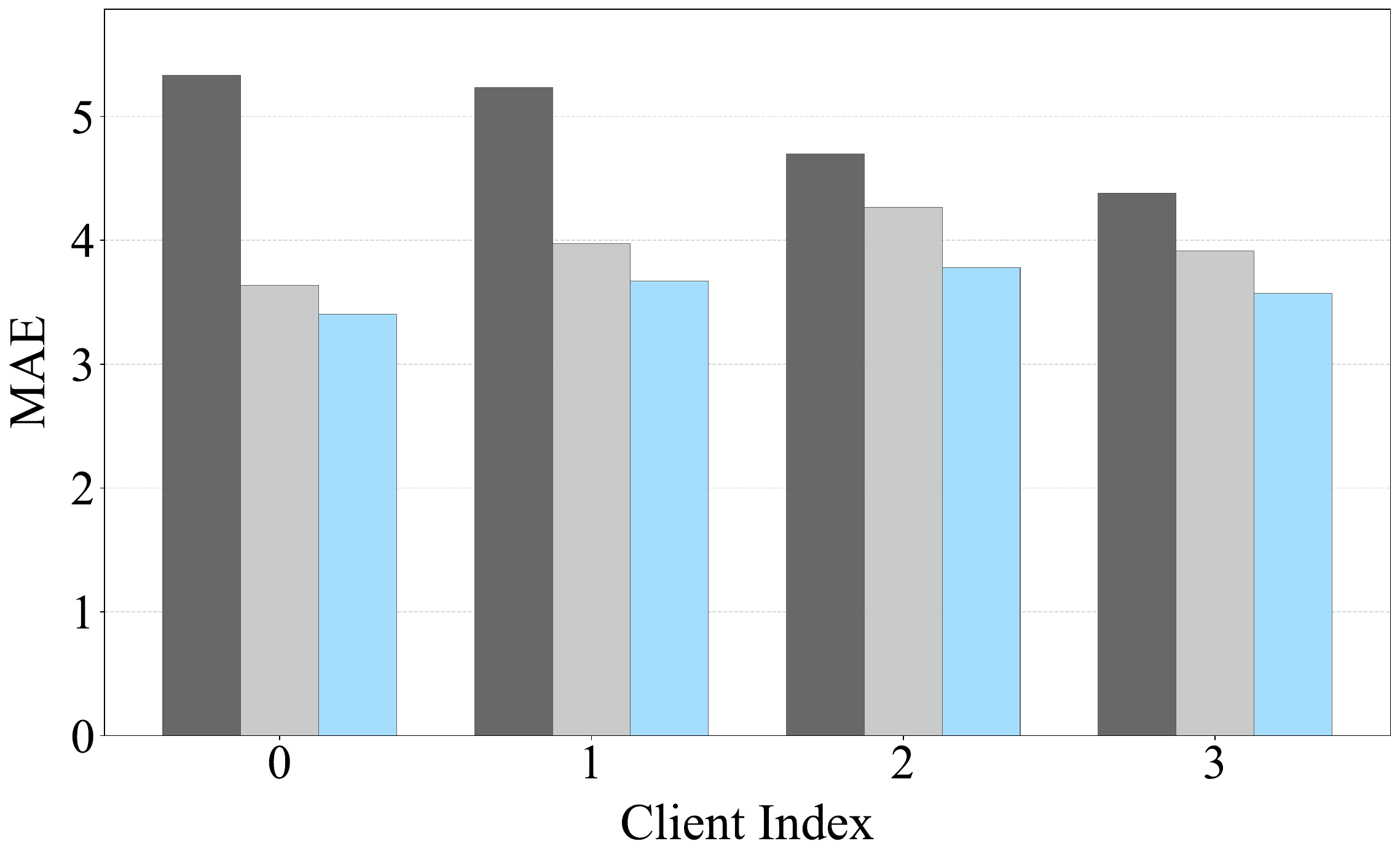}
        \subcaption{GEO-Methylation}
        % \label{subfig:b}
    \end{subfigure}
    \hfill
    \begin{subfigure}[t]{0.46\textwidth}
        \centering
        \includegraphics[width=\linewidth, valign=t]{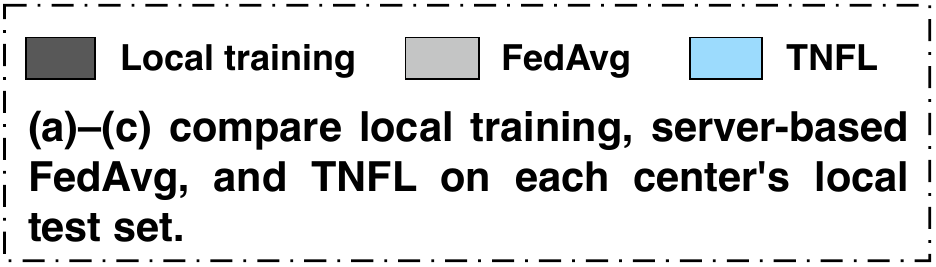}
        % \caption{Subfig c}
        % \label{subfig:c}
    \end{subfigure}

    \par
    % \vspace{4pt}
    \noindent\makebox[\linewidth]{\dotfill}
    % \vspace{2pt}

    % ========== 第三排：左侧文字块(0.48) + 右侧子图d(0.48) ==========
    \begin{subfigure}[t]{0.48\textwidth}
        \centering
        \includegraphics[width=\linewidth, height=0.52\linewidth]{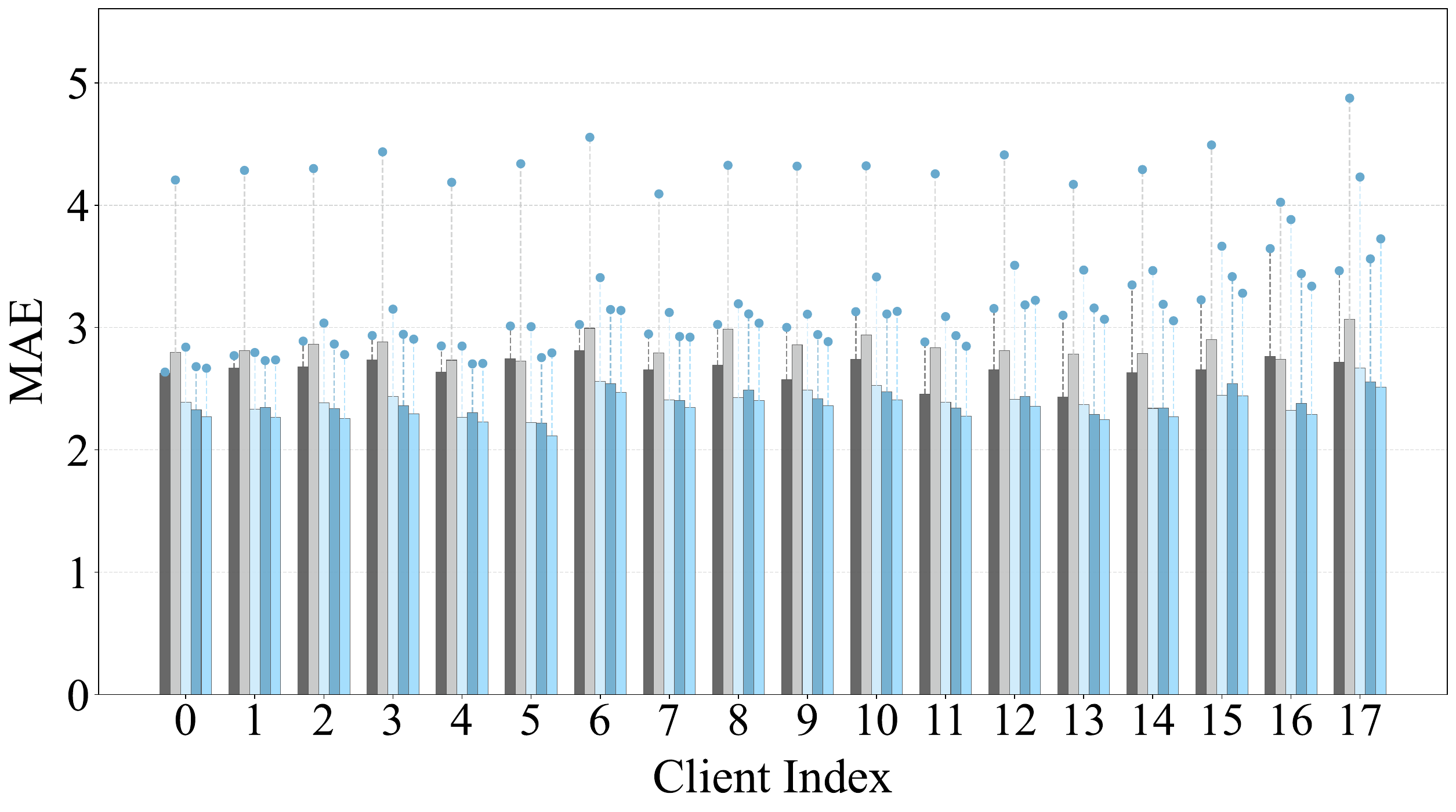}
        \subcaption{UKB-Center}
        % \label{subfig:e}
    \end{subfigure}
    \hfill
    \begin{subfigure}[t]{0.48\textwidth}
        \centering
        \includegraphics[width=\linewidth, height=0.52\linewidth]{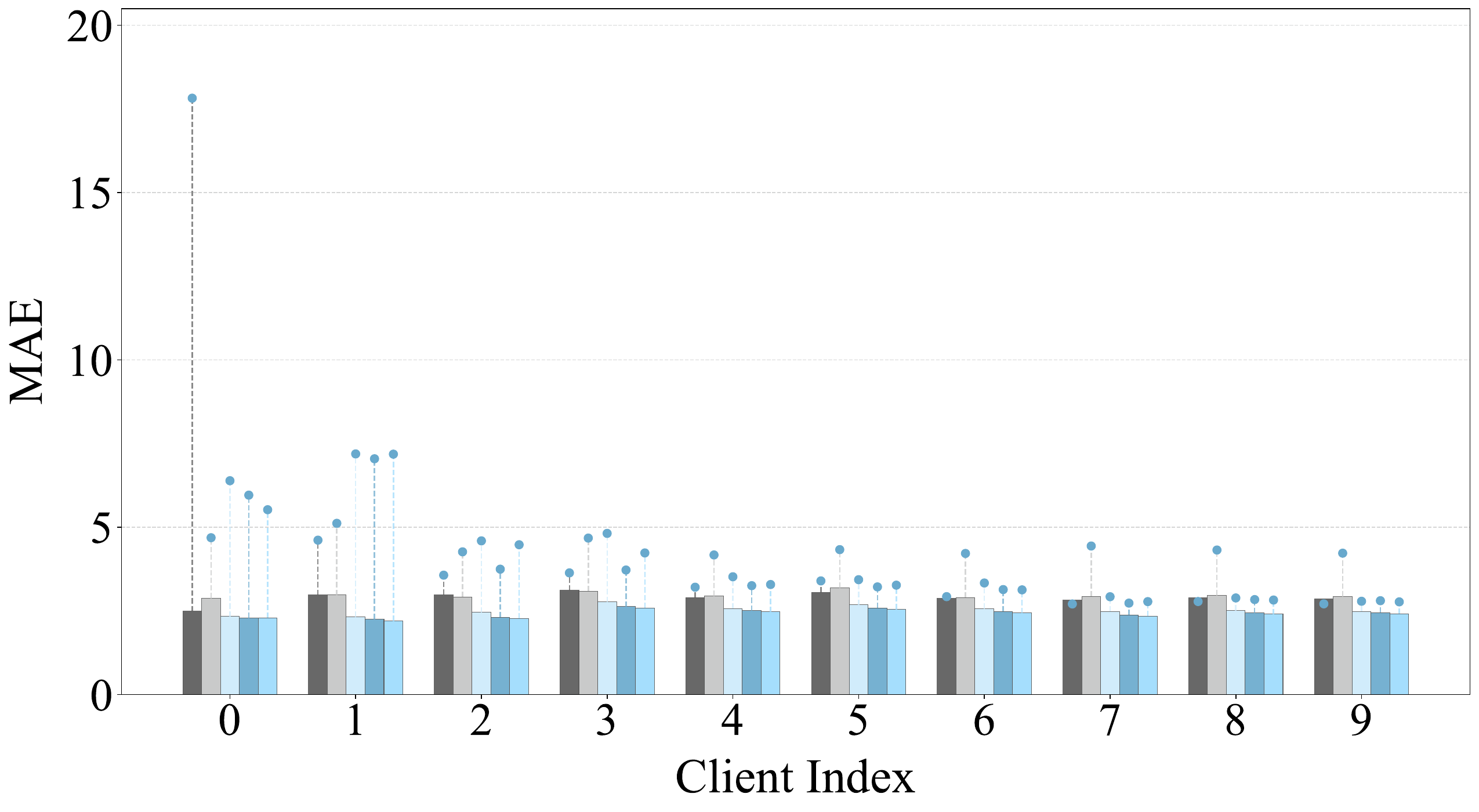}
        \subcaption{UKB-Imbalance}
        % \label{subfig:f}
    \end{subfigure}

    % \vspace{0.2em}

    % ========== 第四排：子图e(0.48) + 子图f(0.48) ==========
    \begin{subfigure}[t]{0.48\textwidth}
        \centering
        \includegraphics[width=\linewidth, height=0.52\linewidth, valign=t]{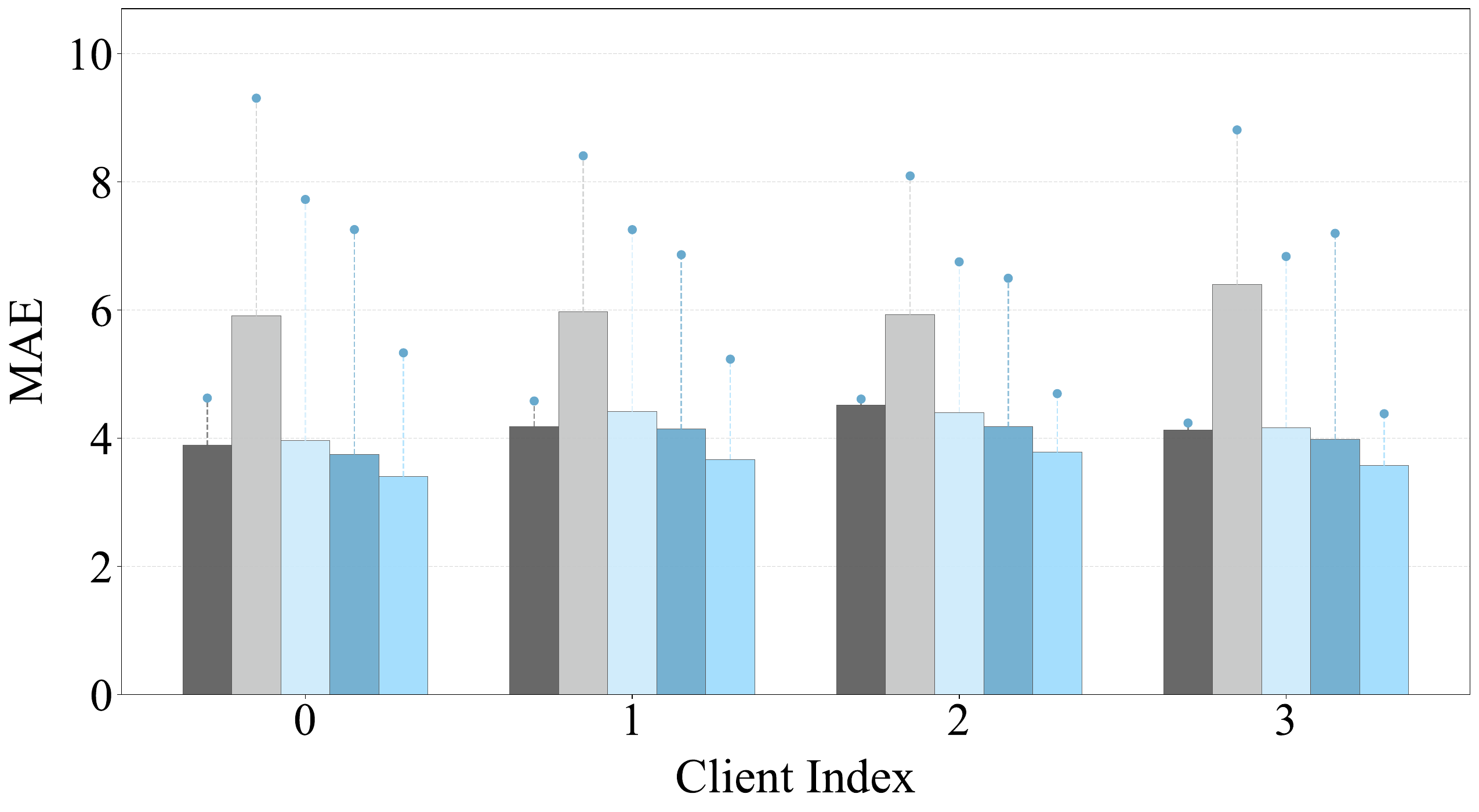}
        \subcaption{GEO-Methylation}
        % \label{subfig:e}
    \end{subfigure}
    \hfill
    \begin{subfigure}[t]{0.46\textwidth}
        \centering
        \includegraphics[width=\linewidth, valign=t]{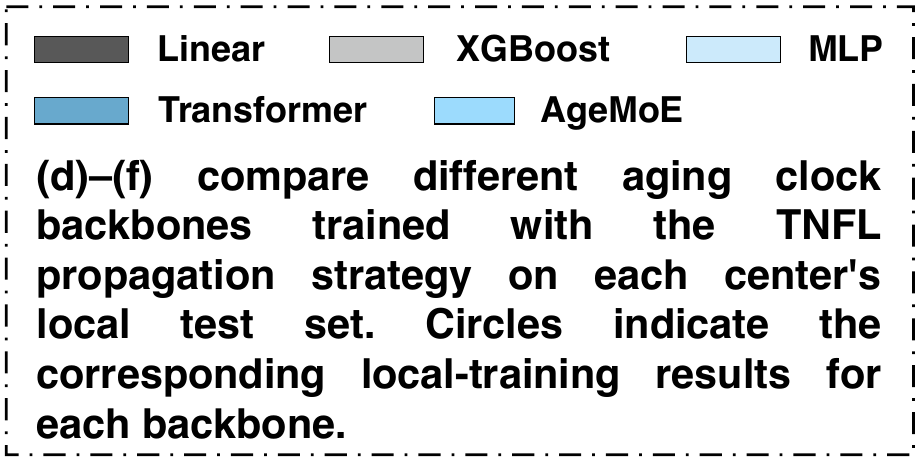}
        % \caption{Subfig f}
        % \label{subfig:f}
    \end{subfigure}

    \par
    % \vspace{4pt}
    \noindent\makebox[\linewidth]{\dotfill}
    % \vspace{4pt}

    % ========== 第五排：左侧文字块(0.48) + 右侧子图g(0.48) ==========
    \begin{subfigure}[t]{0.48\textwidth}
        \centering
        \includegraphics[width=\linewidth]{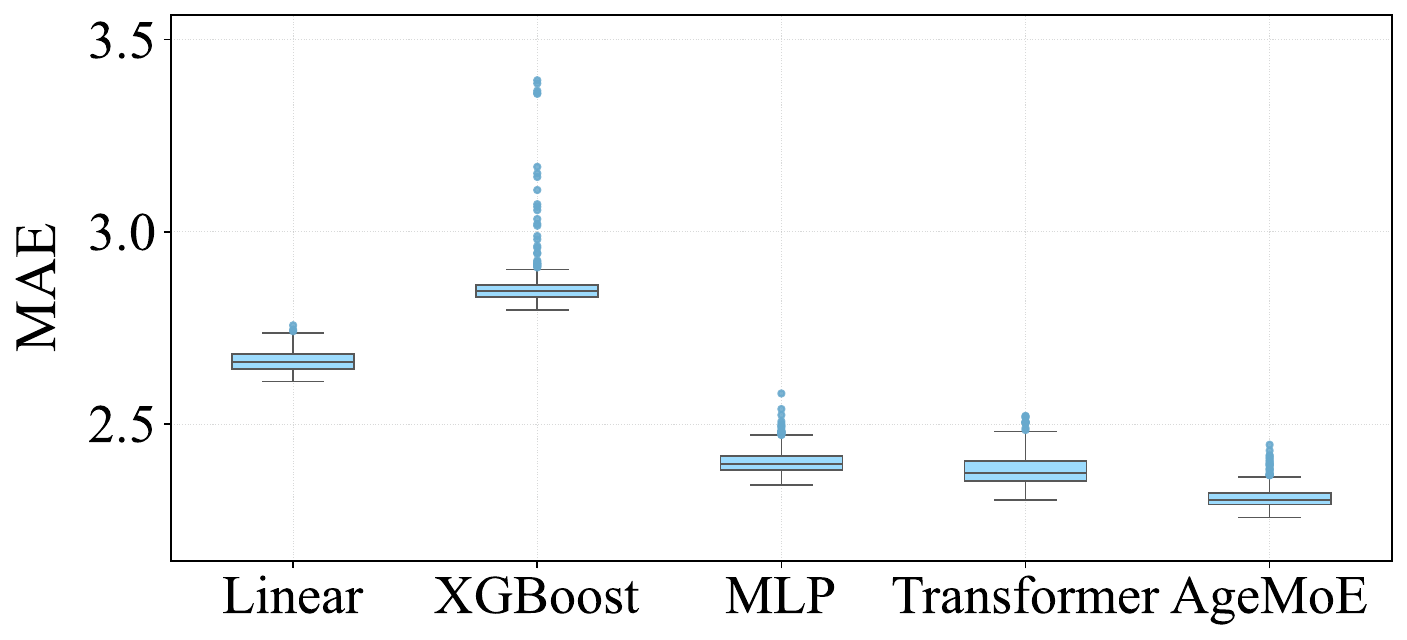}
        \subcaption{UKB-Center}
        % \label{subfig:h}
    \end{subfigure}
    \hfill
    \begin{subfigure}[t]{0.48\textwidth}
        \centering
        \includegraphics[width=\linewidth]{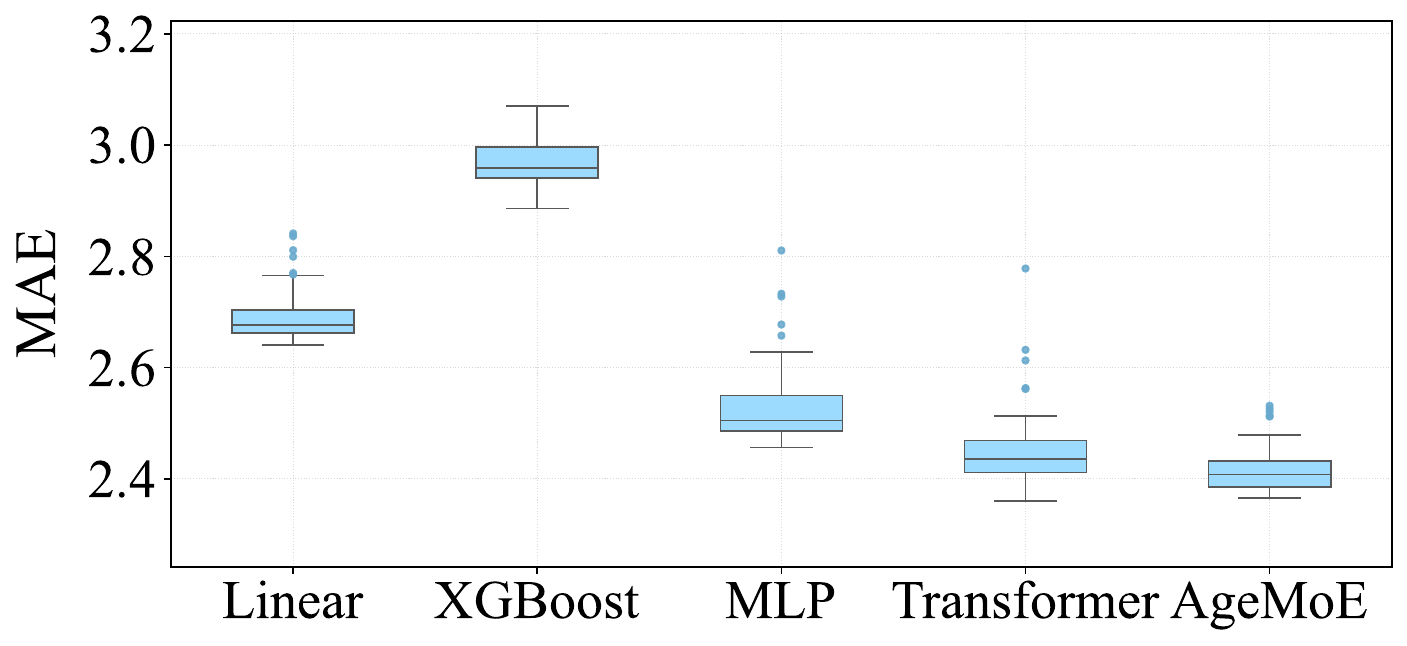}
        \subcaption{UKB-Imbalance}
        % \label{subfig:i}
    \end{subfigure}

    % \vspace{0.2em}

    % ========== 第六排：子图h(0.48) + 子图i(0.48) ==========
    \begin{subfigure}[t]{0.48\textwidth}
        \centering
        \includegraphics[width=\linewidth, valign=t]{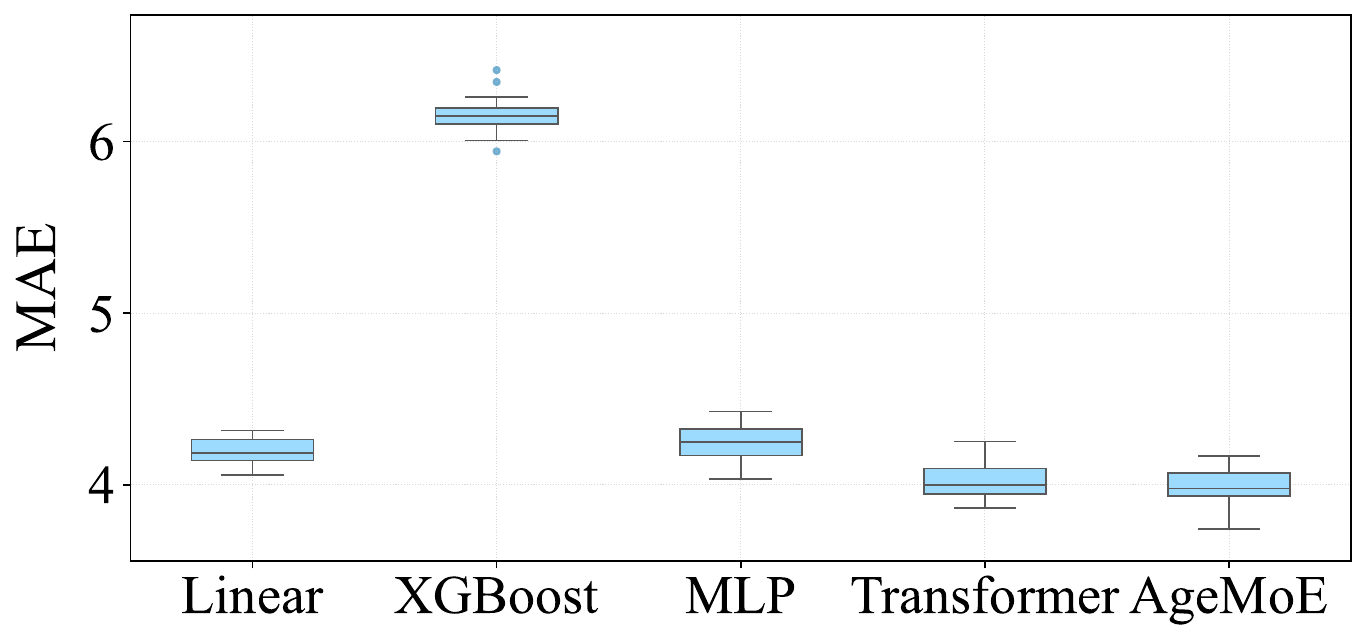}
        \subcaption{GEO-Methylation}
        % \label{subfig:h}
    \end{subfigure}
    \hfill
    \begin{subfigure}[t]{0.46\textwidth}
        \centering
        \includegraphics[width=\linewidth, valign=t]{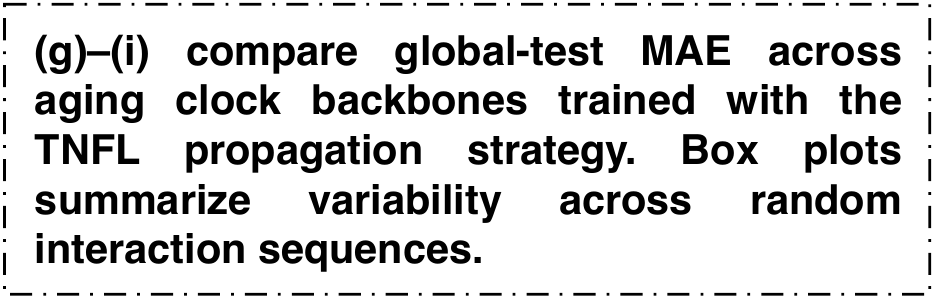}
        % \caption{Subfig i}
        % \label{subfig:i}
    \end{subfigure}

    \caption{Predictive performance of \model and its compatibility with different aging-clock backbones across multi-center datasets. Panels (a)–(c) compare \model with local training and server-based FedAvg, whereas panels (d)–(i) evaluate different aging-clock backbones under the \model propagation strategy. AgeMoE denotes the main implementation with generative replay.}
    \label{fig:mae_comparison_all}
\end{figure}

\subsection{Computational Evaluation of \model}
We next evaluate \model from the computational perspective, focusing on predictive performance and robustness under multi-center learning. Predictive performance is assessed across limited-data centers and different aging-clock backbones, while robustness is examined across varying interaction orders, heterogeneous center conditions, and trust-network structures. Together, these evaluations characterize how effectively \model addresses the key computational challenges of multi-center aging-clock modeling.

\subsubsection{Predictive Performance}
We first evaluate whether \model improves aging-clock prediction when individual centers have limited local data, compared with independent local training and using server-based FedAvg~\cite{mcmahan2017communication} as a centralized federated learning reference. Performance is measured by mean absolute error (MAE) on each center's local test set~\footnote{We use the terms ``client'' and ``center'' interchangeably throughout this paper to denote participating institutions in the federated learning setting.}. Local training and FedAvg provide reference settings outside the trust-network protocol, whereas \model is evaluated through trust-network-based sequential propagation. For this comparison, we instantiate multiple valid single-pass trust sequences by randomly permuting client indices so that every center is visited once, with each consecutive model transmission governed by a directed trust relation. Results for \model are averaged across these sampled sequences to reduce dependence on any particular propagation order. Experiments are conducted on three multi-center datasets covering multiple omics modalities and client configurations: UKB-Center and UKB-Imbalance from the UK Biobank~\cite{sudlow2015uk}, and GEO-Methylation from the Gene Expression Omnibus (GEO)~\cite{edgar2002gene}. Results are summarized in Fig.~\ref{fig:mae_comparison_all}(a)--(c).

Across datasets, the results support three main findings. \textbf{First}, \model consistently achieves lower MAE than independent local training, demonstrating the value of integrating information from distributed centers. \textbf{Second}, \model achieves predictive performance comparable to server-based FedAvg without relying on central aggregation. On the UKB-Center dataset, the average MAE reduction relative to local training is 0.69 years for \model and 0.70 years for server-based FedAvg, showing that trust-network-based pairwise propagation can provide effective collaborative modeling. \textbf{Third}, \model further improves over FedAvg under more heterogeneous configurations, with additional average MAE reductions of 0.12 years on UKB-Imbalance and 0.34 years on GEO-Methylation, indicating that \model remains effective under greater cross-center heterogeneity. These results demonstrate that \model enables effective collaborative aging-clock prediction when individual centers are limited by local data, while maintaining performance competitive with centralized federated learning.

We further examine whether these predictive benefits extend across different aging-clock backbones. Specifically, we consider linear regression~\cite{hastie2009elements}, XGBoost~\cite{chen2016xgboost}, multilayer perceptrons (MLP)~\cite{rumelhart1986learning}, and transformers~\cite{vaswani2017attention} as alternative aging-clock backbones within the \model propagation strategy. The main \model implementation uses AgeMoE as the discriminative predictor together with generative replay for cross-center knowledge preservation. Performance is assessed on local test sets to capture center-specific predictive ability and on a global test set to evaluate generalization across centers. To account for variability across interaction orders, results are averaged over multiple randomly sampled center sequences.

As shown in Fig.~\ref{fig:mae_comparison_all}(d)--(f), all evaluated aging-clock backbones improve over their local-training counterparts when trained with the trust-network-constrained \model propagation strategy, indicating that \model is not tied to a specific aging-clock architecture. In addition, the main \model implementation consistently achieves the strongest overall performance and exhibits lower variance on global test sets across different interaction sequences (Fig.~\ref{fig:mae_comparison_all}(g)--(i)). It achieves median MAE reductions of 0.54 years on UKB-Center, 0.55 years on UKB-Imbalance, and 2.55 years on GEO-Methylation. Notably, XGBoost also benefits substantially from the \model propagation strategy, achieving a 2.60-year MAE reduction on GEO-Methylation relative to local training. These results show that the trust-network-based learning mechanism can accommodate diverse discriminative aging-clock backbones, while AgeMoE provides the main predictive implementation used for subsequent analysis.

To further examine the contribution of generative replay to cross-center predictive generalization, we perform an ablation study by removing the generative component. Removing the generative component reduces cross-client generalization, particularly under settings with more pronounced distributional heterogeneity, supporting its role in preserving previously acquired information during cross-center learning. Detailed results are provided in Appendix~\ref{app:ablation}. Together, these results show that \model maintains effective predictive performance under limited local data, across different aging-clock backbones, and with heterogeneous cross-center distributions.

\subsubsection{Robustness Evaluation}
We next evaluate the robustness of \model under variations in interaction order, center-specific measurement conditions, and trust-network structures. We first examine whether sequential cross-center learning remains stable across different interaction orders and preserves previously acquired knowledge. Following the sequence-level instantiation used in the baseline evaluation, each sampled interaction order corresponds to a valid single-pass trust sequence in which all centers are visited once and direct model transmission occurs only between consecutive centers. We evaluate different aging-clock backbones trained with the \model propagation strategy across multiple randomly sampled interaction orders for each dataset. Three complementary metrics are used to characterize order sensitivity, sequential performance degradation, and knowledge forgetting, with detailed definitions and calculation procedures provided in Appendix~\ref{app:robustness}.

\begin{figure}[htbp]
    \centering
    % 全局间距优化：减小列间距，保证布局紧凑
    \setlength{\tabcolsep}{0.5em}
    \renewcommand{\arraystretch}{1.0}

    % ========== 第一排：左侧文字块(0.48) + 右侧子图a(0.48) ==========
    \begin{subfigure}[t]{0.48\textwidth}  % [t] 顶端对齐
        \centering
        \includegraphics[width=\linewidth]{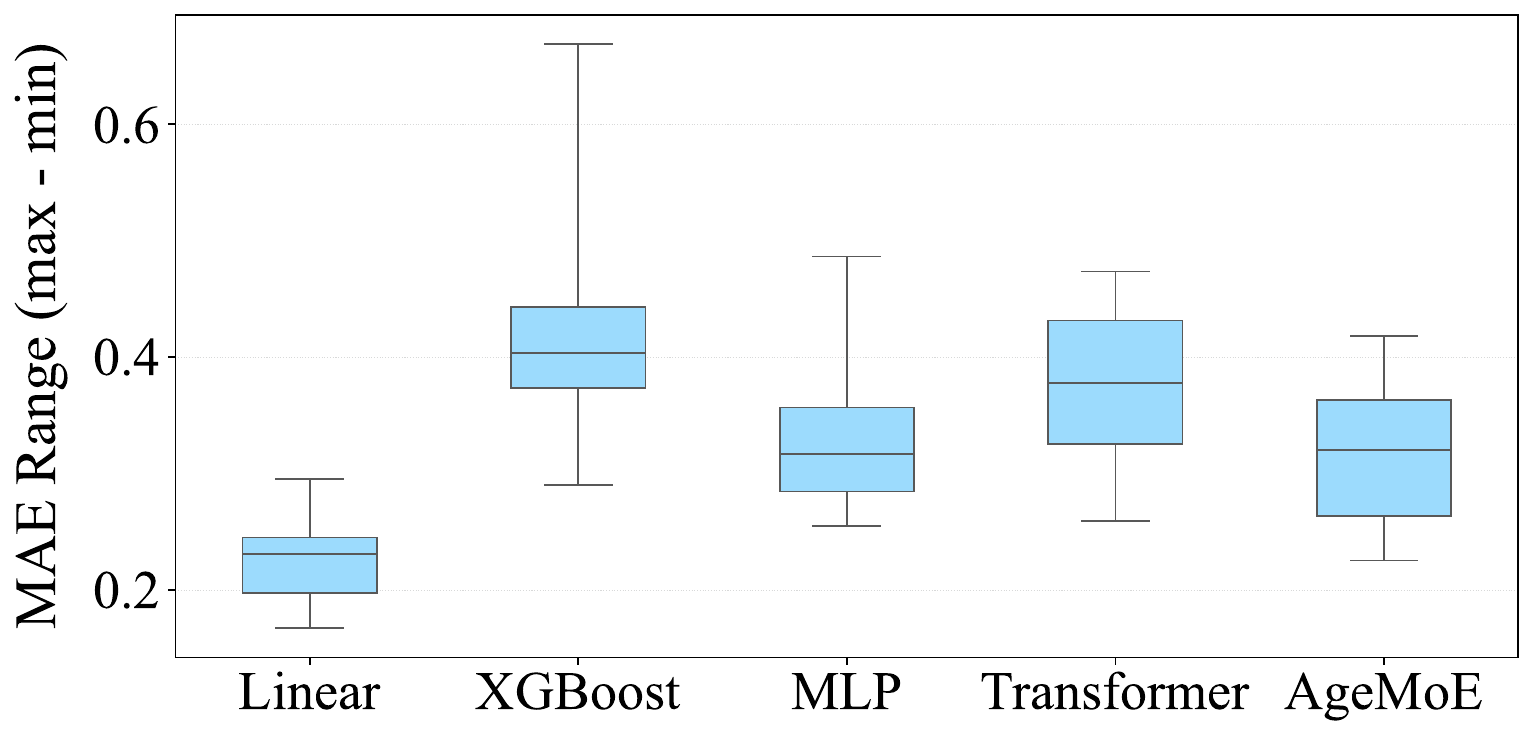}
        \subcaption{UKB-Center}
    \end{subfigure}
    \hfill  % 自动填充列间空白
    \begin{subfigure}[t]{0.48\textwidth}  % [t] 顶端对齐
        \centering
        \includegraphics[width=\linewidth]{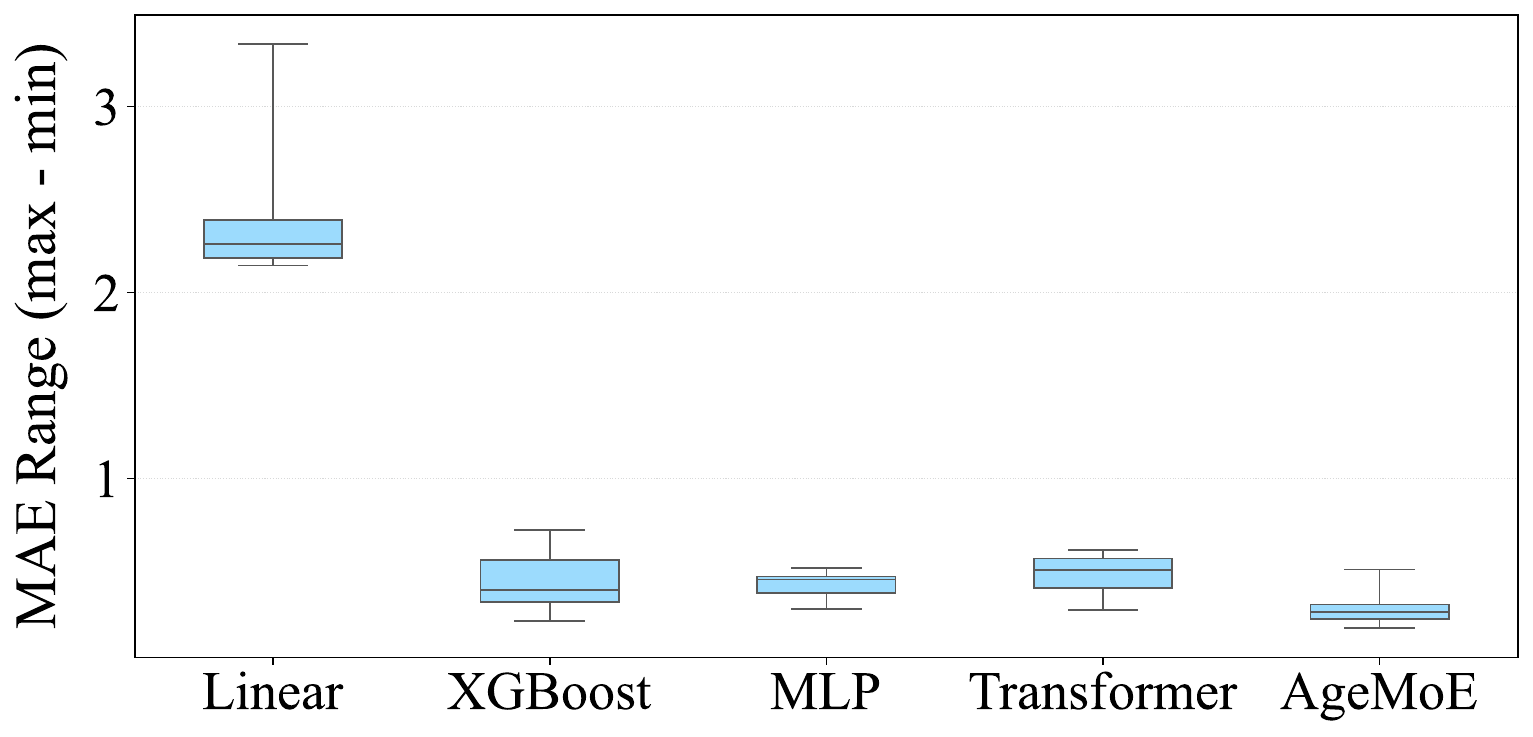}
        % 可选：添加子图标题（不需要则注释）
        \subcaption{UKB-Imbalance}
        % \label{subfig:2a}
    \end{subfigure}

    % \vspace{0.2em}  % 行间距，可按需调整

    % ========== 第二排：子图b(0.48) + 子图c(0.48) ==========
    \begin{subfigure}[t]{0.48\textwidth}
        \centering
        \includegraphics[width=\linewidth, valign=t]{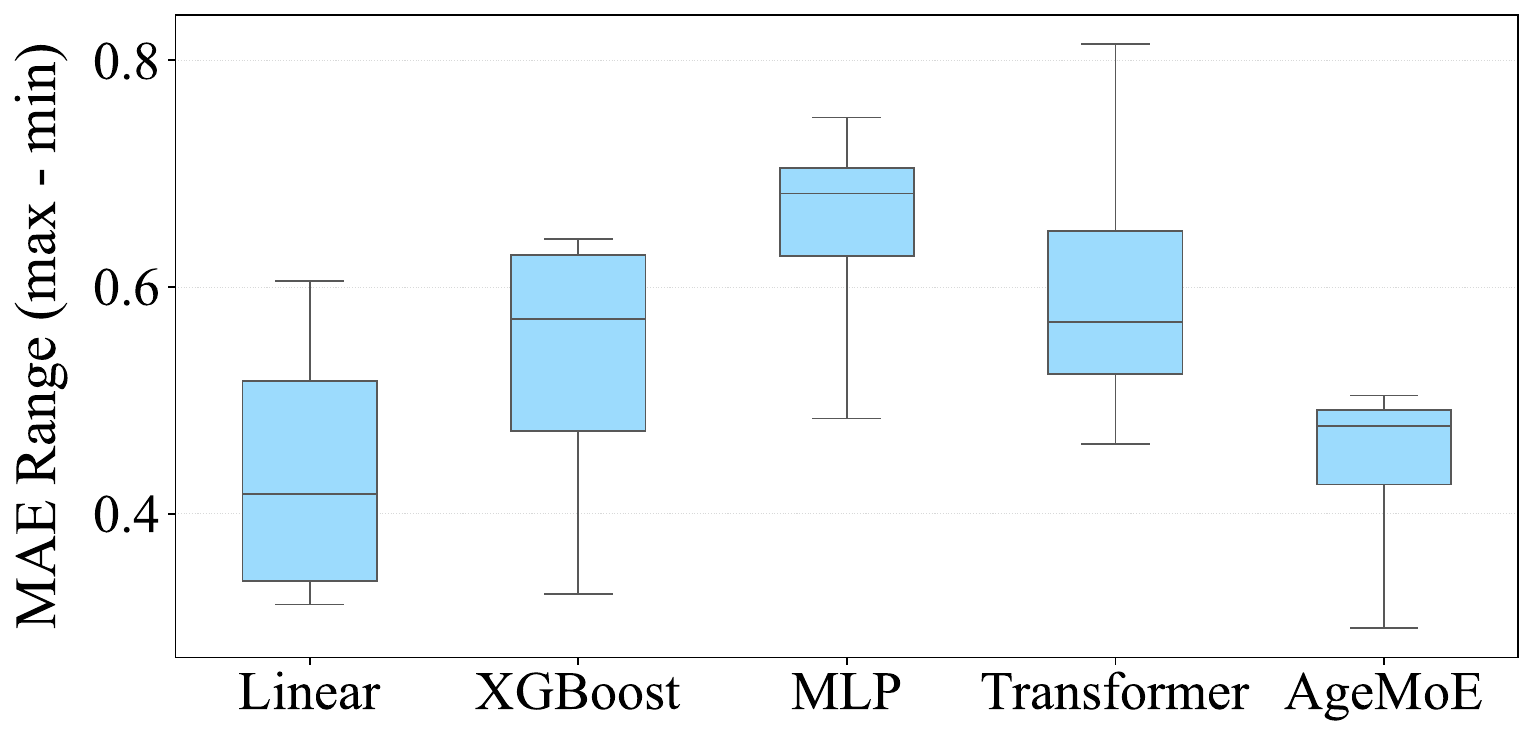}
        \subcaption{GEO-Methylation}
        % \label{subfig:b}
    \end{subfigure}
    \hfill
    \begin{subfigure}[t]{0.48\textwidth}
        \centering
        \includegraphics[width=\linewidth, valign=t]{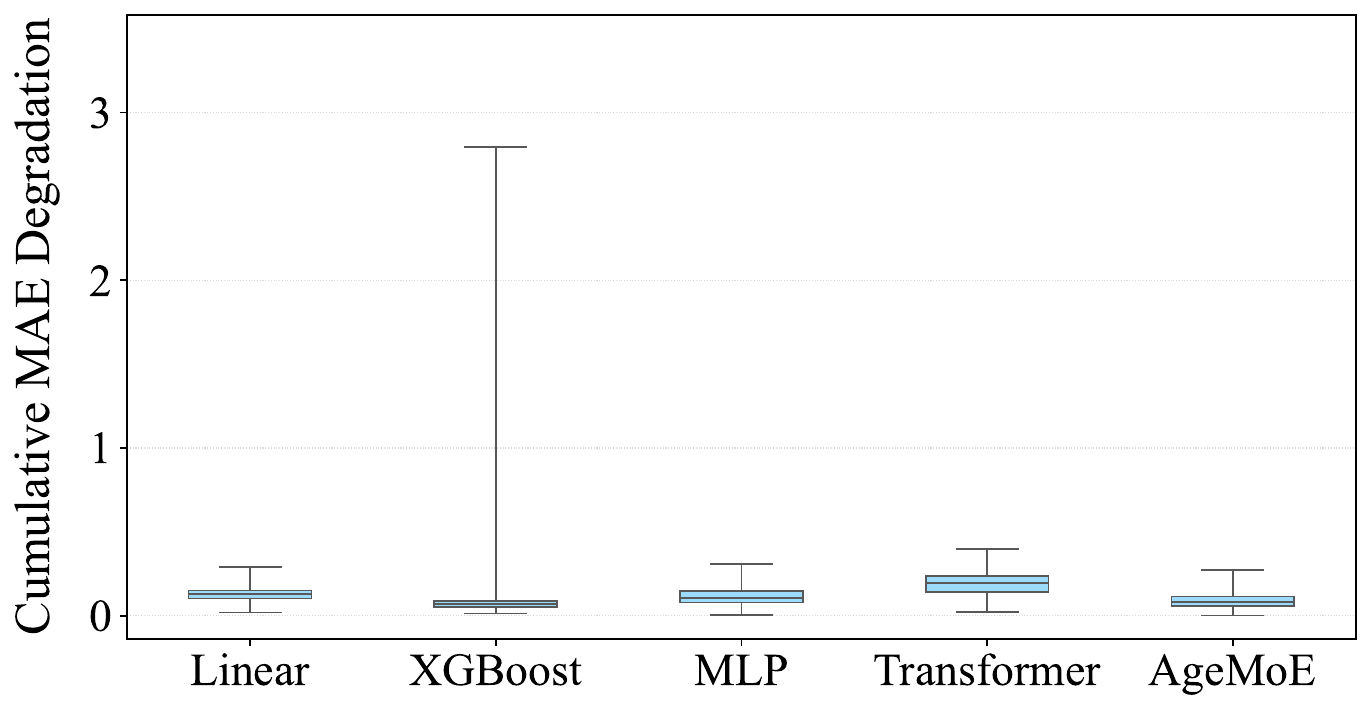}
        \subcaption{UKB-Center}
        % \label{subfig:c}
    \end{subfigure}

    % \par
    % \vspace{4pt}
    % \noindent\makebox[\linewidth]{\dotfill}
    % \vspace{4pt}

    % ========== 第三排：左侧文字块(0.48) + 右侧子图d(0.48) ==========
    \begin{subfigure}[t]{0.48\textwidth}
        \centering
        \includegraphics[width=\linewidth]{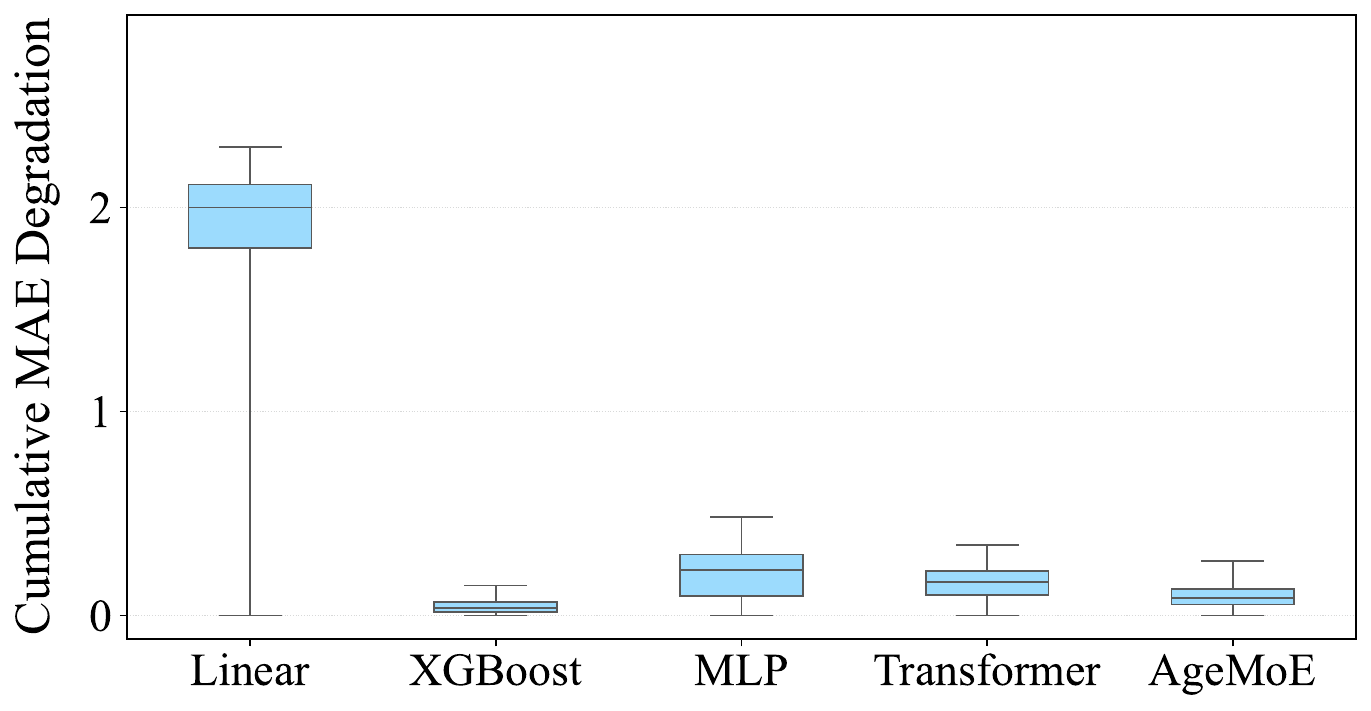}
        \subcaption{UKB-Imbalance}
        % \label{subfig:e}
    \end{subfigure}
    \hfill
    \begin{subfigure}[t]{0.48\textwidth}
        \centering
        \includegraphics[width=\linewidth]{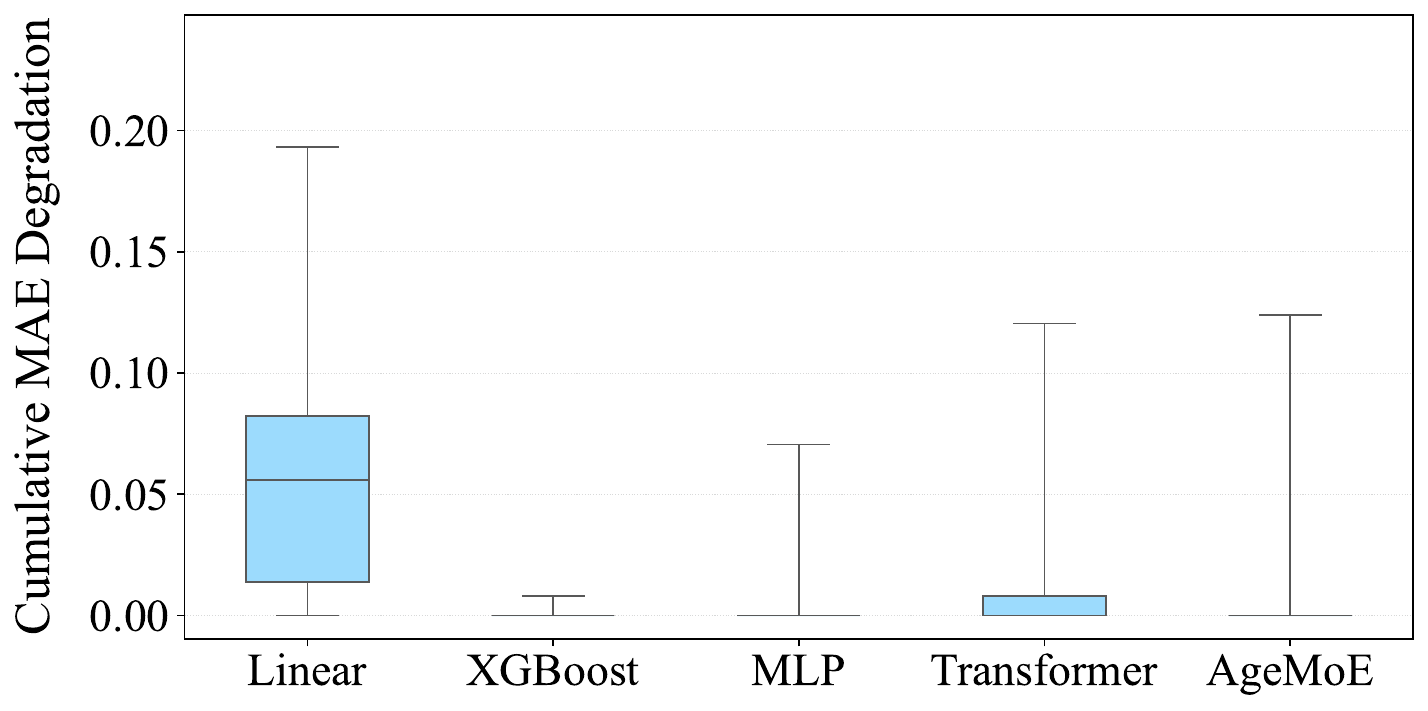}
        \subcaption{GEO-Methylation}
        % \label{subfig:f}
    \end{subfigure}

    % \vspace{0.2em}

    % ========== 第四排：子图e(0.48) + 子图f(0.48) ==========
    \begin{subfigure}[t]{0.48\textwidth}
        \centering
        \includegraphics[width=\linewidth, valign=t]{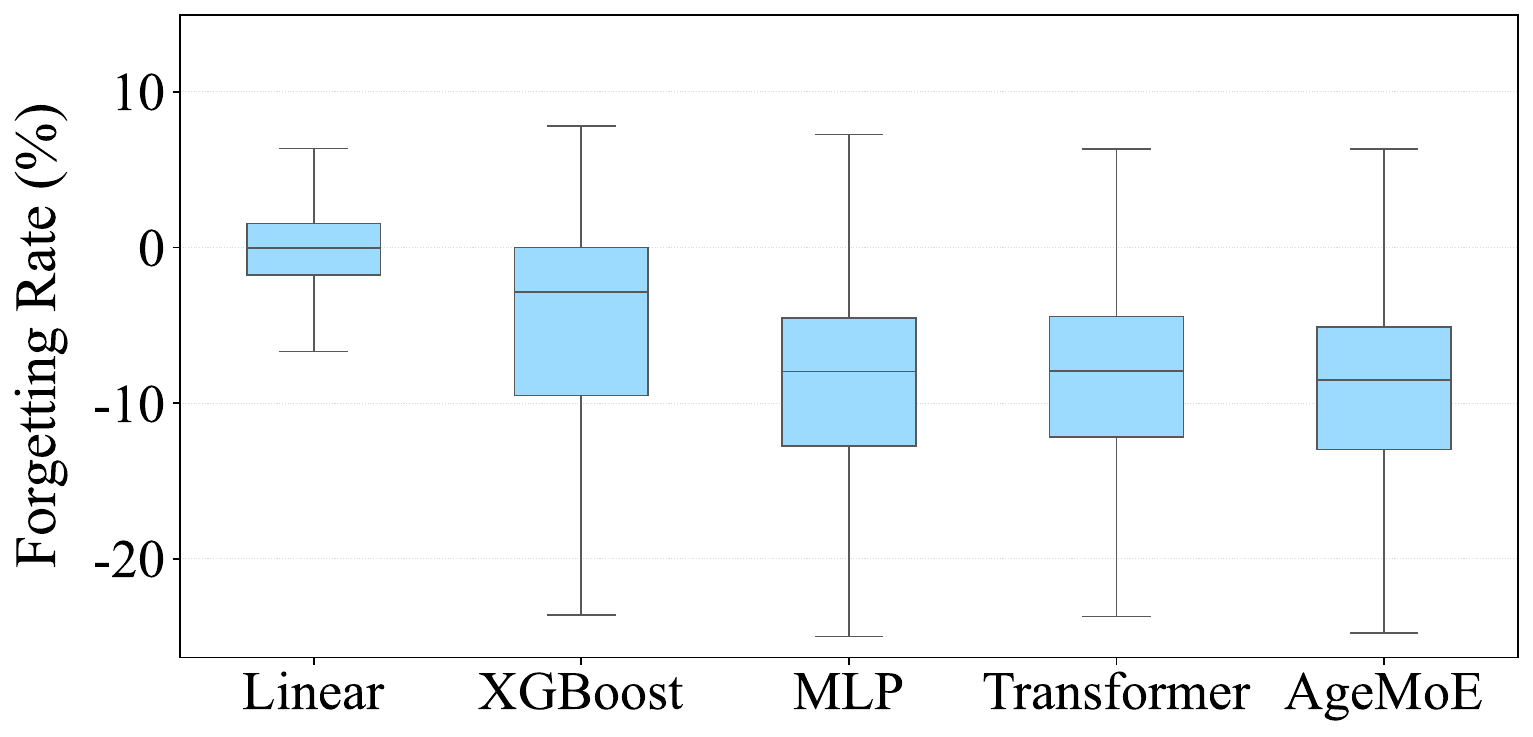}
        \subcaption{UKB-Center}
        % \label{subfig:e}
    \end{subfigure}
    \hfill
    \begin{subfigure}[t]{0.48\textwidth}
        \centering
        \includegraphics[width=\linewidth, valign=t]{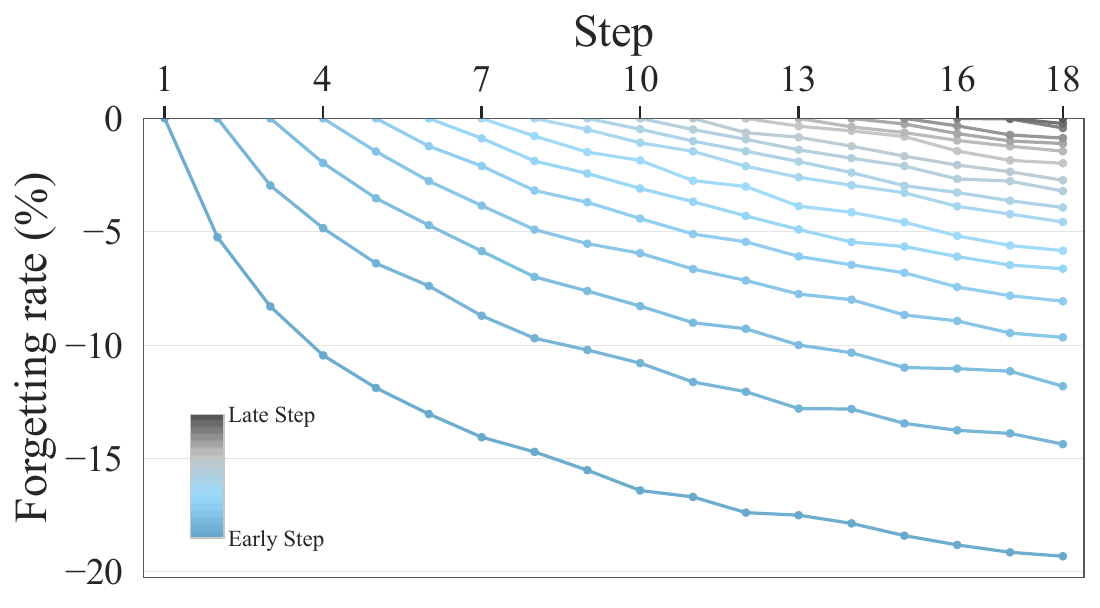}
        \subcaption{UKB-Center}
        % \label{subfig:f}
    \end{subfigure}

    \caption{Interaction-order robustness of aging clock models under \model propagation. (a)–(c) show the client-level MAE range, computed as the maximum minus minimum MAE across sampled interaction orders on local test sets; lower values indicate greater robustness to interaction-order changes. (d)–(f) show cumulative MAE degradation on the global test set, computed by accumulating only MAE increases between successive center updates; lower values indicate more stable performance progression. (g) shows forgetting rates on UKB-Center, computed as the relative MAE change from an intermediate model to the final model for clients in the first half of each sequence; positive values indicate performance degradation after subsequent updates, whereas zero or negative values indicate knowledge preservation or improvement. (h) further visualizes step-wise forgetting dynamics on UKB-Center across all sampled interaction sequences, where each point summarizes the median forgetting rate for clients visited at a given step and evaluated after later updates; values below zero indicate that later updates do not systematically degrade earlier client performance.}
    \label{fig:robustness_comparison_all}
\end{figure}

\noindent \textbf{Client-level MAE range.}
We measure robustness for each client as the range of MAE values, computed as the maximum minus minimum MAE across interaction orders on the client's local test set. Smaller ranges indicate more stable center-specific performance under different propagation sequences. As shown in Fig.~\ref{fig:robustness_comparison_all}(a)--(c), the AgeMoE-based main implementation equipped with generative replay exhibits stable performance, with ranges comparable to or smaller than those of alternative aging-clock backbones. The ranges remain below 0.5 years across datasets and drop below 0.3 years on UKB-Imbalance, indicating limited sensitivity of center-specific prediction to interaction order.

\noindent \textbf{Cumulative MAE degradation.}
We further assess whether global predictive performance deteriorates during sequential propagation. For each interaction order, this metric is calculated by summing only the positive increases in global-test MAE between consecutive intermediate model states, i.e., $\max(\text{current MAE} - \text{previous MAE}, 0)$, accumulated over the propagation process. This metric captures the extent of performance degradation introduced by successive center updates; lower values indicate more stable sequential updates. As shown in Fig.~\ref{fig:robustness_comparison_all}(d)--(f), the main \model implementation exhibits minimal cumulative degradation, with values below 0.2 years on the two UK Biobank datasets and near zero on GEO-Methylation. These results indicate that successive center updates do not lead to substantial cumulative deterioration in global predictive performance.

\noindent \textbf{Forgetting rate.}
We define forgetting for a client visited at a given step as the relative change in MAE from the intermediate model obtained after that client update to the final model produced at the end of the same interaction order. Specifically, it is calculated as $(\text{final MAE} - \text{intermediate MAE}) / \text{intermediate MAE} \times 100\%$, where both MAE values are evaluated on the same client's local test set. Positive values therefore indicate that performance on an earlier-visited client becomes worse after subsequent center updates, suggesting forgetting of previously learned center-specific information. To focus on early-stage effects, this metric is computed on the first half of clients in each sequence, where such information is most likely to be overwritten by later updates. We use UKB-Center for this analysis because its larger number of clients and diverse sample characteristics provide a suitable setting to observe potential forgetting effects. Results show that the main \model implementation consistently maintains or improves performance on earlier-visited clients after subsequent center updates (Fig.~\ref{fig:robustness_comparison_all}(g)).

We further analyze step-wise forgetting dynamics on UKB-Center, where each step denotes the position of a visited center within an interaction sequence and the MAE immediately after that center update serves as the reference for later evaluations. For each visited step and later evaluation step, Fig.~\ref{fig:robustness_comparison_all}(h) reports the median forgetting rate across all sampled interaction sequences, rather than a single sequence-specific trajectory. The median trajectories remain mostly below zero, indicating that later center updates do not systematically degrade performance on previously visited centers and often further improve it.

These results show that \model remains stable across different interaction orders and largely preserves information acquired from previously visited centers. We next examine whether this robustness extends to variation in center-specific measurements.

\noindent \textbf{Robustness to simulated measurement variability.}
To evaluate how \model handles practical variations in data collection, we simulate measurement differences across clients, including factors such as instrument calibration, experimental procedures, and batch effects. Each feature is perturbed in two ways: a client-specific shift, representing systematic deviations shared by all samples within a center, and sample-level random noise, capturing local measurement fluctuations. Despite these perturbations, the main \model implementation shows only limited performance degradation, indicating that predictive performance remains stable under simulated center-specific measurement variability. Full results of this perturbation analysis are provided in Appendix~\ref{app:perturb}.

\noindent \textbf{Robustness across trust-network structures.}
Finally, we evaluate whether \model remains effective when collaboration is constrained by more complex directed trust-network structures rather than single-pass interaction sequences. We simulate model propagation over network-based trust structures, where centers may participate multiple times and propagation sequences are not restricted to a single client permutation. Under these simulated conditions, \model maintains predictive performance comparable to that observed with randomly sampled interaction orders, and the relative performance ranking among aging-clock backbones remains largely consistent. These results indicate that the trust-network-based propagation mechanism remains effective across different feasible collaboration structures and more complex multi-center interaction patterns. Detailed experimental results are provided in Appendix~\ref{app:network}.

Together, these analyses show that \model maintains robust behavior across variations in interaction order, center-specific measurement conditions, and feasible trust-network structures.

\section{Discussion}\label{sec3}
\noindent\textbf{Summary of Main Findings.}
The biological analyses show that \model captures aging-related molecular patterns across multiple interaction orders. Age-dependent expert routing identifies individual proteins associated with aging-clock prediction, and these proteins are enriched in aging-related biological processes and supported by external biological annotations. Extending beyond individual effects, the model further identifies biologically coherent pairwise protein interactions that repeatedly organize into higher-order subnetworks spanning neural, vascular, endocrine, metabolic, inflammatory, and stress-related systems. From the computational perspective, \model achieves effective aging-clock prediction when local data are limited, remains compatible with different predictive backbones, and benefits from generative replay for cross-center generalization. It also exhibits stable behavior across different interaction orders, preserves information acquired from previously visited centers, and remains robust under simulated measurement variability and different trust-network structures. Together, these findings show that \model supports reliable multi-center aging-clock learning while enabling biological analysis of protein effects and interactions from individual to higher orders.

\noindent \textbf{Comparison with Existing Work.}
Standard federated learning provides the canonical paradigm for collaborative model training across distributed centers, typically relying on a central server to coordinate local updates and aggregate them into a shared global model~\cite{mcmahan2017communication}. Beyond this conventional setting, trust-related FL incorporates trust into collaborative learning to characterize client reliability or regulate communication among selected parties~\cite{zhang2020enabling,chen2023trustnetfl,he2019central}, with OPS~\cite{he2019central} further supporting collaboration over a directed trust graph. Decentralized FL removes the fixed central aggregator and instead coordinates learning through model exchange, aggregation, or synchronization among connected clients~\cite{warnat2021swarm,sun2022decentralized}. Sequential FL transfers models across centers so that information can accumulate through successive local updates~\cite{chang2018distributed,kampfederated,wang2024one}; CWT~\cite{chang2018distributed} and FedELMY~\cite{wang2024one} adopt cumulative cross-center training, while FedDC~\cite{kampfederated} transfers multiple models with periodic aggregation, and FedELMY further regularizes local updates relative to the preceding model to reduce drift. A detailed comparison of these paradigms with \model is summarized in Table~\ref{tab:related_work_comparison}. These approaches address different aspects of cross-center collaboration, but none is designed to jointly meet the requirements of multi-center aging-clock modeling considered here. In contrast, \model uses directed pairwise trust relations to determine feasible model propagation without relying on a globally trusted aggregation server. Along this propagation process, cumulative cross-center learning allows data-limited centers to benefit from previously accumulated information, while generative replay helps preserve acquired knowledge across heterogeneous center updates.

\begin{table*}[!t]
\centering
\caption{Comparison of representative federated learning methods for four key challenges in multi-center aging-clock modeling.
\Checkmark indicates that the corresponding challenge is explicitly addressed by a dedicated mechanism, whereas \ding{55} indicates that no such mechanism is provided.}
\label{tab:related_work_comparison}

\small
\setlength{\tabcolsep}{3.2pt}
\renewcommand{\arraystretch}{1.22}

\begin{tabular}{@{}
>{\centering\arraybackslash}p{2.55cm}
>{\centering\arraybackslash}p{2.35cm}
>{\centering\arraybackslash}p{1.75cm}
>{\centering\arraybackslash}p{2.00cm}
>{\centering\arraybackslash}p{2.05cm}
>{\centering\arraybackslash}p{2.20cm}
@{}}
\hline

\textbf{Category} &
\raisebox{0.18\baselineskip}{
    \makecell[c]{\textbf{Representative}\\\textbf{Method}}
} &
\parbox[c][1.05cm][c]{1.75cm}{
    \centering\footnotesize\bfseries
    Limited Local\\Data
} &
\parbox[c][1.05cm][c]{2.00cm}{
    \centering\footnotesize\bfseries
    Partial Asymmetric\\Trust
} &
\parbox[c][1.05cm][c]{2.05cm}{
    \centering\footnotesize\bfseries
    Prediction and\\Interpretation
} &
\parbox[c][1.05cm][c]{2.20cm}{
    \centering\footnotesize\bfseries
    Model Drift and\\Knowledge Forgetting
}
\\ \hline

\multirow[c]{3}{*}{\textbf{Trust-related FL}}
& TrustFL~\cite{zhang2020enabling}
& \ding{55} & \ding{55} & \ding{55} & \ding{55}\\

& TrustNetFL~\cite{chen2023trustnetfl}
& \ding{55} & \ding{55} & \ding{55} & \ding{55}\\

& OPS~\cite{he2019central}
& \ding{55} & \Checkmark & \ding{55} & \ding{55}\\
\hline

\multirow{2}{*}{\raisebox{-1.80\baselineskip}{\textbf{Decentralized FL}}}
& Swarm Learning~\cite{warnat2021swarm}
& \raisebox{-0.35\baselineskip}{\Checkmark}
& \raisebox{-0.35\baselineskip}{\ding{55}}
& \raisebox{-0.35\baselineskip}{\ding{55}}
& \raisebox{-0.35\baselineskip}{\ding{55}}\\

& Decentralized FedAvg~\cite{sun2022decentralized}
& \raisebox{-0.35\baselineskip}{\ding{55}}
& \raisebox{-0.35\baselineskip}{\ding{55}}
& \raisebox{-0.35\baselineskip}{\ding{55}}
& \raisebox{-0.35\baselineskip}{\ding{55}}\\
\hline

\multirow[c]{3}{*}{\textbf{Sequential FL}}
& CWT~\cite{chang2018distributed}
& \Checkmark & \ding{55} & \ding{55} & \ding{55}\\

& FedDC~\cite{kampfederated}
& \Checkmark & \ding{55} & \ding{55} & \ding{55}\\

& FedELMY~\cite{wang2024one}
& \Checkmark & \ding{55} & \ding{55} & \Checkmark\\
\hline

\textbf{Ours}
& \textbf{\model}
& \Checkmark & \Checkmark & \Checkmark & \Checkmark\\
\hline

\end{tabular}
\end{table*}

Machine-learning and deep-learning studies on aging clocks have also increasingly moved beyond age prediction toward biological interpretation~\cite{galkin2021deepmage,de2022pan}. Some studies identify molecular features that contribute most strongly to age prediction and examine their biological relevance, for example by analyzing selected CpG sites or their associations with aging-related phenotypes~\cite{vijayakumar2022pan}. Other approaches further relate model-identified molecular features to genes, biological processes, and pathways~\cite{martinez2023ncae,prosz2024biologically}, while more recent work has examined how these aging-related molecular patterns vary across age stages or population subgroups~\cite{lin2026deepstrataage}. These studies have substantially improved the interpretability of aging-clock models, but they mainly focus on which molecular features are important and what biological functions or patterns they represent. Our analysis further examines aging-related molecular organization across different orders, covering individual protein effects as well as pairwise and higher-order protein relationships. This allows us to characterize not only which proteins are relevant to aging-clock prediction, but also how their relationships extend beyond isolated pairwise associations into higher-order organization.

Taken together, \model addresses the key challenges of multi-center aging-clock modeling and enables a unified investigation of which proteins are relevant to aging-clock prediction and how their effects and interactions are organized across different orders.

\noindent \textbf{Limitations and Future Directions in Aging Clock Modeling.}
The proposed trust-network-based framework achieves effective performance across multiple molecular aging-clock datasets, but several limitations remain. From an algorithmic perspective, the current framework assumes that the trust network is specified before collaborative learning and that model propagation proceeds along feasible trust sequences derived from this network. Future work could investigate adaptive routing under dynamic trust relations, where subsequent propagation decisions are adjusted according to the evolving learning state or collaboration context. It would also be valuable to develop model- or data-dependent termination criteria that determine when additional cross-center propagation provides limited incremental benefit, allowing the sequential learning process to terminate adaptively rather than relying on a predefined propagation process. Second, using chronological age as the supervision signal may favor features with strong empirical associations with age, which may limit the diversity of captured aging-related signals. Future work should therefore consider alternative prediction targets beyond chronological age, such as healthspan or phenotypic age, to better reflect functional aspects of biological aging. Third, the current analysis considers each omics modality separately, which captures only part of the molecular complexity underlying aging. Integrating multi-omics data in future work may provide a more comprehensive representation of molecular changes across biological layers and reduce dependence on any single data source. Finally, future work could extend \model to real-world multi-center deployments to further evaluate its behavior under naturally occurring data heterogeneity, institutional governance policies, and communication constraints.

\section{Methods}\label{sec4}
This section describes \model for multi-center aging-clock modeling. We first formulate the multi-center problem setting and the computational requirements that motivate the framework design. Based on these requirements, \model is developed around two complementary aspects: how collaborative learning is organized under constrained inter-center trust, and how an aging-clock model is progressively learned while retaining predictive utility, interpretability, and previously acquired information across centers. We then formalize the trust-network-based learning framework and its model components, followed by the training and optimization procedures. Finally, we characterize the proposed trust relation relative to participant-level behavioral models and establish the security properties arising along valid trust sequences. The section concludes with the experimental design and evaluation settings used throughout the study.

\subsection{Problem Formulation and Multi-Center Challenges}
\noindent \textbf{Problem Definition.}
We formulate the aging clock task as a supervised regression problem, where the goal is to predict an individual’s chronological age from high-dimensional molecular measurements, such as proteomic or DNA methylation features. Formally, let $\mathbf{x} \in \mathbb{R}^d$ denote the molecular feature vector of a given sample, and let $y \in \mathbb{R}$ represent the corresponding chronological age. The objective is to learn a mapping $\mathcal{F}_\Theta: \mathbb{R}^d \rightarrow \mathbb{R}$ from molecular features to chronological age by minimizing a predefined loss function, typically the mean squared error, over the training data.

\noindent \textbf{Multi-Center Setting.}
Let there be $K$ centers, each holding a local dataset $\mathcal{D}_k = {(\mathbf{x}_i, y_i)}_{i=1}^{n_k}$, where $\mathbf{x}_i \in \mathbb{R}^d$ denotes the molecular features of sample $i$ and $y_i \in \mathbb{R}$ its chronological age, and $n_k$ is the number of samples at center $k$. The datasets remain locally held without direct sharing of individual-level data, and may differ across centers in sample size and molecular feature distributions. Cross-center collaboration is conducted through model communication subject to the relations among participating centers.

\noindent \textbf{Challenges in Multi-Center Aging-Clock Modeling.}
The above setting gives rise to four coupled challenges. \textbf{Limited Local Data} can restrict reliable aging-clock training at individual centers and may also be insufficient for local generative modeling, creating a need to leverage information across centers. Such collaboration is further constrained by \textbf{Partial and Asymmetric Trust}, where model communication may be permitted only between selected and directional pairs of centers. Meanwhile, \textbf{Discriminative and Interpretable Prediction} requires the aging clock to retain age prediction while supporting analysis of the learned age-related patterns. Finally, cross-center heterogeneity gives rise to \textbf{Model Drift and Knowledge Forgetting}, as continued adaptation to center-specific distributions may shift the model away from previously learned patterns and weaken knowledge acquired from other centers.

These challenges motivate \model as a unified framework for progressive aging-clock learning over constrained inter-center relations. Rather than training each center independently or relying on a globally trusted coordinator, \model allows an evolving model to accumulate information through authorized cross-center propagation. Within this process, the aging clock remains a discriminative predictor whose internal routing behavior can be interpreted, while generative replay helps retain information acquired from previously visited centers as the model adapts to heterogeneous local distributions. Together, these considerations shape the overall design of \model.

\subsection{Directed Trust Relations and Trust-Network Formulation}
To formalize the constrained inter-center relations described above, we introduce a directed, pairwise, and message-specific trust relation for model communication in \model. Each relation specifies the obligations associated with an authorized transmission between two centers, and the collection of such relations forms the directed trust network used by \model.

\noindent \textbf{Protocol Message.}
Let $A$ and $B$ denote two participating centers. We use $X_{A\rightarrow B}$ to denote the protocol-authorized message that carries the model parameters transmitted from $A$ to $B$ in a pairwise interaction. The concrete parameter components included in $X_{A\rightarrow B}$ depend on the instantiated \model implementation and training procedure. The message does not include raw local data, individual-level molecular measurements, local gradients, explicit parameter-update vectors, or local training computations, all of which remain local to the corresponding center. Here, $X_{A\rightarrow B}$  refers specifically to a pairwise message transmitted during the training process; release of the final trained model is specified separately in the \model protocol.

\noindent\textbf{Definition 1 (Directed Pairwise Trust Relation).}
For a protocol-authorized transmission of message $X_{A\rightarrow B}$ from center $A$ to center $B$, 
$\operatorname{Trust}(A,B,X_{A\rightarrow B})$ denotes the directed, pairwise, and message-specific trust relation governing this transmission. This relation imposes the following obligations:

\begin{enumerate}
    \item \textbf{Sender-side authenticity.}
    Center $A$ agrees to transmit $X_{A\rightarrow B}$ to center $B$ and guarantees that it does not poison, forge, or maliciously tamper with $X_{A\rightarrow B}$.

    \item \textbf{Receiver-side confidentiality.}
    Center $B$ shall not use $X_{A\rightarrow B}$ received from center $A$ to infer private information contained in or represented by $X_{A\rightarrow B}$.
\end{enumerate}

\noindent\textbf{Remark 1.}
If $X_{A\rightarrow B}$ is derived from a model state influenced by previously participating centers, the receiver-side confidentiality obligation also applies to private information from those centers insofar as it is represented in $X_{A\rightarrow B}$.

\noindent\textbf{Remark 2.}
Center $A$ may optionally encrypt $X_{A\rightarrow B}$ such that downstream centers of $B$ cannot directly access $X_{A\rightarrow B}$ or infer the private information contained in it.

The directed trust relation is non-transitive; specifically,
$\operatorname{Trust}(A,B,X_{A\rightarrow B})$ and
$\operatorname{Trust}(B,C,X_{B\rightarrow C})$ do not imply a direct trust
relation between $A$ and $C$.

\noindent\textbf{Trust Network.}
Let $\mathcal{V}=\{v_1,\ldots,v_K\}$ denote the set of participating centers. 
The directed trust network is represented as
\begin{equation}
    \mathcal{G}=(\mathcal{V},\mathcal{E}),
\end{equation}
where $(v_i,v_j)\in \mathcal{E}$ indicates that the protocol-authorized message
$X_{v_i\rightarrow v_j}$ can be transmitted from $v_i$ to $v_j$ under
$\operatorname{Trust}(v_i,v_j,X_{v_i\rightarrow v_j})$. The trust network is treated as a given collaboration structure determined by pre-established relationships among the participating centers. \model organizes collaborative training through feasible model-propagation sequences over this network, as described next.

% \red{A valid trust sequence is an ordered sequence of centers
% \[
% \mathcal{S}=[s_1,s_2,\ldots,s_T]
% \]
% such that, for every pair of consecutive centers,
% \[
% (s_t,s_{t+1})\in \mathcal{E},
% \qquad t=1,\ldots,T-1.
% \]
% Equivalently, each message transmission in the sequence satisfies
% \[
% \operatorname{Trust}
% \left(
% s_t,
% s_{t+1},
% X_{s_t\rightarrow s_{t+1}}
% \right),
% \qquad t=1,\ldots,T-1.
% \]
% The sequence $\mathcal{S}$ defines a feasible route for sequential model propagation
% over the trust network. Centers in $S$ are not required to be distinct.}

\begin{figure}[!t]
\centering
\includegraphics[width=\linewidth]{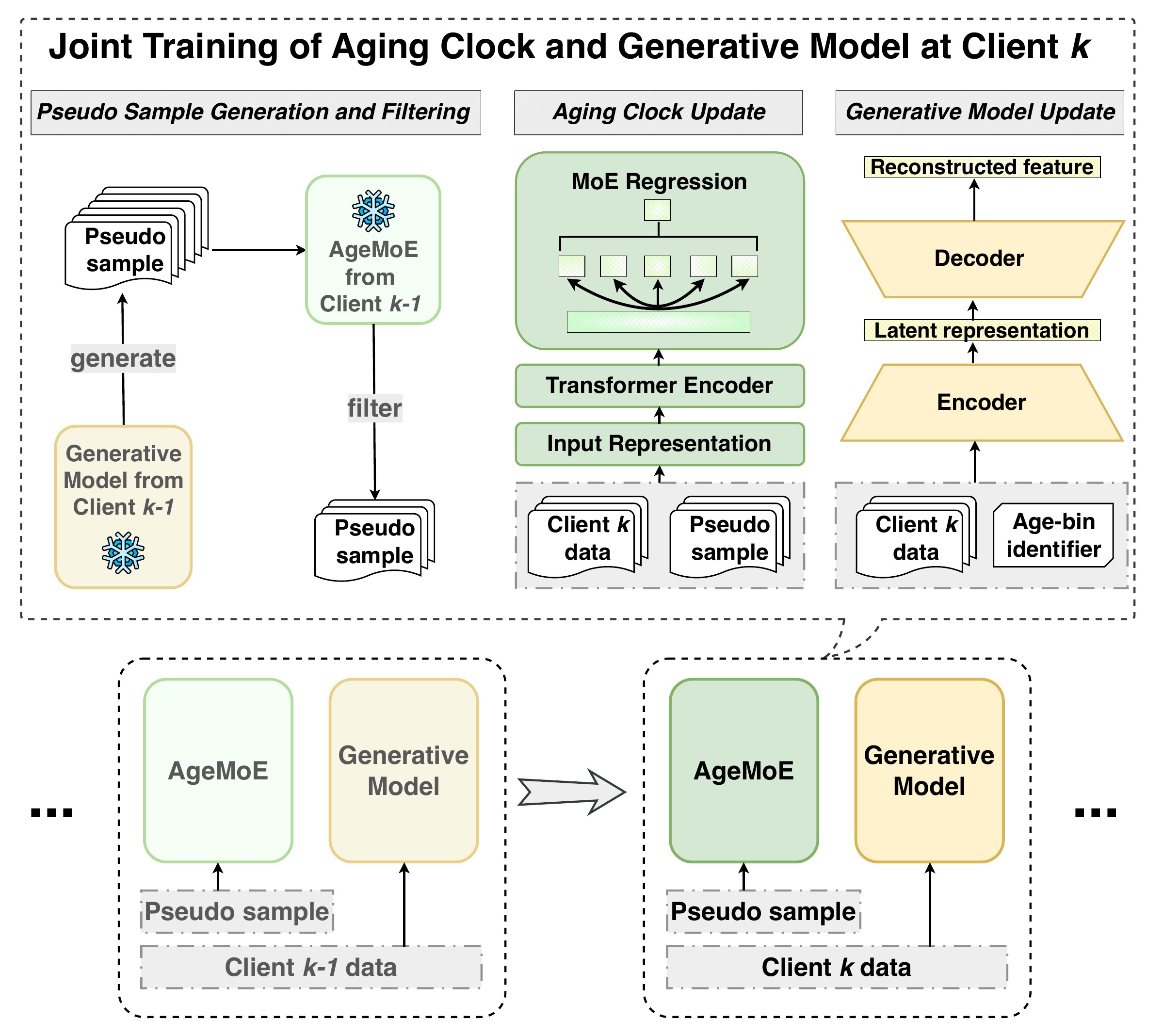}
\caption{Sequential training procedure of \model. The AgeMoE aging clock and generative model are jointly propagated across centers along a valid trust sequence composed of directed trust relations. At each client, the generative model first synthesizes pseudo-samples approximating previously encountered client distributions, followed by prediction-consistency filtering using the received aging clock. The filtered pseudo-samples are then combined with local data to update the AgeMoE aging clock, while the generative model is updated using local client data.}
\label{fig:sfl_framework}
\end{figure}
\subsection{Trust-Network-Based Federated Learning Framework}
\subsubsection{Framework Overview}
Building on the directed trust network defined above, \model organizes multi-center learning as a progressive model-propagation process over feasible trust-constrained sequences. Rather than training separate models at individual centers or relying on centralized aggregation, an evolving model state is successively transmitted and updated across centers under pre-established directed trust relations. Each participating center therefore continues from the knowledge accumulated through preceding updates and further adapts the model using its local data.

Within this propagation process, the main implementation of \model employs AgeMoE as the discriminative aging-clock predictor, enabling age estimation while exposing age-dependent routing behavior for interpretation. A generative replay mechanism is further maintained alongside the aging-clock model to retain information acquired from previously visited centers during continued adaptation to heterogeneous local distributions. The overall architecture of \model is illustrated in Fig.~\ref{fig:sfl_framework}.

\subsubsection{Trust-Constrained Training Sequence} 
Given the directed trust network $\mathcal{G}=(\mathcal{V},\mathcal{E})$, \model performs collaborative training along a finite sequence of centers
\begin{equation}
    S=[s_1,s_2,\ldots,s_T], \qquad s_t\in\mathcal{V},
\end{equation}
where each consecutive pair must correspond to a pre-established directed trust relation:
\begin{equation}
    (s_t,s_{t+1})\in \mathcal{E},\qquad t=1,\ldots,T-1.
\end{equation}
Equivalently, the protocol-authorized transmission from $s_t$ to $s_{t+1}$ is governed by
\begin{equation}
    \operatorname{Trust}(s_t,s_{t+1},X_{s_t\rightarrow s_{t+1}}).
\end{equation}
The sequence neither creates nor extends trust relations and it only composes pre-established directed trust relations into a feasible training route.

\subsubsection{Sequential Training Protocol}
Given a valid training sequence $S=[s_1,\ldots,s_T]$, \model maintains an evolving model state that is successively updated by the centers appearing in $S$. The first center $s_1$ starts from the initial model parameters $\Theta_0$ and performs local training on $\mathcal{D}_{s_1}$. For each subsequent step $t=2,\ldots,T$, center $s_t$ receives the protocol-authorized message $X_{s_{t-1}\rightarrow s_t}$ from the preceding center under the corresponding directed trust relation. Let $\Theta_{t-1}$ denote the model state available to center $s_t$, with $\Theta_0$ denoting the initial state for $s_1$. Each center then updates the received model using its local dataset:
\begin{equation}
\Theta_t =
\operatorname{Update}\left(
\Theta_{t-1},\mathcal{D}_{s_t}
\right),
\qquad t=1,\ldots,T,
\end{equation}
where $\operatorname{Update}(\cdot)$ denotes the local training process. Each center therefore continues training from the model state accumulated through preceding centers rather than training an independent model from scratch. For $t<T$, the updated parameters $\Theta_t$ are carried by the protocol message $X_{s_t\rightarrow s_{t+1}}$ and transmitted to $s_{t+1}$ under the corresponding directed trust relation. All transmissions follow the protocol-authorized message scope defined above, while local data and local optimization information remain at the corresponding center. In the main implementation, the auxiliary generative model is additionally propagated and updated together with the aging-clock model, as detailed in the subsequent training formulation.

% \noindent \textbf{Sequential Pairwise Training Protocol.}
% In the \model framework, the shared model evolves through a sequence of pairwise updates over the trust network. Let $\mathcal{S}=[s_1,...,s_T]$ denote a valid training sequence defined by the trust network, where $s_t$ is the center visited at step $t$. The first center $s_1$ initializes the training process with model parameters $\Theta_0$. At each subsequent step $t>1$, center $s_t$ receives the protocol-authorized message $X_{s_{t-1}\rightarrow s_t}$ from the preceding center under the corresponding directed trust relation. Let $\Theta_{t-1}$ denote the model parameters carried by this message. Center $s_t$ updates these parameters using its local dataset $\mathcal{D}_{s_t}$:
% \begin{equation}
%     \Theta_{t} = Update(\Theta_{{t-1}}, \mathcal{D}_{s_t}),
% \end{equation}
% where $Update(\cdot)$ represents the local training process. For $t<T$, after local training, the updated parameters $\Theta_t$ are carried by $X_{s_t\rightarrow s_{t+1}}$ and transmitted to the next center in the training sequence. This protocol allows each center to contribute to the evolving shared model using only its own local data. Only protocol-authorized model-parameter messages are transmitted between centers, while raw data, individual-level molecular measurements, local gradients, explicit parameter-update vectors, and local training computations remain local.

\subsubsection{Protocol Termination and Model Release}
\noindent \textbf{Protocol Termination.}
Given a predefined finite trust-constrained training sequence $S=[s_1,\ldots,s_T]$, the current \model protocol terminates after center $s_T$ completes the final local update, yielding the final aging-clock model $\Theta_T$. The stopping condition is therefore defined by completion of the prescribed sequence, which may include repeated visits to a center when permitted by the trust-network topology.

% Unlike one-shot sequential FL~\cite{wang2024one}, where each client communicates with its adjacent client only once, \model defines termination by completion of a predefined trust-constrained sequence, in which a center may appear multiple times when permitted by the trust-network topology.}

\noindent \textbf{Final Model Release.}
After termination, the final aging-clock model $\Theta_T$ is released to the participating centers according to a predefined protocol-level release policy, which can be represented as
\begin{equation}\label{eq:release}
    \operatorname{Release}(\Theta_T,\mathcal{V}).
\end{equation}
This release is treated as an output of the collaborative protocol rather than an additional pairwise training transmission. It therefore does not require the terminal center $s_T$ to establish a direct trust relation with every participating center, nor does it imply that $s_T$ is globally trusted. Accordingly, final model release is governed separately from the pairwise trust relations that constrain model propagation during training.

% The release operation may further be implemented through a protocol-managed anonymized distribution mechanism, such that participating centers receive $\Theta_T$ without learning which center served as the terminal node of the training sequence. Accordingly, final model release is governed separately from the pairwise trust relations that constrain model propagation during training.

% \blue{\subsubsection{Interaction Dependence and Stabilization}
% Under trust-network-constrained sequential propagation, the learned model can depend on the training sequence because each center adapts the received model to its local data distribution. When local distributions differ across centers, successive updates may shift the evolving model toward the distributions of recently visited centers, leading to model drift, while continued adaptation may also weaken information acquired from earlier centers and result in knowledge forgetting. Different valid training sequences can therefore produce different learning trajectories and predictive behavior. To mitigate these effects, the main implementation of \model incorporates a knowledge-preservation mechanism to stabilize sequential learning across heterogeneous centers, as detailed in the following sections.}

\subsection{Model Architecture}
\subsubsection{AgeMoE Aging Clock Model}
To model heterogeneous molecular patterns that arise across distributed client datasets, we implement an age-aware Mixture-of-Experts (AgeMoE) aging clock that maps tabular omics profiles to a scalar age prediction. The model combines a lightweight Transformer encoder with a soft MoE regression head, allowing multiple experts to specialize in different predictive patterns while a routing network dynamically determines their contributions for each sample. This design enables the model to capture diverse age-related signals across heterogeneous clients while maintaining a flexible and expressive prediction framework.

\noindent \textbf{Input Representation.}
Omics datasets generally consist of high-dimensional numerical measurements, such as protein abundances or DNA methylation levels, along with a small number of categorical annotations, for example sex or tissue. We integrate both numerical and categorical features to capture complementary age-related signals that are informative for the aging clock. Directly combining raw numerical values with categorical indicators can hinder learning due to differences in scale and representation. To address this, we first project numerical features $\mathbf{x}^{\text{num}} \in \mathbb{R}^{d_1}$ into a shared embedding space:
\begin{equation}\label{eq:numerical_mapping}
    \mathbf{e}_{\text{num}} = f_{\theta_{\text{num}}}(\mathbf{x}^{\text{num}}),
\end{equation}
where $f_{\theta_{\text{num}}}$ is the mapping function parametrized with $\theta_{\text{num}}$, which can generally be implemented as a simple linear transformation. For the categorical features $\mathbf{x}^{\text{cat}} \in \mathbb{R}^{d_2}$, we map each category $\mathbf{x}_c^{\text{cat}} \in \mathbb{R}$ to a dense embedding vector using learnable feature-specific embedding tables $Emb_c$:
\begin{equation}\label{eq:categorical_mapping}
    \mathbf{e}_{\text{c}} = Emb_c(\mathbf{x}_c^{\text{cat}}), \quad c=1, ..., d_2.
\end{equation}
We then aggregate all embeddings into a unified representation:
\begin{equation}\label{eq:repre_com}
    \mathbf{h}_0 = \frac{1}{d_2}\left(\sum_{c=1}^{d_2} \mathbf{e}_c \;+\; \mathbf{e}_{\text{num}} \right).
\end{equation}
This unified representation $\mathbf{h}_0$ serves as the input for subsequent model components, enabling integrated processing of both numerical and categorical features.

\noindent \textbf{Transformer Encoder Backbone.}
To capture complex relationships among integrated feature representations, we employ a Transformer encoder as a flexible feature extractor. The self-attention mechanism enables the model to adaptively reweight feature interactions and learn expressive representations from high-dimensional omics inputs. Formally, given the aggregated input representation $\mathbf{h}_0$, we first add a learnable positional embedding and pass the result through the Transformer encoder:
\begin{equation}\label{eq:tf_encoder}
    \mathbf{h} = \text{TransformerEncoder}_{\theta_t}(\mathbf{h}_0 \; + \; \mathbf{p}),
\end{equation}
where $\mathbf{p}$ denotes the positional embedding and $\mathbf{h}$ represents the encoded feature vector produced by the Transformer backbone. The positional embedding introduces a consistent structural signal to the input representation, which facilitates stable representation learning across encoder layers. The resulting representation $\mathbf{h}$ is then used as the input to the subsequent MoE regression head for age prediction.

\noindent \textbf{Mixture-of-Experts Regression for Aging Prediction.}
To model heterogeneous aging patterns in omics data, the AgeMoE aging clock employs a MoE regression module on top of the backbone representation. Since aging signals arise from multiple biological processes and exhibit diverse patterns across individuals, a single predictor may not capture all variability. To address this, we implement multiple expert heads and employ a learnable routing network that assigns input-dependent weights to each expert, enabling the model to combine specialized predictions into a final age estimate.

Given the encoded feature representation $\mathbf{h}$ from the Transformer backbone, we first compute routing weights using a routing network:
\begin{equation}\label{eq:routing}
    \bm{\pi} = g_{\theta_{r}}(\mathbf{h}) \in \mathbb{R}^{E},
\end{equation}
where $g_{\theta_{r}}(\cdot)$ denotes the routing function and $E$ is the number of experts. We apply a softmax function to obtain normalized routing weights satisfying $\sum_{e=1}^{E} \bm{\pi}^e=1$. These weights determine how strongly each expert contributes to the final prediction.

We implement each expert as an independent regression head that maps the shared representation to an age estimate. For expert $e=1, ..., E$, the predicted age is
\begin{equation}\label{eq:expert_head}
    \hat{y}^e = s_{\phi_e}(\mathbf{h}).
\end{equation}
We then combine expert outputs with routing weights to produce the final prediction:
\begin{equation}\label{eq:final_prediction}
    \hat{y} = \sum_{e=1}^E \bm{\pi}^e \cdot \hat{y}^e.
\end{equation}

\noindent \textbf{Loss Functions for AgeMoE Aging Clock.}
We train the AgeMoE predictor with two complementary objectives. The first is a base regression loss, which ensures accurate age estimation:
\begin{equation}\label{eq:regression_loss}
    \mathcal{L}_{reg} = \frac{1}{n_k}\sum_{i=1}^{n_k} (\hat{y}_i \;-\; y_i)^2
\end{equation}
where $n_k$ denotes the number of samples in center $k$, $\hat{y}_i$ is the predicted age, and $y_i$ is the true age for sample $i$.

Second, we incorporate an age-aware routing regularizer to encourage samples with similar ages to exhibit similar routing distributions. This constraint promotes smoother expert specialization across the age spectrum and helps stabilize the routing behavior of the MoE model. Let $\bm{\pi}_i \in \mathbb{R}^{E}$ denote the routing weights for sample $i$. For each sample, we consider its $l$ nearest neighbors in the age space and measure the difference between their routing distributions using the symmetric KL divergence, whose symmetric formulation removes directional bias and provides a balanced measure of distributional similarity:
\begin{equation}\label{eq:kl}
    D_{\text{symKL}}(\bm{\pi}_i, \bm{\pi}_j) = D_{\text{KL}}(\bm{\pi}_i || \bm{\pi}_j) \;+\; D_{\text{KL}}(\bm{\pi}_j || \bm{\pi}_i).
\end{equation}
Then, the routing regularization loss is defined as:
\begin{equation}\label{eq:target_loss}
    \mathcal{L}_{tr} = \frac{1}{n_k}\sum_{i=1}^{n_k} \frac{1}{l}\sum_{{j \in \mathcal{N}_l(i)}} \omega_{ij} \cdot D_{\text{symKL}}(\bm{\pi}_i, \bm{\pi}_j),
\end{equation}
where $\mathcal{N}_l(i)$ denotes the set of $l$ nearest neighbors of sample $i$ in the age space and
\begin{equation}\label{eq:tr_weight}
    \omega_{ij} = \text{exp}(-\frac{|y_i - y_j|}{\tau}),
\end{equation}
assigns larger weights to pairs with closer ages, with $\tau$ controlling the decay rate.

The overall loss for training the AgeMoE aging clock is:
\begin{equation}\label{eq:overall_loss}
    \mathcal{L}_{AgeMoE}(\Theta) = \mathcal{L}_{reg} + \lambda \mathcal{L}_{tr},
\end{equation}
where $\Theta=\{\theta_{\text{num}}, \{Emb_c\}_{c=1}^{d_2}, \theta_t, \theta_r, \{\phi_e\}_{e=1}^E\}$ denotes the set of all parameters in the aging clock model, including the numerical feature mapping, categorical embedding tables, Transformer encoder, routing network, and expert heads, and $\lambda$ is a hyperparameter that balances the regression and regularization terms.

\subsubsection{Generative Replay for Cross-Center Knowledge Preservation}
As the evolving aging-clock model is sequentially updated across heterogeneous centers, continued adaptation to new local data may weaken information acquired from previously visited centers. To preserve such information during cross-center learning, \model incorporates generative replay through an auxiliary conditional generative model. The generator synthesizes age-conditioned pseudo-samples representing information accumulated from preceding centers, which are replayed together with local data during subsequent model updates.

We implement the generative replay module using a conditional variational autoencoder (CVAE) that models the conditional distribution of omics features given age-related information. In addition to the age value, we introduce a discrete age-bin identifier $b$ defined over the global age range with a fixed bin width. Conditioning on both the continuous age value and the age-bin indicator allows the model to capture coarse age-level information while preserving fine-grained variation within each bin. The CVAE consists of an encoder and a decoder parameterized by neural networks. The encoder maps the input sample and its conditioning variables to a latent representation:
\begin{equation}\label{eq:cvae_encoder}
    q_\psi(\mathbf{z}|\mathbf{x},y,b),
\end{equation}
where $\mathbf{z}$ denotes the latent variable. The decoder reconstructs the feature vector from the latent variable together with the conditioning information:
\begin{equation}\label{eq:cvae_decoder}
    p_\varphi(\mathbf{x}|\mathbf{z},y,b).
\end{equation}
The model is trained by minimizing the following objective:
\begin{equation}\label{eq:cvae_loss}
    \mathcal{L}_{cvae}(\Psi) = ||\mathbf{x}-\hat{\mathbf{x}}||_2^2 \;+\; \beta D_{\text{KL}}(q_\psi(\mathbf{z}|\mathbf{x},y,b) || \mathcal{N}(0,I)),
\end{equation}
where $\Psi=\{\psi, \varphi\}$ denotes all parameters in the generative model, including the encoder parameters and the decoder parameters, and $\hat{\mathbf{x}}$ denotes the reconstructed feature vector generated by the decoder. The first term measures the reconstruction error between the input features and their reconstruction, while the second term regularizes the latent distribution toward a standard Gaussian prior. After training, the CVAE can generate pseudo-samples by sampling $\mathbf{z} \sim \mathcal{N}(0,I)$ and decoding it together with specified age conditions $(y,b)$. These synthesized samples are replayed alongside local data during subsequent model updates to help preserve information acquired from previously visited centers.

% \subsubsection{Feasible Training Sequence from the Trust Network}
% Given the directed inter-institutional trust network $\mathcal{G}=(\mathcal{V},\mathcal{E})$, we construct the training sequence $\mathcal{S}=[s_1,\ldots,s_T]$ as a directed traversal of $\mathcal{G}$ that covers the participating centers. For every pair of consecutive centers, $(s_t,s_{t+1})\in\mathcal{E}$, and the corresponding message transmission is governed by $\operatorname{Trust}(s_t,s_{t+1},X_{s_t\rightarrow s_{t+1}})$. Centers may be revisited when required by the network topology. The resulting sequence determines the execution order of the collaborative updates in the subsequent optimization procedure.

% \red{\noindent\textbf{Remark 3 (Termination of Sequential Training).}
% SeqPairFL currently terminates after completing a predefined finite trust sequence $\mathcal{S}=[s_1,\ldots,s_T]$. The model obtained at $s_T$ is the final state of this sequence, but additional valid propagation steps or repeated visits to earlier centers may yield further improvement. Adaptive termination and revisitation strategies that balance predictive gains against communication and computation costs are left for future work.}
\subsection{Training and Optimization}
\subsubsection{Joint Training of Aging Clock and Generative Model}
In \model, the aging-clock and generative-model parameters are transmitted between successive centers through protocol-authorized messages along directed trust relations. To preserve information from previously visited clients while adapting to the current client’s data, we use the generative replay mechanism introduced above. The generative model produces pseudo-samples approximating previously encountered client distributions, and the aging clock uses these samples together with local data during model updates. This strategy helps preserve previously acquired information as the aging clock and generator are updated sequentially across centers.

At the initial step, center $s_1$ is assigned the model parameters $\Theta_0$ and $\Psi_0$. At each subsequent step $t>1$, center $s_t$ receives the protocol-authorized message $X_{s_{t-1}\rightarrow s_t}=(\Theta_{t-1},\Psi_{t-1})$ from the preceding center $s_{t-1}$. For the initial center $s_1$, pseudo-sample generation and prediction-consistency filtering are omitted because no preceding-center model state is available. The joint training procedure at each subsequent center consists of three steps:
\begin{enumerate}
    \item \textbf{Pseudo-sample generation and prediction-consistency filtering}
    
    \noindent We first generate pseudo-samples using the received generative model to represent information from previously visited clients. Specifically, we sample $n_{pse}$ latent vectors $\mathbf{z}_i \sim \mathcal{N}(0,I)$ and decode feature vectors conditioned on target ages $y_i$ and age-bin identifiers $b_i$:
    \begin{equation}\label{eq:sample_generation}
        \hat{\mathbf{x}}_i \sim p_{\varphi_{{t-1}}}(\mathbf{x}|\mathbf{z}_i,y_i,b_i), \quad i=1, ..., n_{pse},
    \end{equation}
    where target ages and age-bin identifiers are sampled from the age range maintained by the generator. Because generated samples may vary in quality, we further apply prediction-consistency filtering to retain reliable pseudo-samples. The aging clock received from the preceding center serves as a reference predictor, $\hat{y}_{ref}(\hat{\mathbf{x}}_i)=\mathcal{F}_{\Theta_{{t-1}}}(\hat{\mathbf{x}}_i)$. A pseudo-sample is kept if its predicted age is sufficiently consistent with the conditioning label:
    \begin{equation}\label{eq:filter}
        |\hat{y}_{ref}(\hat{\mathbf{x}}_i)-y_i| \leq \epsilon.
    \end{equation}
    The pseudo-samples satisfying this condition are retained to form the filtered set $\tilde{\mathcal{D}}_{s_t}$. This filtering step excludes pseudo-samples with inconsistent age predictions and improves the reliability of generated supervision during subsequent training.

    \item \textbf{Aging clock update}

    \noindent After obtaining the filtered pseudo-samples $\tilde{\mathcal{D}}_{s_t}$, we update the aging clock using them together with the current center’s data $\mathcal{D}_{s_t}$. The combined dataset allows the model to learn from the current center while retaining information from previously encountered client distributions. The aging clock parameters are optimized by minimizing the AgeMoE objective:
    \begin{equation}\label{eq:aging_clock_update}
        \Theta_t = \arg\min_{\Theta} \mathcal{L}_{AgeMoE}\big(\Theta; \mathcal{D}_{s_t} \cup \tilde{\mathcal{D}}_{s_t} \big).
    \end{equation}
    Through this update, the generated samples provide complementary supervision for preserving information acquired from previously visited clients during sequential model updates.

    \item \textbf{Generative model update}

    \noindent Finally, we update the generative model from the received parameters $\Psi_{{t-1}}$ using the current center’s data $\mathcal{D}_{s_t}$. This sequential update incorporates the current center’s data distribution into the propagated generator for subsequent interactions. The model parameters are optimized by minimizing the CVAE objective:
    \begin{equation}\label{eq:cvae_update}
        \Psi_{t} = \arg\min_\Psi\mathcal{L}_{cvae}(\Psi;\mathcal{D}_{s_t}).
    \end{equation}
    The updated generative model is then passed along the training sequence together with the aging clock.
\end{enumerate}
\begin{algorithm}[!t]
\caption{Sequential Federated Training over a Directed Trust Network}
\label{alg:sfl_training}
\begin{algorithmic}[1]
\REQUIRE Center datasets $\{\mathcal{D}_k\}_{k=1}^{K}$, initial parameters $\Theta_0$, $\Psi_0$, number of pseudo-samples $n_{pse}$, filtering threshold $\epsilon$, training sequence $\mathcal{S} = [s_1, ..., s_T]$
\ENSURE Final aging-clock model $\Theta_T$ and generative-model parameters $\Psi_T$

\STATE Initialize aging clock parameters $\Theta \leftarrow \Theta_0$
\STATE Initialize generative model parameters $\Psi \leftarrow \Psi_0$

\FOR{$t=1$ to $T$}

    \STATE \textbf{Receive the authorized model-parameter message for center $s_t$}
    \STATE $\Theta_{t}^{(0)} \leftarrow \Theta$
    \STATE $\Psi_{t}^{(0)} \leftarrow \Psi$

    \STATE \textbf{Pseudo-sample generation}
    \FOR{$i=1$ to $n_{pse}$}
        \STATE Sample latent vector $\mathbf{z}_i \sim \mathcal{N}(0,I)$
        \STATE Sample replay age conditions $(y_i,b_i)$
        \STATE Generate a pseudo sample $\hat{\mathbf{x}}_i$ with Eq.~(\ref{eq:sample_generation})
    \ENDFOR

    \STATE \textbf{Prediction-consistency filtering}
    \STATE Compute $\hat{y}_{ref}(\hat{\mathbf{x}}_i)=\mathcal{F}_{\Theta_{t}^{(0)}}(\hat{\mathbf{x}}_i)$ for all generated pseudo-samples
    \STATE Construct the filtered pseudo-sample set $\tilde{\mathcal{D}}_{s_t}$ according to Eq.~(\ref{eq:filter})

    \STATE \textbf{Aging clock update}
    \STATE Update aging clock parameters $\Theta_{t}$ by optimizing the AgeMoE objective in Eq.~(\ref{eq:aging_clock_update})
    \STATE Set propagated aging clock parameters $\Theta \leftarrow \Theta_{t}$

    \STATE \textbf{Generative model update}
    \STATE Update generative model parameters $\Psi_{t}$ by optimizing the CVAE objective in Eq.~(\ref{eq:cvae_update})
    \STATE Set propagated generative model parameters $\Psi \leftarrow \Psi_{t}$

\ENDFOR

\STATE $\Theta_T \gets \Theta$, $\Psi_T \gets \Psi$
\STATE \textbf{Final model release}
\STATE Release $\Theta_T$ to all participating centers in $\mathcal{V}$ according to the predefined protocol-level release policy in Eq.~(\ref{eq:release})
\RETURN $\Theta_T$

\end{algorithmic}
\end{algorithm}
Overall, this joint training procedure allows the aging clock and generative model to be updated together as they propagate across centers through protocol-authorized transmissions under directed trust relations. Filtered pseudo-samples support replay from previously encountered client distributions, while sequential generator updates incorporate new local data for subsequent interactions.

\subsubsection{Sequential Optimization Formulation}

Given a valid training sequence $S=[s_1,\ldots,s_T]$, the \model training process can be summarized as a sequence of recursively initialized local optimization steps. Let $\mathcal{D}_{s_t}$ denote the local dataset of the center visited at step $t$, and let $\tilde{\mathcal{D}}_{s_t}$ denote the filtered pseudo-sample set used for generative replay. For the initial center $s_1$, no replay set from preceding centers is available, and we define $\tilde{\mathcal{D}}_{s_1}=\varnothing$. For each subsequent step $t>1$, the filtered replay set is constructed as
\begin{equation}
\tilde{\mathcal{D}}_{s_t}
=
\left\{
(\hat{x}_i,y_i)
\;\middle|\;
\hat{x}_i \sim p_{\phi_{t-1}}(x \mid z_i,y_i,b_i),
\;
\left|
F_{\Theta_{t-1}}(\hat{x}_i)-y_i
\right|
\leq \epsilon
\right\}.
\end{equation}
The aging-clock parameters are updated as
\begin{equation}
\Theta_t
=
\arg\min_{\Theta}
\mathcal{L}_{\mathrm{AgeMoE}}
\left(
\Theta;
\mathcal{D}_{s_t}\cup\tilde{\mathcal{D}}_{s_t}
\right),
\qquad
\Theta_t^{(0)}=\Theta_{t-1},
\end{equation}
and the generative-model parameters are updated as
\begin{equation}
\Psi_t
=
\arg\min_{\Psi}
\mathcal{L}_{\mathrm{cvae}}
\left(
\Psi;
\mathcal{D}_{s_t}
\right),
\qquad
\Psi_t^{(0)}=\Psi_{t-1},
\end{equation}
for $t=1,\ldots,T$, where $\Theta_0$ and $\Psi_0$ denote the initial aging-clock and generative-model parameters, respectively. Here, $\Theta_t$ and $\Psi_t$ denote the model parameters after the update at step $t$, while $\Theta_t^{(0)}$ and $\Psi_t^{(0)}$ denote their local initialization before optimization. For $t>1$, these initializations are inherited from the protocol-authorized message transmitted by the preceding center under the corresponding directed trust relation. After the final update, TNFL outputs the aging-clock model $\Theta_T$. The complete update procedure is summarized in Algorithm~1.

\subsection{Trust Characterization and Security Properties}
The directed trust relation defines asymmetric and message-specific obligations for the sender and receiver, and therefore differs in scope from conventional participant-level behavioral models. This subsection clarifies this distinction and establishes a security property arising along sequences of separately
established pairwise trust relations.

\subsubsection{Relation to Semi-Honest and Fully Honest Models}
\begin{figure}[!t]
    \centering
    \includegraphics[width=0.82\linewidth]{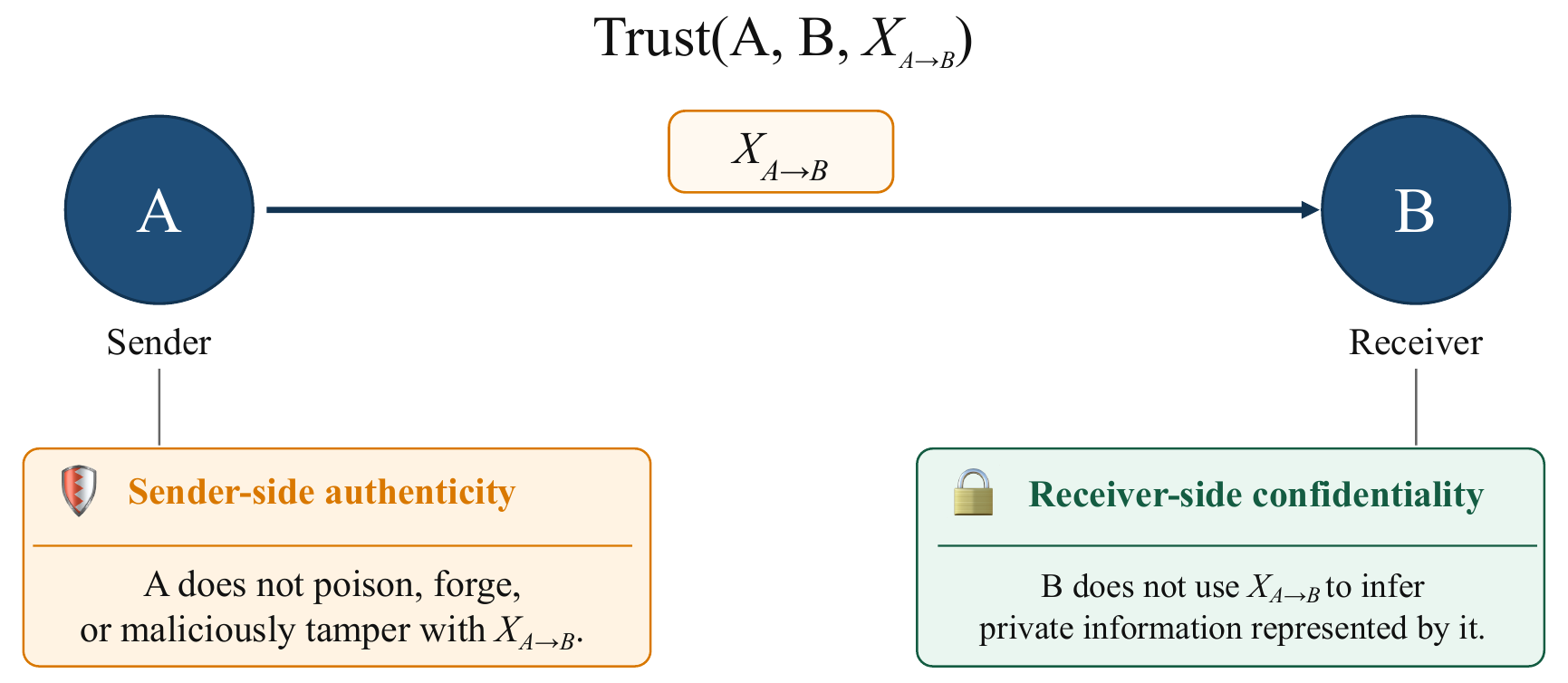}
    \caption{Illustration of the sender-side authenticity and receiver-side confidentiality obligations under
    $\operatorname{Trust}(A,B,X_{A\rightarrow B})$.}
    \label{fig:directed_trust}
\end{figure}

Semi-honest and fully honest models characterize the behavior of an individual
participant. A semi-honest participant follows the prescribed protocol and does not
poison, forge, or tamper with protocol messages, but may use legitimately obtained
information to infer private information. A fully honest participant additionally
refrains from such inference. In contrast,
$\operatorname{Trust}(A,B,X_{A\rightarrow B})$ is a directed and pairwise relation
that assigns different obligations to the sender and receiver of a specific message. As illustrated in Fig.~\ref{fig:directed_trust}, the relation imposes sender-side authenticity on $A$ and receiver-side confidentiality on $B$. Building on this relation-level characterization, Table~\ref{tab:trust_comparison} applies the semi-honest and fully honest models to both $A$ and $B$ and contrasts their requirements with those imposed by $\operatorname{Trust}(A,B,X_{A\rightarrow B})$.

\begin{table}[t!]
    \centering
    \caption{Comparison of trust with semi-honest and fully honest models.}
    \label{tab:trust_comparison}
    \small
    \setlength{\tabcolsep}{4pt}
    \renewcommand{\arraystretch}{1.35}
    \begin{tabular}{@{}lcccc@{}}
        \toprule
        \textbf{Setting}
        & \shortstack{\textbf{Requires $A$}\\\textbf{not to poison?}}
        & \shortstack{\textbf{Requires $B$}\\\textbf{not to poison?}}
        & \shortstack{\textbf{Requires $A$}\\\textbf{not to infer?}}
        & \shortstack{\textbf{Requires $B$}\\\textbf{not to infer?}} \\
        \midrule
        \shortstack[l]{Both $A$ and $B$ are\\semi-honest}
        & Yes
        & Yes
        & No
        & No \\
        \specialrule{0.4pt}{1pt}{4pt}
        \shortstack[l]{Both $A$ and $B$ are\\fully honest}
        & Yes
        & Yes
        & Yes
        & Yes \\
        \specialrule{0.4pt}{1pt}{4pt}
        $\operatorname{Trust}(A,B,X_{A\rightarrow B})$
        & Yes
        & No
        & No
        & Yes \\
        \bottomrule
    \end{tabular}
\end{table}

\subsubsection{Security Properties along Trust Sequences}
\model composes multiple directed trust relations through sequential model
propagation, whereas Definition~1 specifies a receiver-side confidentiality obligation at the level of an individual transmission. This subsection analyzes the security property induced by
this composition and establishes downstream non-inference for upstream centers along
a valid trust sequence.

\noindent\textbf{Proposition 1 (Downstream Confidentiality along a Trust Sequence).}
Let
\[
S=[v_1,v_2,\ldots,v_T]
\]
be a sequence of participating centers, where
$X_{v_i\rightarrow v_{i+1}}$ denotes the message transmitted from $v_i$ to
$v_{i+1}$. For every $i\in\{1,\ldots,T-1\}$, the information about preceding model states exposed to $v_{i+1}$ through the \model training protocol is contained in $X_{v_i\rightarrow v_{i+1}}$. If
\[
\operatorname{Trust}
\left(
v_i,v_{i+1},X_{v_i\rightarrow v_{i+1}}
\right)
\]
holds for every $i\in\{1,\ldots,T-1\}$, then, for any
$1\leq i<j\leq T$, center $v_j$ does not use its received message to infer the
private information of center $v_i$.

\noindent\textit{\textbf{Proof.}}
We prove the proposition by induction on the number of centers $n$ in a sequence
prefix.

\noindent\textit{\textbf{Base case}} ($n=2$).
By
\[
\operatorname{Trust}
\left(
v_1,v_2,X_{v_1\rightarrow v_2}
\right)
\]
and the confidentiality requirement in Definition~1, center $v_2$ does not use
$X_{v_1\rightarrow v_2}$ to infer the private information of center $v_1$.
Therefore, the proposition holds for the first two centers.

\noindent\textit{\textbf{Induction hypothesis}} ($n$).
Assume that, for the sequence prefix
\[
(v_1,v_2,\ldots,v_n),
\]
for any $1\leq i<j\leq n$, center $v_j$ does not use its received message to
infer the private information of center $v_i$.

\noindent\textit{\textbf{Induction step}} ($n+1$).
By the induction hypothesis, the proposition holds for the first $n$ centers.
Moreover, by
\[
\operatorname{Trust}
\left(
v_n,v_{n+1},X_{v_n\rightarrow v_{n+1}}
\right),
\]
the confidentiality requirement in Definition~1 ensures that $v_{n+1}$ does not
use $X_{v_n\rightarrow v_{n+1}}$ to infer the private information of center
$v_n$. By Remark~1, $v_{n+1}$ also does not use
$X_{v_n\rightarrow v_{n+1}}$ to infer the private information of any preceding
center $v_i$, where $1\leq i<n$. Therefore, for every $1\leq i\leq n$,
center $v_{n+1}$ does not use $X_{v_n\rightarrow v_{n+1}}$ to infer the private
information of center $v_i$. Hence, the proposition holds for the first $n+1$
centers.

By mathematical induction, for any $1\leq i<j\leq T$, center $v_j$ does not use
its received message to infer the private information of center $v_i$.
\hfill$\square$

% \section{Experiments}
\subsection{Experimental Design and Evaluation Setup}

Following the methodological formulation above, we next describe the experimental design used to evaluate \model. This includes the datasets and multi-center construction, model configurations, and evaluation settings used throughout the experiments.

\subsubsection{Datasets and Multi-Center Construction}

We use two molecular aging-clock datasets covering distinct omics modalities: proteomic data from the UK Biobank and DNA methylation data curated from the Gene Expression Omnibus (GEO). These datasets provide complementary molecular modalities for constructing the multi-center evaluation settings used in this study.

\noindent \textbf{\textit{Data Source}}

\noindent \textbf{UK Biobank Proteomics Dataset.}
The UK Biobank proteomics dataset consists of 46,988 samples, with each sample characterized by expression measurements of 2,920 plasma proteins. The chronological age of participants ranges from 39 to 72 years.

\noindent \textbf{GEO DNA Methylation Dataset.}
The DNA methylation dataset is assembled from publicly available studies deposited in the Gene Expression Omnibus (GEO), a widely used repository for functional genomics data. This dataset comprises 6,392 samples profiled using the Illumina HumanMethylation450 BeadChip array, with methylation levels measured at 485,514 CpG sites and participant ages spanning from 20 to 80 years. These samples were collected and curated through the EWAS Data Hub hosted by the National Genomics Data Center, which aggregates and standardizes DNA methylation array data and associated metadata from multiple public studies.

\noindent \textbf{\textit{Data Preprocessing}}

For both datasets, quality control and preprocessing are performed prior to model training. Molecular features with missing values in more than 10\% of samples are first removed to mitigate the impact of sparsely observed signals. For the remaining features, missing values are imputed using mean values computed within the corresponding sample group; for the DNA methylation dataset, imputation is conducted separately within each tissue to preserve tissue-specific methylation patterns. After filtering and imputation, molecular features are standardized to ensure comparable scales across samples. For the UK Biobank proteomics dataset, standardization is performed across all samples, whereas for the GEO methylation dataset it is carried out within each tissue type to account for inherent tissue-specific differences in molecular measurements.

In addition, for the GEO dataset, CpG sites are further filtered based on their age association. For each site, Pearson correlations between methylation levels and chronological age are computed within individual tissues, and a weighted average correlation is obtained using tissue sample proportions as weights. Sites are retained only if their weighted correlation exceeds a data-driven threshold determined from the upper quantile of the correlation distribution and if the threshold is satisfied in at least 50\% of tissues. This procedure preserves CpG sites with consistent age-related signals across multiple tissues while reducing noise from weakly associated loci.

\begin{figure}[!t]
    \centering
    % 全局间距优化：减小列间距，保证布局紧凑
    \setlength{\tabcolsep}{0.5em}
    \renewcommand{\arraystretch}{1.0}

    % ========== 第一排：左侧文字块(0.48) + 右侧子图a(0.48) ==========
    \begin{subfigure}[t]{0.48\textwidth}  % [t] 顶端对齐
        \centering
        \includegraphics[width=\linewidth]{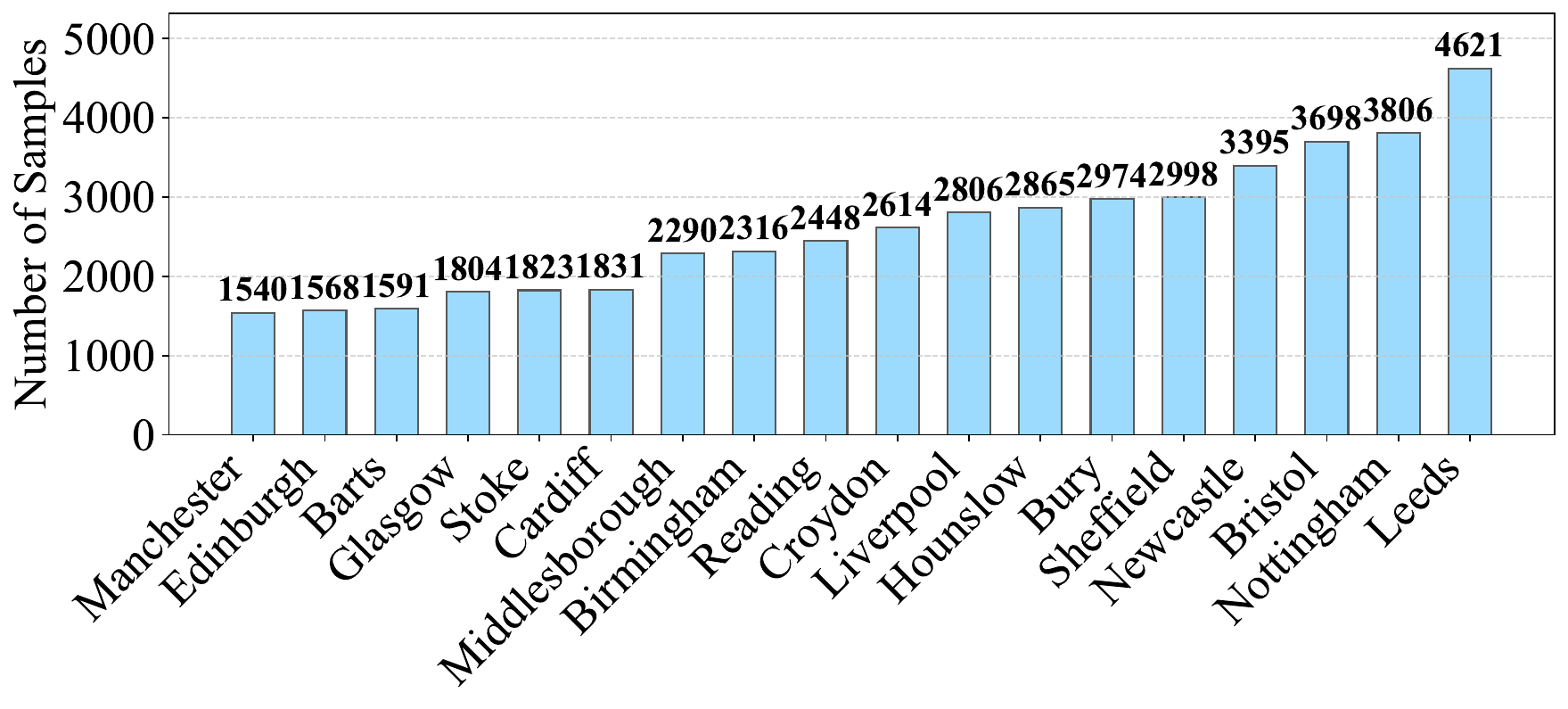}
        \subcaption{UKB-Center}
    \end{subfigure}
    \hfill  % 自动填充列间空白
    \begin{subfigure}[t]{0.48\textwidth}  % [t] 顶端对齐
        \centering
        \includegraphics[width=\linewidth]{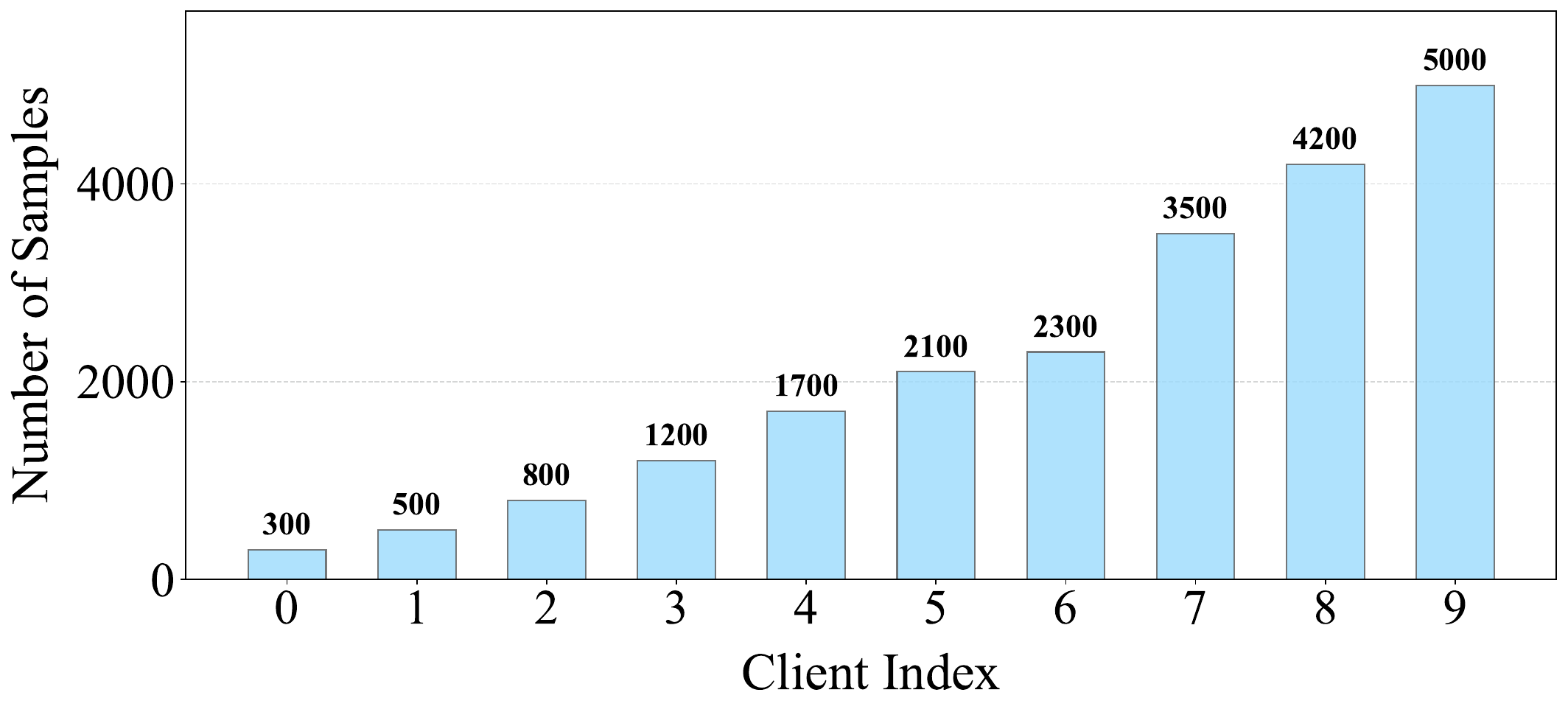}
        % 可选：添加子图标题（不需要则注释）
        \subcaption{UKB-Imbalance}
        % \label{subfig:2a}
    \end{subfigure}

    \vspace{0.2em}  % 行间距，可按需调整

    % ========== 第二排：子图b(0.48) + 子图c(0.48) ==========
    \begin{subfigure}[t]{0.48\textwidth}
        \centering
        \includegraphics[width=\linewidth, valign=t]{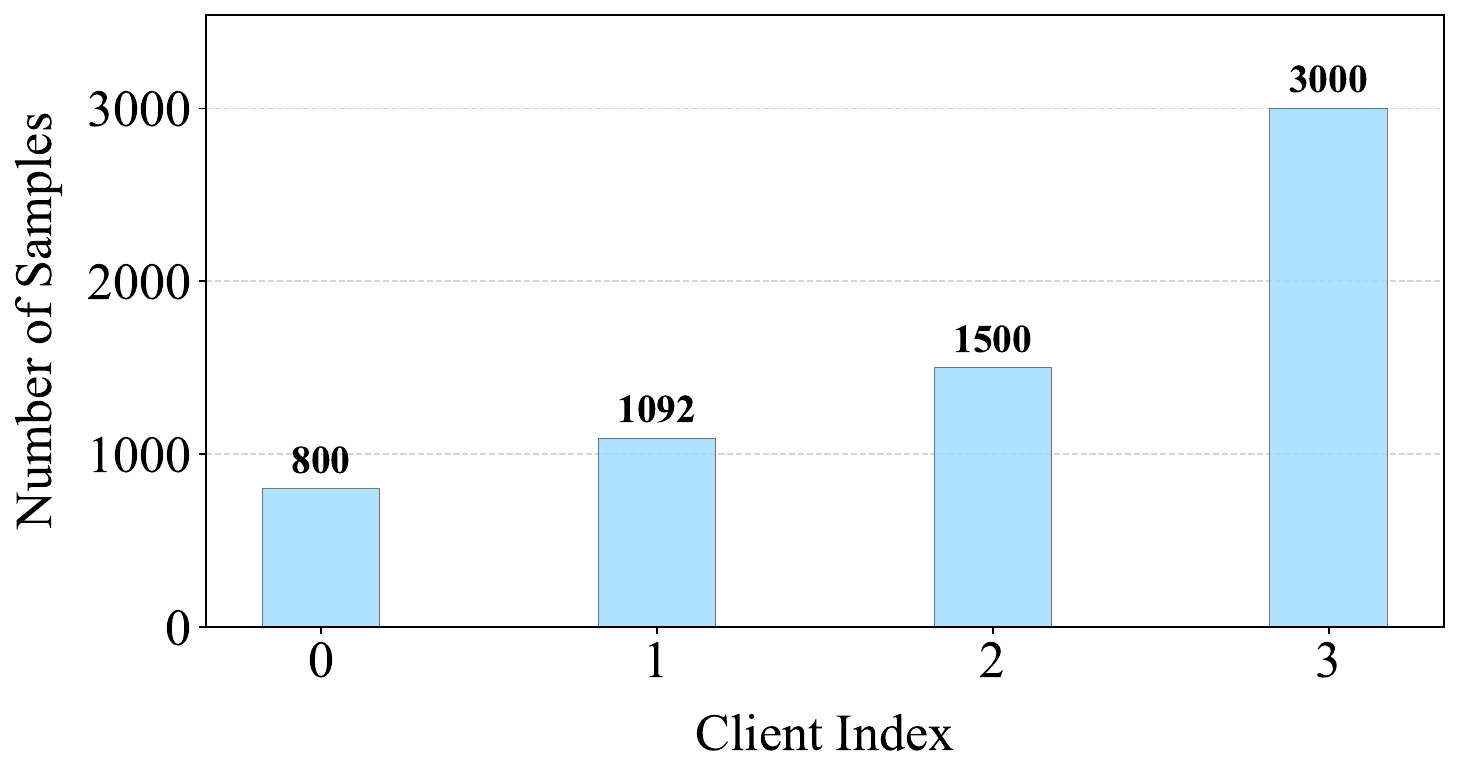}
        \subcaption{GEO-Methylation}
        % \label{subfig:b}
    \end{subfigure}
    \hfill
    \begin{subfigure}[t]{0.48\textwidth}
        \centering
        \includegraphics[width=\linewidth, valign=t]{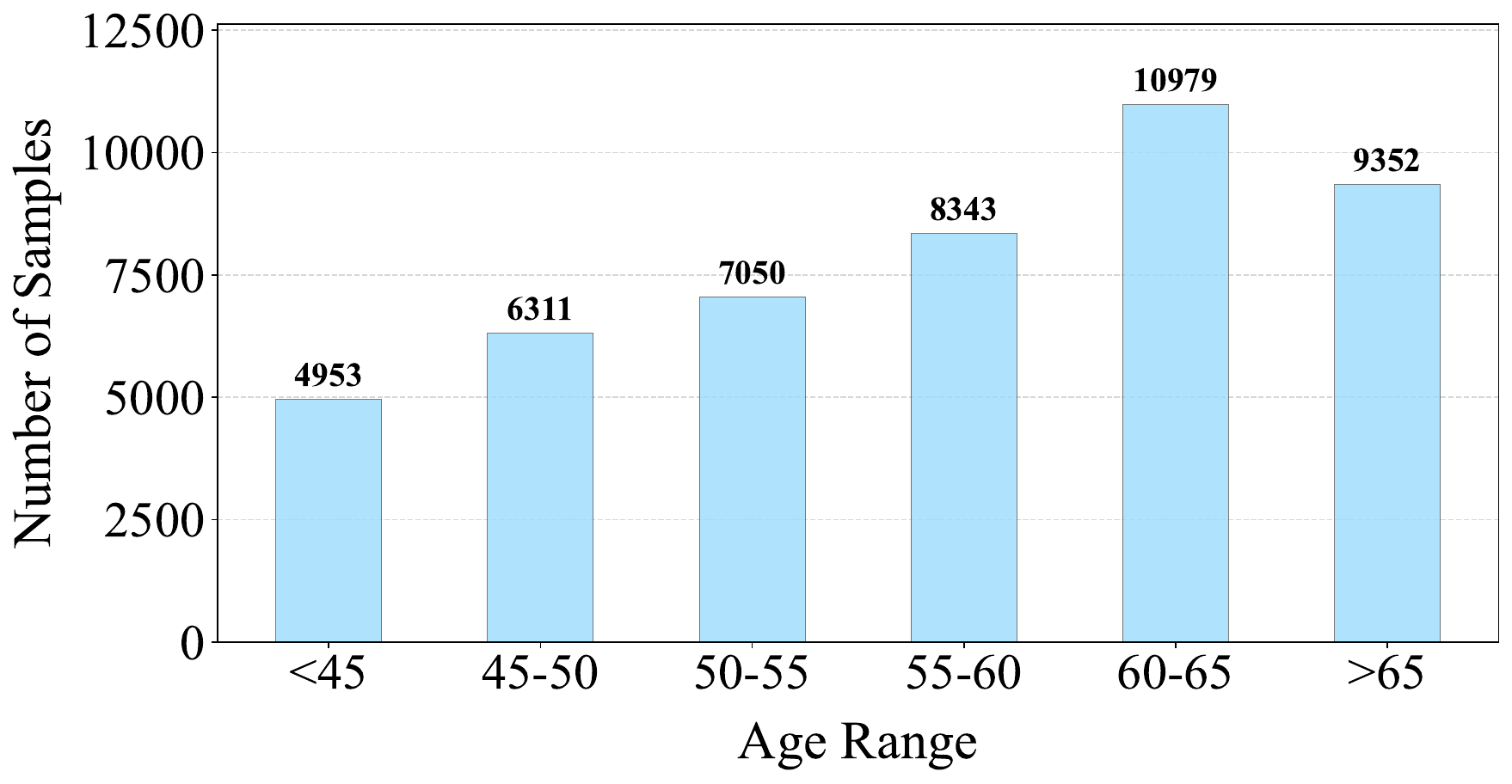}
        \subcaption{UKB-AgeSplit}
        % \label{subfig:c}
    \end{subfigure}
    \caption{Distribution of client sample sizes in constructed multi-center datasets.}
    \label{fig:dataset}
\end{figure}
\noindent \textbf{\textit{Multi-Center Construction}}

To emulate realistic multi-center learning environments, we construct federated clients by partitioning each dataset according to practical data-distribution scenarios. A summary of the constructed multi-center settings is provided in Fig.~\ref{fig:dataset}, which presents the sample size distribution across clients for each dataset.

\noindent \textbf{UKB-Center.}
For the UK Biobank proteomics dataset, samples are grouped based on their recruitment assessment centers. This partitioning naturally yields 18 clients corresponding to geographically distributed collection sites, reflecting the data silos typically encountered in large-scale biomedical collaborations.

\noindent \textbf{UKB-Imbalance.}
In addition, to evaluate model robustness under severe client-level data imbalance, we construct an alternative setting derived from the UK Biobank dataset. In this scenario, a subset comprising 21,600 samples is organized into 10 clients with highly skewed sample sizes. The resulting distribution exhibits substantial heterogeneity, with the smallest client containing only 300 samples, thereby mimicking extreme imbalance conditions that may arise in real-world collaborative studies.

\noindent \textbf{GEO-Methylation.}
For the GEO DNA methylation dataset, the overall sample size is comparatively smaller and the data originate from multiple studies with heterogeneous sample compositions. To reflect these characteristics, the dataset is partitioned into 4 clients with markedly imbalanced sample sizes, representing multi-center environments in which participating centers contribute datasets of varying scales.

\subsubsection{Model Configurations}

All aging clock models are formulated as supervised regression models that map high-dimensional molecular feature vectors to chronological age. To ensure a controlled comparison within the \model framework, model architectures and hyperparameters are fixed across centers within each experiment.

\noindent \textbf{Conventional Machine Learning Models.}
Linear regression and XGBoost are used as representative non-neural aging clock backbones. Linear regression models age as a linear function of molecular features with an \(\ell_1\)-regularized objective, providing a simple low-capacity reference. XGBoost serves as a stronger tree-based baseline by learning an additive ensemble of regression trees under squared-error loss.

\noindent \textbf{Deep Learning Models.}
We further consider plain MLP and Transformer backbone algorithms under the \model pipeline. The MLP consists of three fully connected layers with GeLU activations, serving as a standard feedforward network for tabular omics data. The Transformer employs a lightweight self-attention encoder with multiple attention heads and position-wise feedforward layers, adapted for tabular input.

\noindent \textbf{AgeMoE Aging Clock.}
Our AgeMoE aging clock consists of a Transformer-based encoder followed by a mixture-of-experts regression head. A CVAE-based generative replay module synthesizes pseudo-samples for replay during subsequent training within the \model pipeline. These pseudo-samples, conditioned on age and discrete age-bin identifiers and filtered to retain only those whose predicted ages match the conditioning label within a tolerance $\epsilon$ (set to 1 year), are used alongside the current client’s real data to train the aging clock. The specific model and training configurations are as follows.

For the \textbf{UK Biobank} experiments, we adopt a lightweight configuration. The aging clock uses a single-layer Transformer encoder with hidden dimension 64, one attention head, and feed-forward dimension 128, followed by a MoE head with 6 experts of hidden size 256. Training is performed with a batch size of 512 and a learning rate of $1e-4$, with a local budget of 40 epochs per stage and early stopping (patience of 5 epochs). The generative replay module produces pseudo-samples whose number is defined relative to the size of the current client’s dataset. Specifically, the replay ratio increases linearly from 0.3 to 0.5 over the training process, meaning that the number of pseudo-samples grows from 30\% to 50\% of the real samples at each client. The generated samples are distributed across 5-year age bins to ensure coverage of the age range. The CVAE used for generation employs a latent variable of dimension 32; both the encoder and decoder are implemented as two-layer MLPs with hidden dimensions 256 and 128, and take as input the molecular features concatenated with a 16-dimensional conditioning vector. The model is trained with a learning rate of $5e-4$, a batch size of 256, and 30 training epochs at each client.

For the \textbf{GEO} experiments, we adopt a higher-capacity configuration to accommodate increased data heterogeneity. The aging clock uses a Transformer encoder with hidden dimension 256, 4 attention heads, and a feed-forward dimension of 512, followed by a MoE head with 12 experts of hidden size 512. Training is performed with a batch size of 512 and a learning rate of $1e-4$, with up to 80 local epochs at each client and early stopping (patience of 20 epochs). The replay mechanism follows the same design as above, where the number of pseudo-samples is defined relative to the size of the current client’s dataset. The replay ratio increases linearly from 0.2 to 0.5, meaning that the number of generated samples grows from 20\% to 50\% of the real samples at each client. The generated samples are distributed across 5-year age bins. The CVAE used for generation employs a latent variable of dimension 128; both the encoder and decoder are implemented as two-layer MLPs with hidden dimensions 512 and 256, and take as input the molecular features concatenated with a 64-dimensional conditioning vector. The model is trained with a learning rate of $3e-4$, a batch size of 512, and 40 training epochs at each client.

\noindent \textbf{Training Configuration Consistency.}
For neural network--based aging clocks, including the MLP, Transformer, and AgeMoE models, we keep the training protocol, optimizer, and major hyperparameter settings consistent where applicable, while retaining model-specific training objectives. This design ensures that performance differences primarily reflect model architecture rather than unrelated training settings. Conventional machine learning baselines, including linear regression and XGBoost, are trained using their standard optimization procedures with fixed hyperparameter settings.

\subsubsection{Evaluation Setup}

\noindent \textbf{Data Splits.}
For each client, available samples are randomly partitioned into training, validation, and test sets with a ratio of 70:10:20. Local test sets are used to evaluate client-specific predictive performance, while a global test set is constructed by aggregating the test samples from all clients to assess overall generalization.

\noindent \textbf{Comparison Methods.}
We compare \model with two reference settings. Local training independently trains an aging clock at each center using only its local data, providing a no-collaboration reference, whereas FedAvg provides a standard server-based federated learning reference with centralized model aggregation. \model is evaluated under the trust-network-constrained setting defined above, where model propagation follows valid trust-constrained sequences composed of feasible pairwise trust relations.

\noindent \textbf{Evaluation Metrics.}
Predictive performance is evaluated using the mean absolute error (MAE). The metric is computed on both local test sets and the global test set to characterize client-level predictive accuracy as well as overall model performance. Interaction-order robustness is further assessed using client-level MAE range, cumulative MAE degradation, and forgetting rate, with detailed definitions provided in Appendix~\ref{app:robustness}.

\noindent \textbf{Interaction Order Generation.}
To examine the sensitivity of the \model protocol to interaction order across valid trust-constrained sequences, multiple valid single-pass trust sequences are instantiated from random client permutations. Specifically, 500 sequences are sampled for the UKB-Center dataset, 50 sequences for the UKB-Imbalance dataset, and 24 sequences for the GEO-Methylation dataset. The number of sequences is determined in accordance with the client scale of each dataset; for the GEO-Methylation dataset with four clients, all possible interaction permutations are evaluated. Model performance is evaluated under each sequence to quantify the robustness of \model to different interaction orders.

\noindent \textbf{Computational Environment.}
All experiments are conducted on a server equipped with an NVIDIA RTX 5090 GPU (32 GB memory) and an Intel Xeon Platinum 8470Q CPU with 25 virtual cores. The software environment is based on Python 3.12 running on Ubuntu 22.04, with model training implemented using PyTorch 2.8.0 and CUDA 12.8 for GPU acceleration.

\section{Declarations}
\begin{itemize}
\item Funding: This work was supported in part by the PolyU Start-up Fund (No. P0059983), the Presidential Young Scholar Scheme (Project No. P0056638), the Research Institute for Artificial Intelligence of Things (RIAIoT) at The Hong Kong Polytechnic University (Project No. P0059914), and the Research Institute for Federated Learning (4\-CG00) at The Hong Kong Polytechnic University. This work also received support from the Joint Research Centre for AI Aging Science and Medicine under the PolyU Academy for Artificial Intelligence. The Department of Aging Science and Medicine is an endowed department supported and funded by GAKKEN HOLDINGS CO., LTD. and Medical Care Service Inc.
\item Competing interests: The authors declare no competing interests.
\item Data availability statement: The data used in this study include publicly available and controlled-access datasets. DNA methylation data were obtained from publicly available GEO-derived resources. The UK Biobank data are available under an approved application and are not publicly available due to access restrictions.
\item Code availability: The complete code used in this study is provided in the Supplementary Materials accompanying this manuscript.
\item Consent for publication: Not applicable. This study does not report identifiable individual-level information.
\item Author contribution: Qiang Yang conceived the main idea and supervised the study. Chunxu Zhang and Bo Li designed the method, developed the model, conducted experiments, analyzed results, and drafted the manuscript. Liu Yang, Di Jiang and Bo Yang provided methodological suggestions and contributed to manuscript revision. Wenliang Wang and Yuan Huang contributed to biological analysis, functional annotation, and interpretation of model-identified molecular signals. Yo-ichi Nabeshima and Akinori Yamamura provided expert biological interpretation from the perspective of aging biology and reviewed the biological relevance of the identified molecular modules and higher-order organization. All authors reviewed and approved the manuscript.
\end{itemize}

\begin{appendices}
\section{Details About Biological Analysis of Key Molecular Features}\label{app:geo_bio_analysis}
\subsection{Selection of Key Omics Features via Age-Dependent Expert Sensitivity}
To identify important molecular features associated with aging, we leveraged the routing weight of an expert that increases with age, providing an aging-related reference representation. This age-dependent behavior indicates that the routing weight of this expert reflects an age-associated routing pattern, which serves as a data-driven basis for assessing feature contributions to aging-related model behavior.

For each protein (or CpG site in the methylation data), we conducted a controlled perturbation analysis. Specifically, each feature was independently varied across ten quantile levels derived from its empirical distribution, while all other features were held fixed. The resulting changes in expert routing weight were recorded. Feature influence was quantified as the difference in expert routing weight between the highest and lowest quantile levels of each feature, and features were subsequently ranked based on this score. 

% Based on this procedure, the top-ranked proteins derived from the UKB-Center dataset are provided in the Supplementary Materials as Top-Proteins.json, and the top-ranked DNA methylation features derived from the GEO-Methylation dataset are provided as Top-Methylation.json.

\subsection{Pathway–Gene Composition of Enriched Functional Clusters}
To complement the pathway enrichment analysis of the top-ranked proteins derived from the UKB-Center proteomic dataset, we provide the full gene-level composition of all significantly enriched ontology clusters. Specifically, each of the 20 enriched pathways identified in the main text is further decomposed to explicitly list the constituent proteins contributing to each functional module. This enables direct inspection of how individual proteins map onto biologically coherent processes. 

% The complete mapping between enriched pathways and their corresponding genes is provided in the Supplementary Materials as pathway-gene.json.

\subsection{Model-based Identification of Synergistic Protein Pairs}
To investigate coordinated protein effects associated with aging, we analyzed pairwise protein interactions learned by the aging clock model. Based on the important proteins identified in the previous section, we constructed candidate protein pairs by exhaustively combining these proteins into all possible two-protein combinations. We then evaluated how each protein pair jointly influenced the biological age estimated by the model on the global test population.

For a protein pair $(i,j)$, we quantified its interaction effect by comparing model outputs under four perturbation states generated on the global test population. Specifically, we set the values of the corresponding proteins to zero across all samples to simulate suppression while leaving all remaining proteins unchanged. The four conditions included: both proteins suppressed ($f_{00}$), only protein $i$ retained at its original values while protein $j$ was suppressed ($f_{10}$), only protein $j$ retained while protein $i$ was suppressed ($f_{01}$), and both proteins retained at their original values ($f_{11}$). Based on these four conditions, we defined the pairwise synergy score as:
\begin{equation}
\text{Synergy}_{ij}=f_{11}-f_{10}-f_{01}+f_{00}.
\end{equation}

A positive synergy score indicates that the joint contribution of the two proteins exceeds the sum of their individual effects, suggesting coordinated influence on aging-related prediction patterns captured by the model. In contrast, negative synergy scores suggest redundant or antagonistic relationships between the two proteins. We ranked all candidate protein pairs according to their synergy scores and focused the main-text analysis on the top 10 synergistic protein pairs. 
% In addition, the complete top 100 ranked pairs were provided in the supplementary file top100-protein-pairs.json as an extended resource for reference.

\subsection{Model-based Discovery of Higher-Order Protein Interactions}
Beyond pairwise protein interactions, we further investigated whether the aging clock model captures coordinated higher-order protein organization associated with aging. To this end, we constructed candidate protein sets from model-identified important proteins for subsequent joint analysis of their collective effects.

To characterize protein coordination from multiple complementary biological perspectives, we performed higher-order synergy analysis under four analytical settings. Specifically, we considered interactions computed from either the final biological age prediction output or the latent expert-routing output of the AgeMoE model, enabling the analysis to capture both phenotype-level coordination and latent functional specialization patterns. In addition, we evaluated these interactions using either the full population distribution or age-restricted young subsets. The full-population setting prioritizes globally stable coordination patterns shared across heterogeneous individuals, whereas the young-subset setting is more sensitive to early-stage or age-dependent interaction signals that may be diluted in the overall population. Together, these complementary settings allow the analysis to capture both globally conserved and context-specific higher-order aging organization.

The group construction process follows a beam-search–style strategy that incrementally builds protein sets from informative starting pairs. Specifically, candidate protein pairs with strong pairwise synergy are first used as initial seeds. These seed pairs are then progressively expanded by adding one protein at a time, forming larger candidate groups. At each expansion step, candidate groups are evaluated based on their overall collective effect under the model, and only a subset of the most promising group configurations is retained for further expansion. Across the four analytical settings, this procedure yields a total of 16 representative protein groups. 

% These protein groups are provided in the Supplementary Materials as group-proteins.json.

\subsection{Biological Analysis on Methylation
Data}
We perform complementary biological analyses on the GEO DNA methylation dataset to assess whether the molecular patterns observed in the proteomic data extend across modalities. Feature importance is defined based on the magnitude of changes in expert routing weights induced by systematic perturbations, consistent with the definition used in the main text. Based on this criterion, the top 300 CpG sites are selected for downstream analysis, representing the most influential methylation features associated with expert routing behavior.

\begin{figure}[!t]
    \centering
    % 全局间距优化：减小列间距，保证布局紧凑
    \setlength{\tabcolsep}{0.5em}
    \renewcommand{\arraystretch}{1.0}

    % ========== 第一排：子图==========
    \begin{subfigure}[t]{\textwidth}
        \centering
        \includegraphics[width=1.\linewidth]{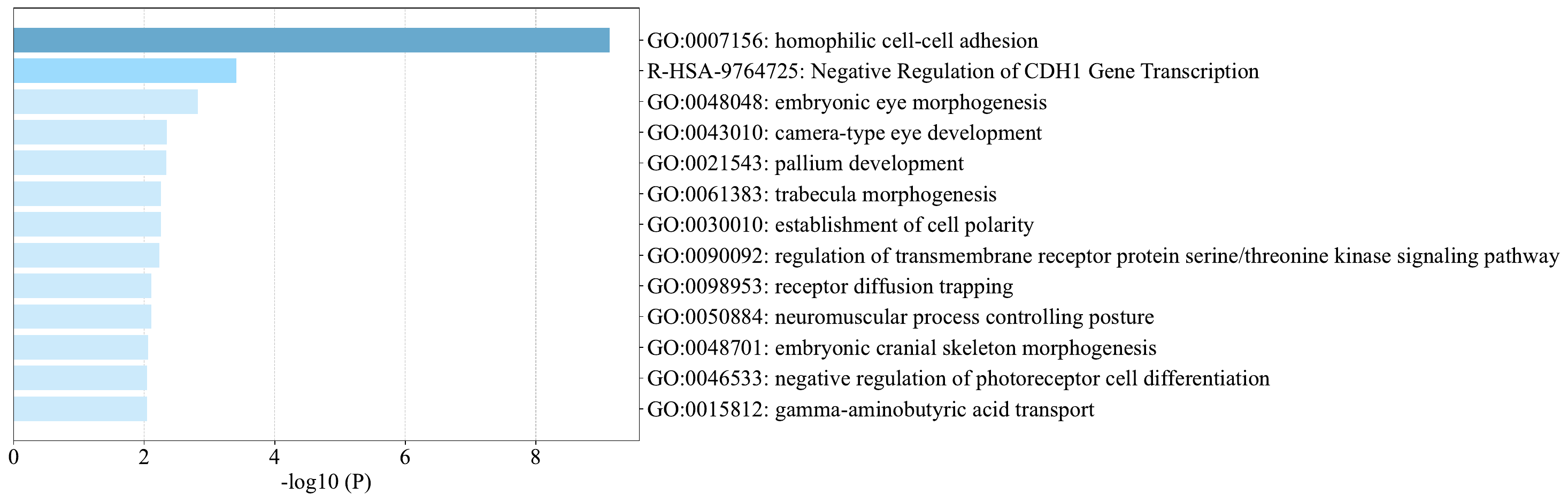}
        \subcaption{Pathway and process enrichment analysis.}
        % \label{subfig:b}
    \end{subfigure}

    \begin{subfigure}[t]{0.58\textwidth}  % [t] 顶端对齐
        \centering
        \includegraphics[width=\linewidth, valign=t]{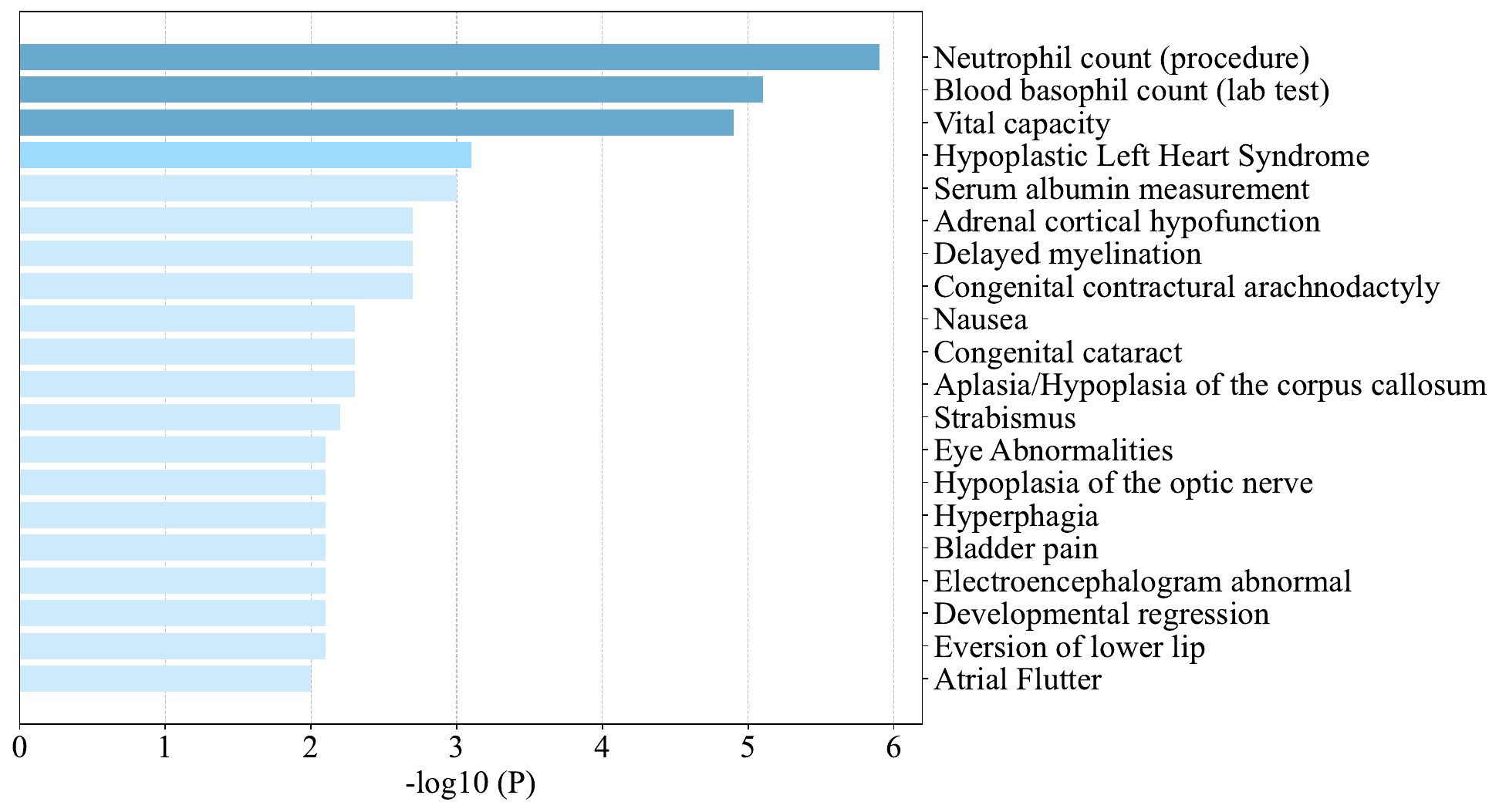}
        \subcaption{Summary of enrichment analysis in DisGeNET.}
    \end{subfigure}
    \hfill  % 自动填充列间空白
    \begin{subfigure}[t]{0.38\textwidth}  % [t] 顶端对齐
        \centering
        \includegraphics[width=\linewidth, valign=t]{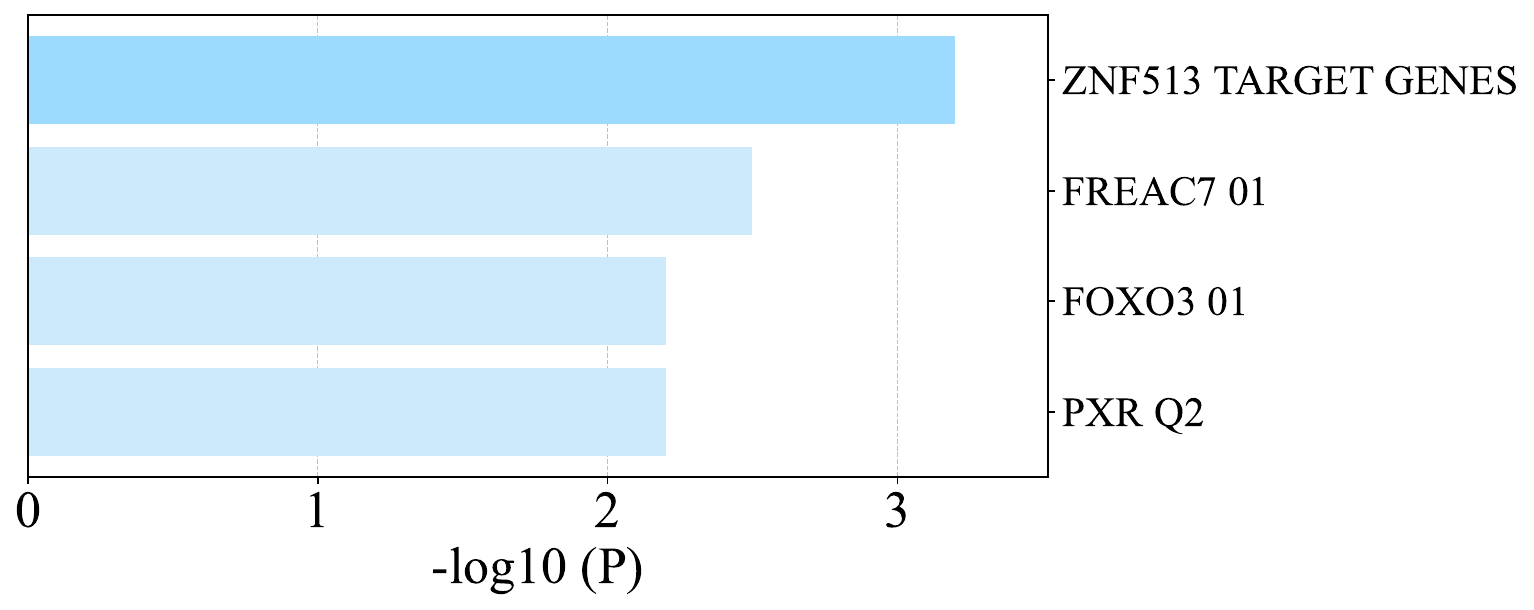}
        % 可选：添加子图标题（不需要则注释）
        \subcaption{Summary of enrichment analysis in Transcription Factor Targets}
        % \label{subfig:2a}
    \end{subfigure}

    \caption{Functional enrichment analysis of aging-related signals identified from the GEO methylation dataset.}
    \label{fig:bio_analysis_geo}
\end{figure}

We first characterize the genomic distribution of these CpG sites using Illumina HumanMethylation450K hg19 annotation~\cite{bibikova2011high}. All CpGs are successfully mapped and are predominantly located in gene-associated regions and CpG islands, indicating a non-random distribution across genomic elements. We then map these CpGs to approximately 245 genes and perform functional enrichment and network analyses using Metascape~\cite{zhou2019metascape} and STRING~\cite{szklarczyk2021string}. As shown in Fig.~\ref{fig:bio_analysis_geo} (a), the enriched terms include cell adhesion, developmental processes, cell-cycle regulation, MAPK signaling, and immune-related functions. To further characterize these features, we perform external annotation analyses. As shown in Fig.~\ref{fig:bio_analysis_geo} (b), DisGeNET~\cite{pinero2020disgenet} associations show links to hematological traits such as neutrophil and basophil counts, suggesting relevance to immune-related processes. Transcription factor target analysis further identifies enrichment for FOXO3 targets~\cite{flachsbart2009association}, a transcription factor previously implicated in longevity and aging-related regulation, which is summarized in Fig.~\ref{fig:bio_analysis_geo} (c).

Overall, these results show that functionally relevant biological processes are identified in both modalities, suggesting that the model captures meaningful molecular signals across different omics data types.

\section{Analysis of the Generative Component}\label{app:ablation}
The AgeMoE model integrates a generative replay module that preserves information from previously visited clients and augments local training with pseudo-samples during cross-client updates. To assess its contribution under the \model training process, we compare the full model with a variant without the generative module, while keeping all other training settings unchanged. As the generative replay component influences how information from different clients is accumulated during collaborative training, its effect is most clearly reflected in the model’s overall generalization across centers. We therefore evaluate performance on the global test set aggregated from all clients.

Experiments are conducted on the three datasets used in the main study, including the UKB-Center, UKB-Imbalance, and GEO-Methylation datasets. In addition, we construct an additional dataset from the UK Biobank cohort to simulate stronger cross-client distributional differences, referred to as UKB-AgeSplit. In this dataset, samples are partitioned into clients according to age intervals of five years (e.g., 40–45, 45–50), resulting in six clients, each containing individuals within a specific age range. This setting reflects practical scenarios in which medical centers serve distinct patient populations, such as pediatric or geriatric hospitals, leading to systematic differences in data distributions across centers.

\begin{figure}[htbp]
    \centering
    % 全局间距优化：减小列间距，保证布局紧凑
    \setlength{\tabcolsep}{0.5em}
    \renewcommand{\arraystretch}{1.0}

    % ========== 第一排：左侧文字块(0.48) + 右侧子图a(0.48) ==========
    \begin{subfigure}[t]{0.48\textwidth}  % [t] 顶端对齐
        \centering
        \includegraphics[width=\linewidth]{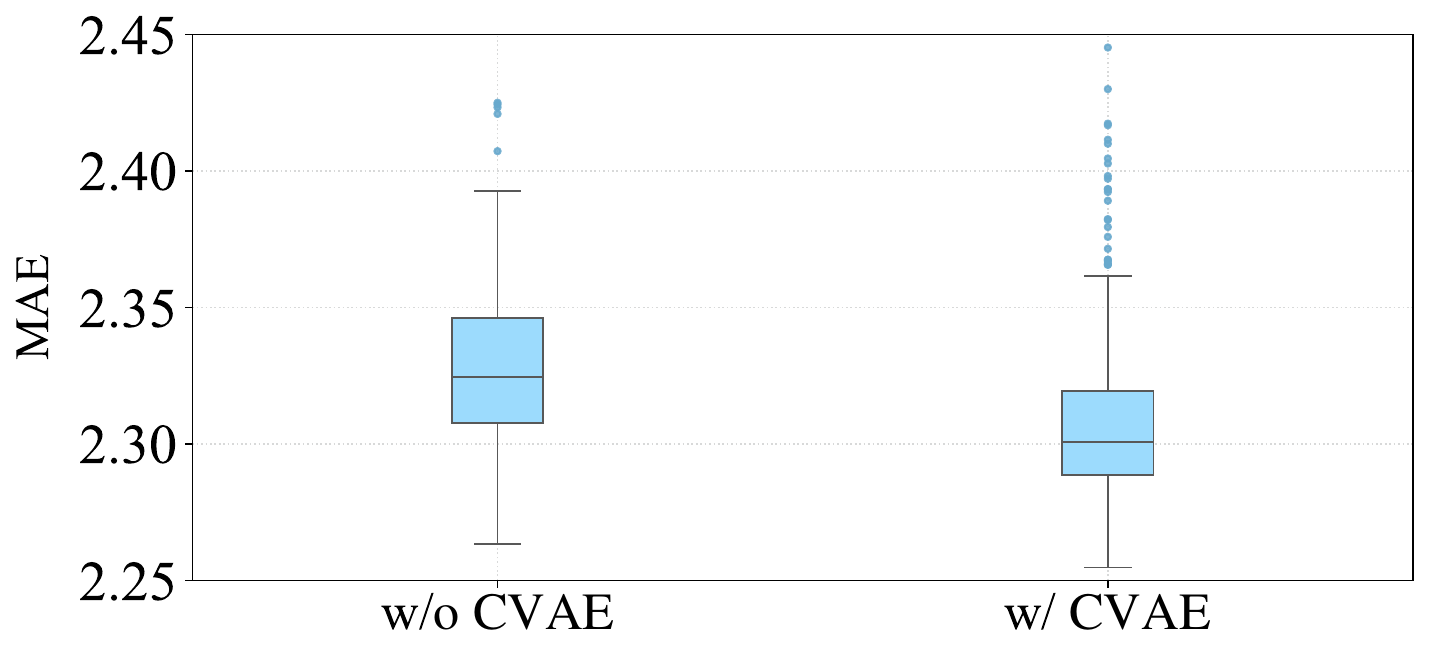}
        \subcaption{UKB-Center}
    \end{subfigure}
    \hfill  % 自动填充列间空白
    \begin{subfigure}[t]{0.48\textwidth}  % [t] 顶端对齐
        \centering
        \includegraphics[width=\linewidth]{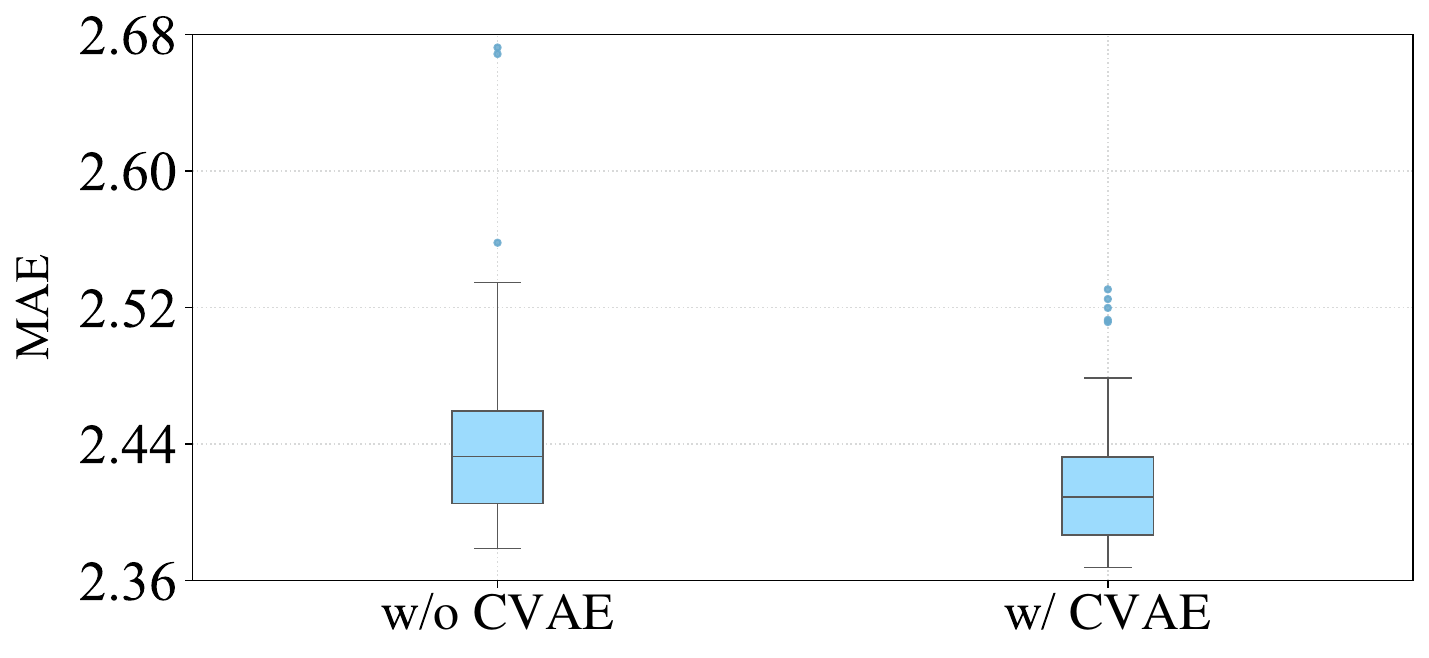}
        % 可选：添加子图标题（不需要则注释）
        \subcaption{UKB-Imbalance}
        % \label{subfig:2a}
    \end{subfigure}

    % \vspace{0.2em}  % 行间距，可按需调整

    % ========== 第二排：子图b(0.48) + 子图c(0.48) ==========
    \begin{subfigure}[t]{0.48\textwidth}
        \centering
        \includegraphics[width=\linewidth, valign=t]{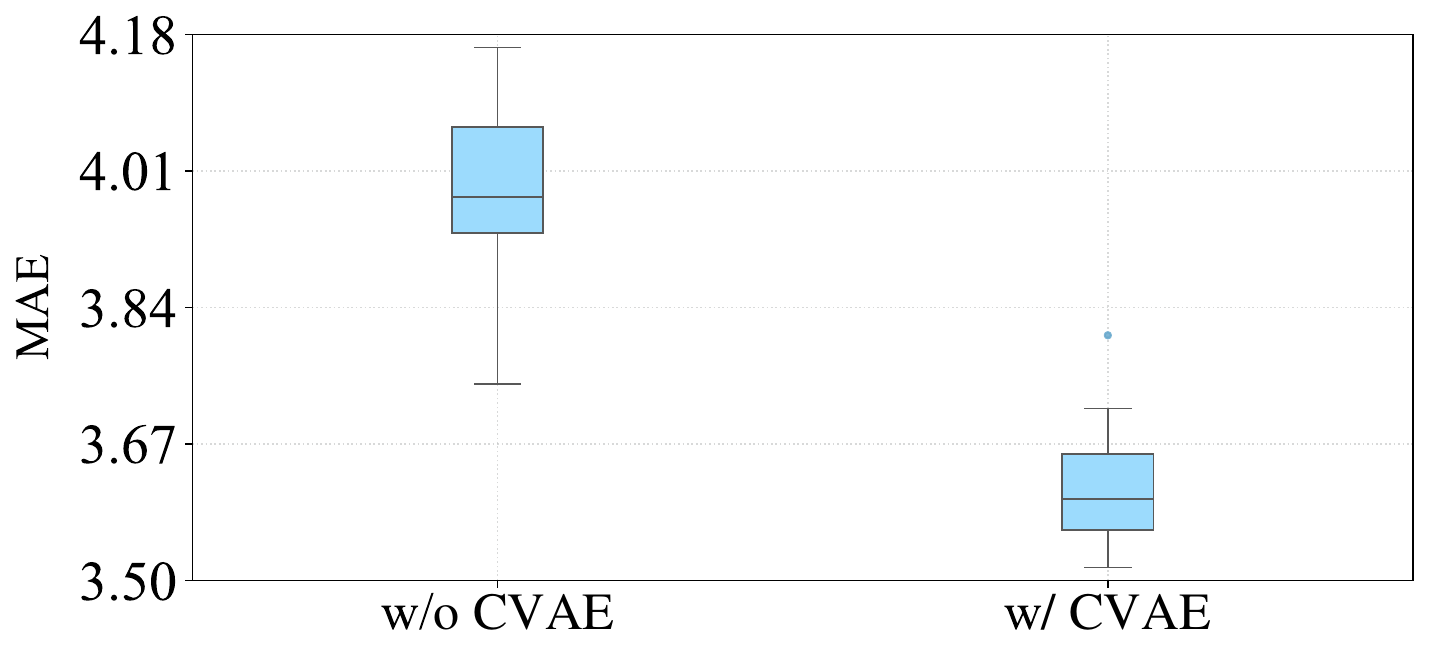}
        \subcaption{GEO-Methylation}
        % \label{subfig:b}
    \end{subfigure}
    \hfill
    \begin{subfigure}[t]{0.48\textwidth}
        \centering
        \includegraphics[width=\linewidth, valign=t]{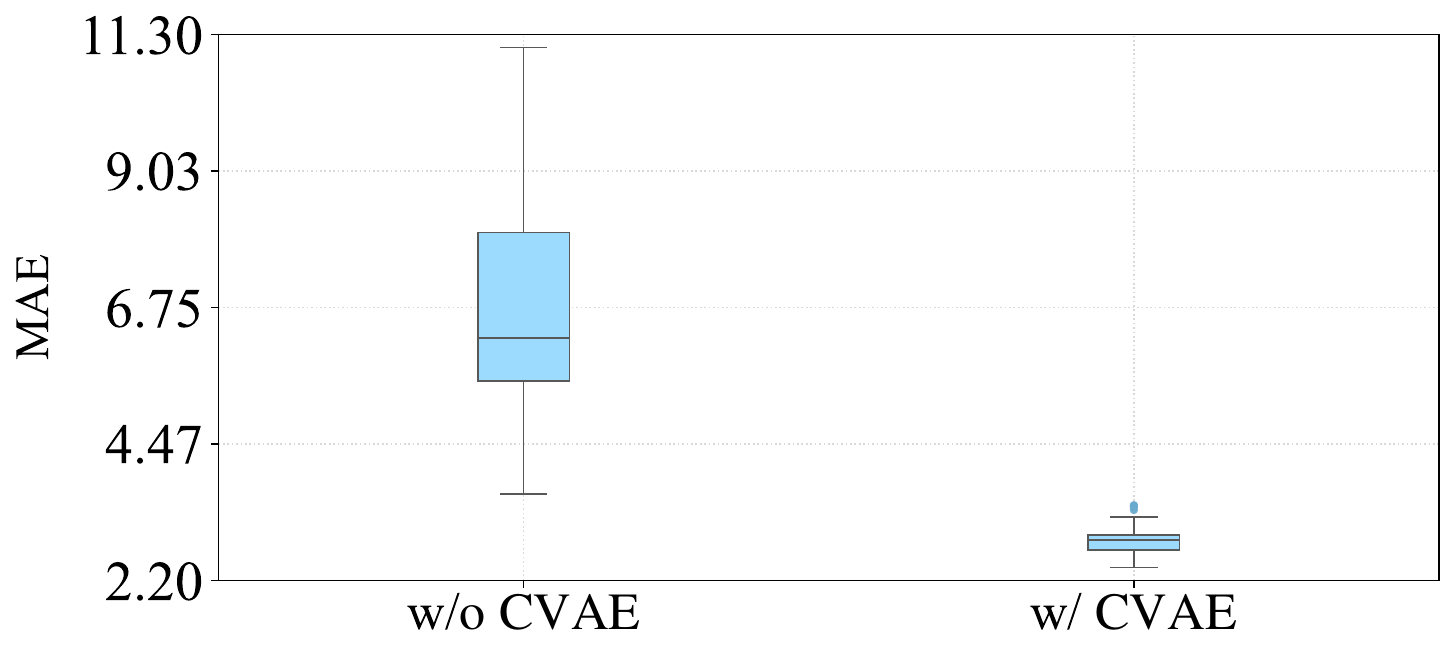}
        \subcaption{UKB-AgeSplit}
        % \label{subfig:c}
    \end{subfigure}

    \caption{Performance comparison of AgeMoE with (w/ CVAE) and without (w/o CVAE) the generative component on global test sets across multiple datasets.}
    \label{fig:ablation}
\end{figure}

Fig.~\ref{fig:ablation} reports the global test performance of the two model variants across all datasets. Incorporating the generative component leads to improved global predictive performance, with larger gains observed for the GEO-Methylation dataset and the age-partitioned UK Biobank dataset, both of which exhibit stronger cross-client distributional differences. These observations indicate that the generative component is particularly beneficial when training involves heterogeneous client data. By retaining representative information from previously visited clients and augmenting local updates with pseudo-samples, the generative model helps maintain coverage of the overall data distribution and preserve information acquired from earlier clients, leading to improved global generalization.

\section{Details on Robustness Metrics}\label{app:robustness}
This section provides formal definitions and calculation procedures for the three robustness metrics used to evaluate \model under varying interaction orders over the trust network. The metrics quantify how sensitive the model’s predictive performance is to different sequences of client interactions.

\subsection{Client-level MAE Range}
For each client $k$, we compute the range of MAE values across all interaction orders on the local test set. Let $\text{MAE}_{k}^{(i)}$ denote the MAE of client $k$ under interaction order $i$, then the client-level range is defined as:
\begin{equation}
    \text{Range}_k = \max_i \text{MAE}_{k}^{(i)} - \min_i \text{MAE}_{k}^{(i)}.
\end{equation}
Smaller $\text{Range}_k$ indicates more stable performance across interaction sequences. To visualize these ranges, we aggregate $\text{Range}_k$ values across all clients for each model and plot them as boxplots, which show the distribution of per-client MAE variability. Each box represents the spread of $\text{Range}_k$ values across clients, providing an intuitive comparison of robustness among models.

\subsection{Cumulative MAE Degradation}
Cumulative MAE degradation quantifies how sequential center updates affect the overall model performance. To capture the effect across all clients rather than individual local behaviors, MAE is measured on the global test set. Let $\text{MAE}_{\text{global}}^{(i,j)}$ denote the global MAE after the $j$-th client update in the $i$-th interaction order. Then the cumulative degradation for order $i$ is defined as:
\begin{equation}
    \text{Degradation}^{(i)} = \sum_{j=2}^{T} \max(0, \text{MAE}_{\text{global}}^{(i,j)} - \text{MAE}_{\text{global}}^{(i,j-1)}),
\end{equation}
where $T$ is the number of update steps. Lower $\text{Degradation}^{(i)}$ indicates that the sequential updates do not substantially increase error. To visualize these cumulative degradation values, we collect $\text{Degradation}^{(i)}$ across all interaction orders for each model and plot them as boxplots. Each box represents the distribution of cumulative MAE increases across sequences, providing a concise comparison of how different models accumulate errors during sequential updates.

\subsection{Forgetting Rate}
Forgetting quantifies the relative performance loss of clients over the course of sequential updates. 
To focus on the retention of knowledge at the client level, we compare each center's MAE immediately after its own update with its MAE under the final model produced at the end of the same propagation sequence. Specifically, for client $k$ appearing at position $j$ in order $i$, let $\text{MAE}_{k}^{(i,j)}$ and $\text{MAE}_{k}^{(i,T)}$ denote the MAE after its own update and at the final model, respectively. 
The forgetting rate is then defined as:
\begin{equation}
    \text{Forgetting}_k^{(i)} = \frac{\text{MAE}_{k}^{(i,T)} - \text{MAE}_{k}^{(i,j)}}{\text{MAE}_{k}^{(i,j)}}\times 100\%.
\end{equation}
Positive values indicate that the client’s performance deteriorates over subsequent updates, while negative values suggest improvement or retention of knowledge. 
To focus on early-stage effects, we calculate $\text{Forgetting}_k^{(i)}$ only for the first half of clients in each sequence, as these clients are more likely to be affected by subsequent updates. To visualize client-level forgetting, we collect $F_k^{(i)}$ for all clients in the first half of each sequence and plot them as boxplots. 
Each box captures the distribution of these forgetting rates across sequences, enabling comparison of how different models vary in client-level performance retention.

\section{Details About Performance Under Simulated Measurement Variability}\label{app:perturb}

To assess \model's stability under simulated multi-center measurement variability, we conducted a perturbation study using the UKB-Center dataset. We simulate measurement variability by modifying each feature with two components: (1) \textbf{Client-level shift}: a systematic deviation shared by all samples from a given center, modeling biases arising from instrument calibration, experimental protocols, batch effects, or site-specific measurement tendencies; (2) \textbf{Sample-level noise}: an independent fluctuation applied to individual samples, capturing variability due to sample processing, measurement precision, or local technical noise. For each feature, the perturbed value was:
\begin{equation}
    \textbf{x}_{perturbed} = \textbf{x}_{original} + \text{client\_shift} + \text{sample\_noise},
\end{equation}
with the perturbation magnitudes set to introduce controlled inter- and intra-center variability (client shift $\sim$35\% of the global feature standard deviation, sample noise $\sim$12\%). This choice introduces controlled deviations representing plausible inter- and intra-center differences.

Predictive performance was evaluated on individual client and global test sets, with results summarized in Table~\ref{tab:perturb}. Minimal performance decline was observed under these perturbations, with the global mean MAE increasing by only 0.07 (from 2.33 on the original data to 2.40 on the perturbed data), indicating that \model maintains stable predictive performance under the simulated measurement variability.

\begin{table}[!t]
\centering
\caption{MAE of \model on individual clients and the global test set under \textbf{original} and \textbf{perturbed} data. Results are split into two groups for readability.}
\label{tab:perturb}
\begin{tabular}{c|ccccccccc}
\hline
\textbf{Client} & 0 & 1 & 2 & 3 & 4 & 5 & 6 & 7 & 8 \\
\hline
\textbf{Original} & 2.2714 & 2.2662 & 2.2637 & 2.2990 & 2.2330 & 2.1098 & 2.4706 & 2.3462 & 2.4036 \\
\textbf{Perturbed} & 2.3878 & 2.3677 & 2.3244 & 2.3749 & 2.3070 & 2.1959 & 2.5332 & 2.4181 & 2.5120 \\
\hline
\end{tabular}

% \vspace{2mm} % 空行分组

\begin{tabular}{c|ccccccccc|c}
\hline
\textbf{Client} & 9 & 10 & 11 & 12 & 13 & 14 & 15 & 16 & 17 & Global \\
\hline
\textbf{Original} & 2.3677 & 2.4112 & 2.2739 & 2.3568 & 2.2513 & 2.2841 & 2.4482 & 2.2943 & 2.5144 & 2.3259 \\
\textbf{Perturbed} & 2.4128 & 2.4621 & 2.3313 & 2.4651 & 2.2944 & 2.3450 & 2.5705 & 2.3496 & 2.5892 & 2.4023 \\
\hline
\end{tabular}
\label{tab:client_mae_split}
\end{table}

\section{Details About Performance Across Trust-Network Structures}\label{app:network}
In the main experiments, interaction sequences are constructed by randomly sampling permutations of clients, providing a controlled setting to evaluate order-dependent variability when each client participates once in a training sequence. To further examine how explicit trust-network structures affect model behavior, we conduct additional experiments on the UKB-Center dataset using trust-network-derived interaction sequences. Specifically, we generate random trust networks over clients with controlled sparsity ranging from $0.1$ to $0.5$ to simulate varying degrees of inter-client trust connectivity, and then identify propagation sequences of variable lengths that cover all clients while respecting directed trust relations. These sequences may include repeated visits to centers when necessary, reflecting feasible interaction patterns in sparsely connected trust networks. The resulting sequences are used in \model without modifying the underlying sequential update mechanism or training protocol.

In this setting, we focus on predictive performance and robustness to interaction patterns. Experiments on model compatibility with different aging-clock backbones are not repeated, as they primarily evaluate the flexibility of the \model propagation strategy with respect to model choice and are not directly determined by how interaction sequences are instantiated. All experiments are conducted on the UKB-Center dataset, which provides a representative setting due to its relatively large number of clients and heterogeneous data distributions across centers.

\begin{figure}[!t]
    \centering
    % 全局间距优化：减小列间距，保证布局紧凑
    \setlength{\tabcolsep}{0.5em}
    \renewcommand{\arraystretch}{1.0}

    % ========== 第一排：左侧文字块(0.48) + 右侧子图a(0.48) ==========
    \begin{subfigure}[t]{0.48\textwidth}  % [t] 顶端对齐
        \centering
        \includegraphics[width=\linewidth]{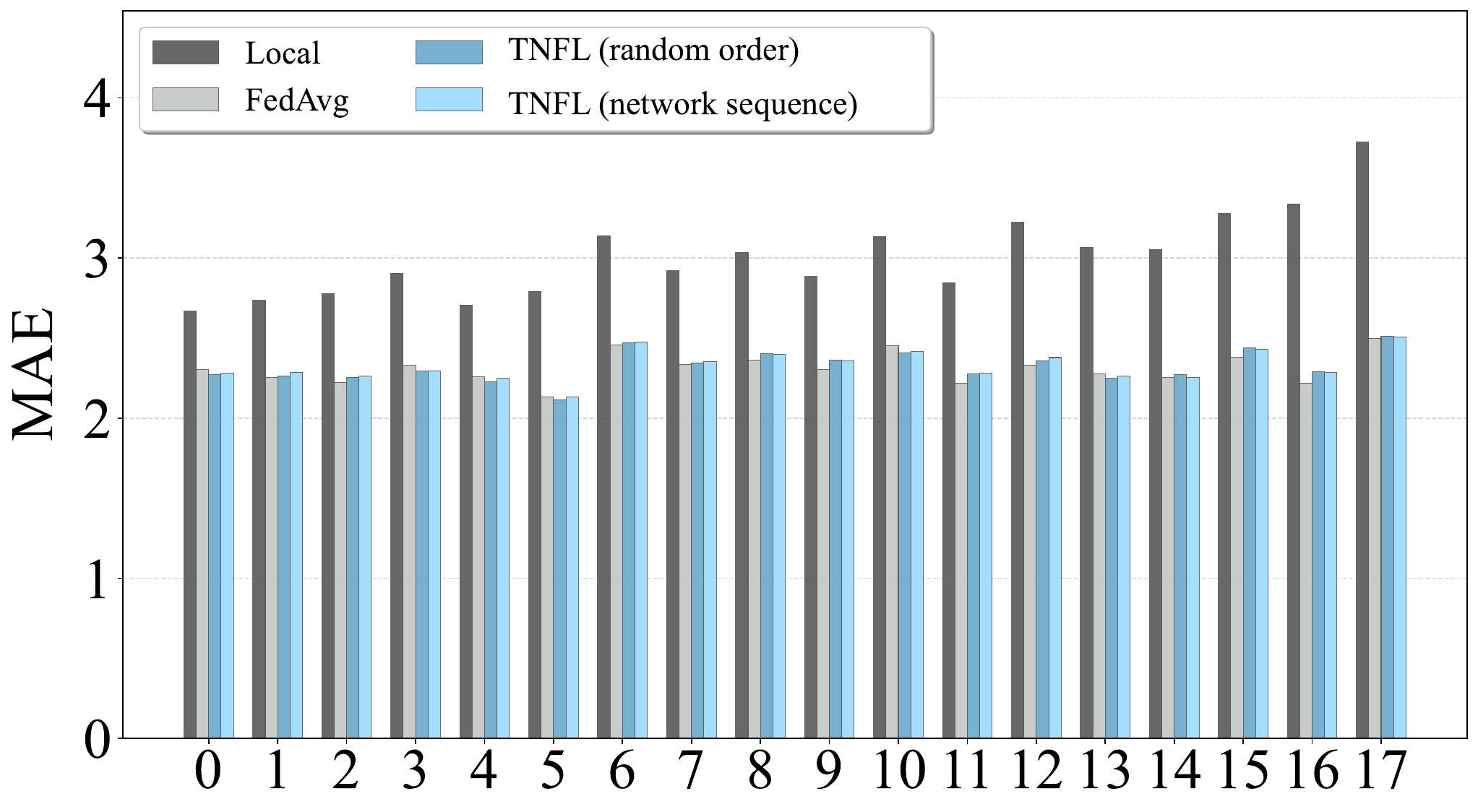}
        \subcaption{Multi-center aging clock performance}
    \end{subfigure}
    \hfill  % 自动填充列间空白
    \begin{subfigure}[t]{0.48\textwidth}  % [t] 顶端对齐
        \centering
        \includegraphics[width=\linewidth]{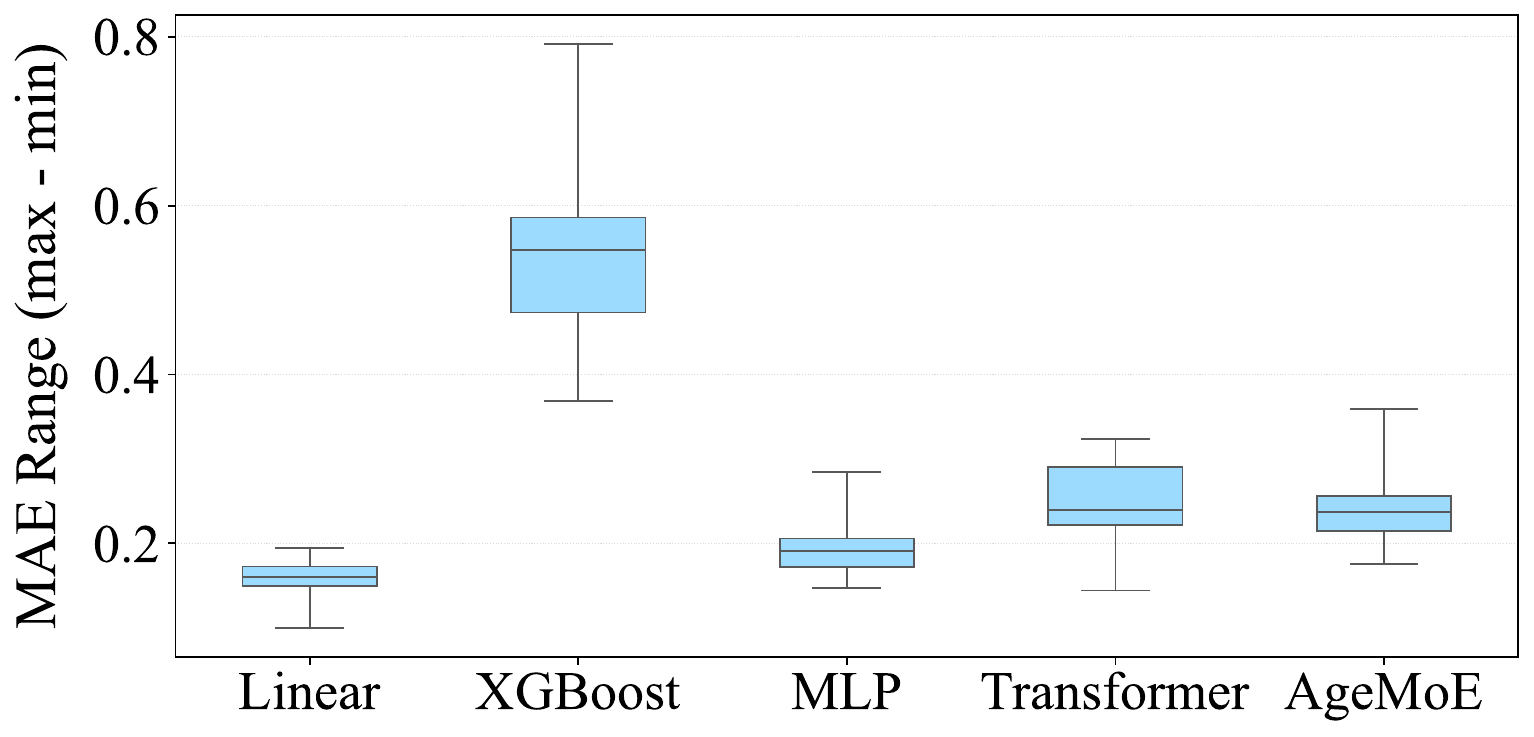}
        % 可选：添加子图标题（不需要则注释）
        \subcaption{Robustness: Client performance range}
        % \label{subfig:2a}
    \end{subfigure}

    \vspace{0.2em}  % 行间距，可按需调整

    % ========== 第二排：子图b(0.48) + 子图c(0.48) ==========
    \begin{subfigure}[t]{0.48\textwidth}
        \centering
        \includegraphics[width=\linewidth, valign=t]{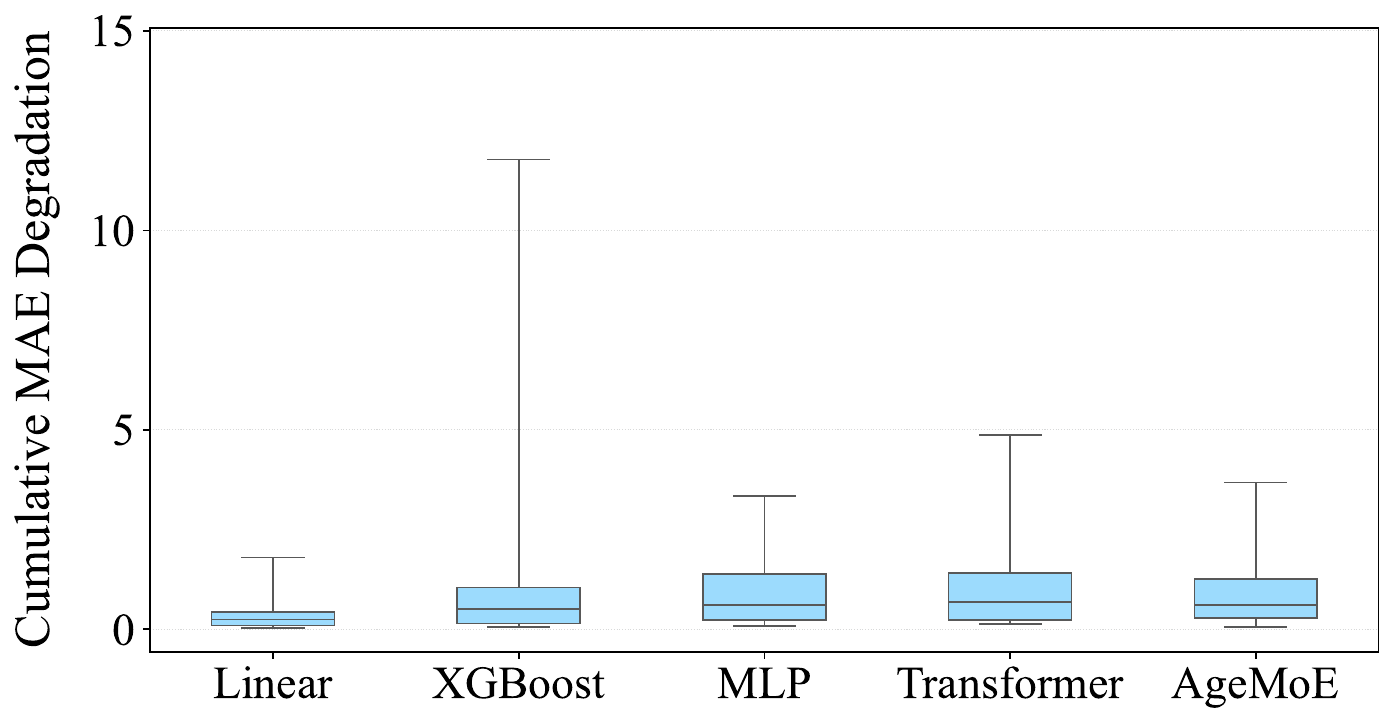}
        \subcaption{Robustness: Cumulative degradation}
        % \label{subfig:b}
    \end{subfigure}
    \hfill
    \begin{subfigure}[t]{0.48\textwidth}
        \centering
        \includegraphics[width=\linewidth, valign=t]{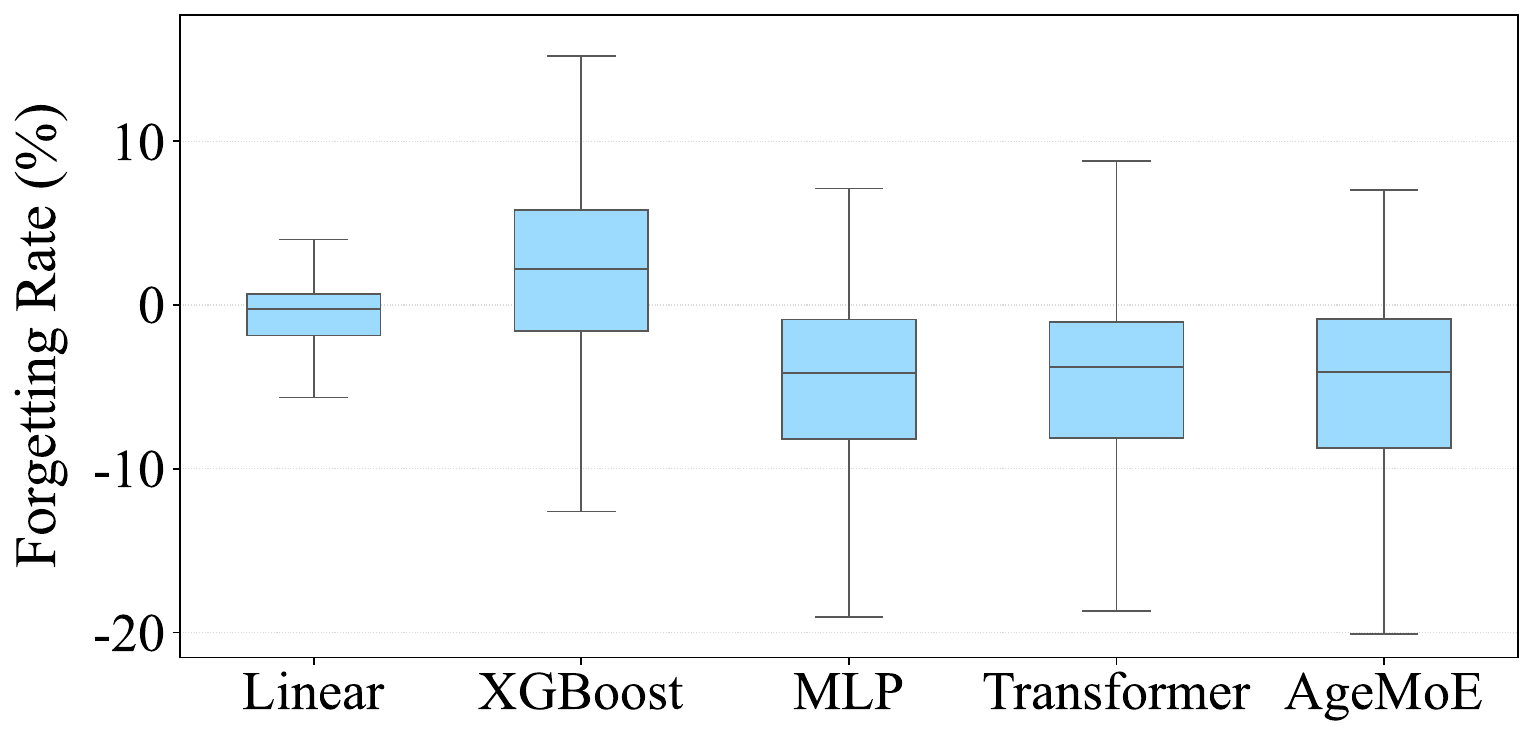}
        \subcaption{Robustness: Forgetting rate}
        % \label{subfig:c}
    \end{subfigure}
    \caption{Performance comparison and robustness evaluation under trust-network-derived interaction sequences.}
    \label{fig:network_supplement}
\end{figure}

We first compare predictive performance across four settings: independent local training, FedAvg, \model with randomly sampled interaction orders, and \model with trust-network-derived interaction sequences. As shown in Fig.~\ref{fig:network_supplement} (a), the performance patterns remain consistent across these settings. \model achieves similar predictive performance under both sequence-generation settings, while maintaining its relative advantages over local training and remaining competitive with FedAvg.

We further evaluate robustness using the same three metrics as in the main study. As shown in Fig.~\ref{fig:network_supplement} (b)-(d), the models exhibit consistent stability patterns under trust-network-derived interaction sequences. \model maintains low variability across interaction arrangements, limited cumulative degradation, and no observable increase in forgetting effects. The cumulative degradation metric aggregates performance changes along the propagation sequence and is therefore naturally dependent on sequence length; since trust-network-derived interaction sequences may involve repeated visits to clients and thus longer propagation sequences, the absolute magnitude is not directly comparable in scale to that under random orders. Nevertheless, the observed cumulative degradation remains small, with a median value below 0.5, indicating stable performance progression throughout training. Overall, these results are consistent with those obtained under randomly sampled interaction orders.

Overall, these results suggest that \model shows broadly consistent behavior across different valid interaction sequences instantiated from the underlying trust network, with predictive performance and robustness metrics exhibiting similar trends across random-permutation and trust-network-derived sequence-generation schemes.

\end{appendices}

%%===========================================================================================%%
%% If you are submitting to one of the Nature Portfolio journals, using the eJP submission   %%
%% system, please include the references within the manuscript file itself. You may do this  %%
%% by copying the reference list from your .bbl file, paste it into the main manuscript .tex %%
%% file, and delete the associated \verb+\bibliography+ commands.                            %%
%%===========================================================================================%%
\newpage
\bibliography{sn-bibliography}% common bib file
%% if required, the content of .bbl file can be included here once bbl is generated
%%\input sn-article.bbl

\end{document}